# Multi-teacher knowledge distillation as an effective method for compressing ensembles of neural networks

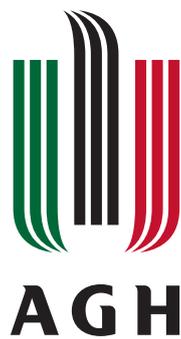

Konrad Zuchniak

AGH University of Science and Technology

Faculty of Computer Science, Electronics, and Telecommunication

Institute of Computer Science

Dissertation for the degree of

*Doctor of Philosophy*

Supervisor: Prof. dr hab. inż. Witold Dzwinel

Co-supervisor: Dr  Robert Szczelina

Kraków, 29.09.2022

# Destylacja wiedzy wielu nauczycieli jako efektywna metoda kompresji zespołów sieci neuronowych

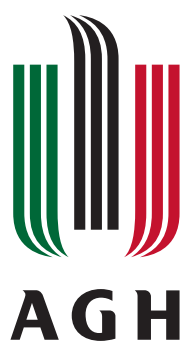

## Konrad Zuchniak

Akademia Górniczo-Hutnicza
im. Stanisława Staszica w Krakowie
Wydział Informatyki, Elektroniki i Telekomunikacji
Instytut Informatyki

*Praca Doktorska*

Promotor: Prof. dr hab. inż. Witold Dzwinel
Promotor pomocniczy: Dr Robert Szczelina

Kraków, 29.09.2022

# Acknowledgements

I am deeply indebted to my supervisor Prof. dr hab. inż. Witold Dzwinel and co-supervisor Dr Robert Szczelina for support in solving research problems and all valuable tips that allowed me to carry out this research to the very end. I am also thankful to the students who collaborated with me for our fruitful cooperation and the experiments they conducted, which helped me in my dissertation. Special thanks to MSc Emilia Majerz and MSc Aleksandra Pasternak. I am also grateful to Prof. Krzysztof Dragan and his team for sharing the DAIS dataset and working together on the automatic corrosion detection project. I would like to extend my sincere thanks to Prof. Stan Matwin and Prof. Jerzy Komorowski for their substantive guidance in this project. Many thanks to my family and friends for their support and understanding. These experiments were carried out within the infrastructure of the ACK Cyfronet computing center, for which I am also grateful.

# Abstract

Deep learning has contributed greatly to many successes in artificial intelligence in recent years. Breakthroughs have been made in many applications such as natural language processing, speech processing, or computer vision. Recently, many techniques have been implemented to enable the training of increasingly larger and deeper neural networks. Today, it is possible to train models that have thousands of layers and hundreds of billions of parameters on powerful GPU or TPU clusters. Large-scale deep models have achieved great success, but the enormous computational complexity and gigantic storage requirements make it extremely difficult to implement them in real-time applications, especially on devices with limited resources, which is very common in the case of offline inference on edge devices.

On the other hand, the size of the dataset is still a real problem in many domains. Data are often missing, too expensive, or impossible to obtain for other reasons. Ensemble learning is partially a solution to the problem of small datasets and overfitting. By training many different models on subsets of the training set, we are able to obtain more accurate and better-generalized predictions. However, ensemble learning in its basic version is associated with a linear increase in computational complexity. We check if there are methods based on Ensemble learning which, while maintaining the generalization increase characteristic of ensemble learning, will be immediately more acceptable in terms of computational complexity. As part of this work, we analyze the various aspects that influence ensemble learning: We investigate methods of quick generation of submodels for ensemble learning, where multiple checkpoints are obtained while training the model only once. We analyzed the impact of the ensemble decision-fusion mechanism and checked various methods of sharing the decisions including voting algorithms. Finally, we used the modified knowledge distillation framework as a decision-fusion mechanism which allows the addition compressing of the entire ensemble model into a weight space of a single machine learning model.

We showed that knowledge distillation can aggregate knowledge from multiple teachers in only one student model and, with the same computational complexity, obtain a better-performing model compared to a model trained in the standard manner. We have developed our own method for mimicking the responses of the teacher's models using a student model, which consists of imitating several teachers at the same time, simultaneously. We tested these solutions on several benchmark datasets. In the end, we presented a wide application use of the efficient multi-teacher knowledge distillation framework, with a description of dedicated software developed by us and examples of two implementations. In the first example, we used knowledge distillation to develop models that could automate corrosion detection on aircraft fuselage. The second example describes the detailed implementation process of the technology in a solution enabling the detection of smoke on observation cameras in order to counteract wildfires in forests.

# Streszczenie

Głębokie uczenie przyczyniło się w ostatnich latach do wielu sukcesów w dziedzinie sztucznej inteligencji. Dokonano przełomu w wielu zastosowaniach, takich jak przetwarzanie języka naturalnego, analiza mowy czy widzenie komputerowe. W ostatnim czasie wdrożono wiele technik umożliwiających trenowanie jeszcze większych i głębszych modeli. Dziś możliwe jest trenowanie modeli, które mają tysiące warstw i setki miliardów parametrów na potężnych klastrach GPU lub TPU.

Wielkoskalowe modele głębokie osiągnęły wielki sukces, ale ogromna złożoność obliczeniowa i gigantyczne wymagania dotyczące zasobów sprzętowych, sprawiają że ich implementacja w aplikacjach wymagających działania w czasie rzeczywistym jest niezwykle trudna, zwłaszcza na urządzeniach o ograniczonych zasobach, co jest bardzo powszechne w przypadku analizy danych offline, bez dostępu do chmury obliczeniowej.

Z drugiej strony rozmiar zbioru danych nadal stanowi prawdziwy problem w wielu dziedzinach. Często brakuje danych, są one zbyt drogie lub niemożliwe do uzyskania z innych powodów. Uczenie zespołowe jest częściowo rozwiązaniem problemu małych zbiorów danych i przetrenowania. Trenując wiele różnych modeli na podzbiorach zbioru treningowego, jesteśmy w stanie uzyskać dokładniejsze i lepiej generalizujące modele. Jednak uczenie zespołowe w swojej podstawowej wersji wiąże się z liniowym wzrostem złożoności obliczeniowej. W niniejszej pracy dokonano sprawdzenia czy istnieją metody oparte na uczeniu zespołowym, które przy zachowaniu charakterystycznego dla uczenia zespołowego przyrostu generalizacji będą jednocześnie bardziej akceptowalne pod względem złożoności obliczeniowej. W ramach tej pracy zbadano różne aspekty, które wpływają na uczenie zespołowe: Metody szybkiego generowania pojedynczych modeli do uczenia zespołowego, gdzie uzyskuje się wiele punktów zapisu wag modelu podczas tylko jednego procesu trenowania. Sprawdzono wpływ zespołowego mechanizmu fuzji decyzji i sprawdzono różne metody uwspólniania decyzji, w tym algorytmy głosowania. Wykorzystano zmodyfikowaną strukturę destylacji wiedzy jako mechanizm fuzji decyzji, który dodatkowo pozwala skompresować wagi wielu modeli całego zespołu w pojedynczy model uczenia maszynowego. Destylacja wiedzy może agregować wiedzę pochodzącą od wielu modeli nauczycieli tylko w jeden model uczenia maszynowego i przy tej samej złożoności obliczeniowej umożliwia uzyskanie lepszej wydajność w porównaniu z modelem wytrenowanym w standardowy sposób bez wsparcia modeli nauczycieli.

Opracowana została autorska metoda naśladowania odpowiedzi modeli nauczycieli za pomocą modelu ucznia, polegającą na naśladowaniu kilku nauczycieli jednocześnie. Powyżej opisane rozwiązania zostały sprawdzone na wybranych zbiorach danych popularnych w domenie uczenia maszynowego. Na koniec przedstawiliśmy szerokie zastosowanie metody wykorzystującej destylację wiedzy z wielu modeli nauczycieli, wraz z opisem dedykowanego oprogramowania oraz z przykładami dwóch wdrożeń.

# Contents

# List of Tables

# List of Figures

<div style="text-align: right">

</div>

# Introduction

## Contents

Deep learning has contributed greatly to many successes in artificial intelligence in recent years. Breakthroughs have been made in many applications, such as natural language processing [1], speech recognition [2] or computer vision [3]. Recently, many techniques have been implemented to enable the training of even larger and deeper models. An example would be the residual connections [4, 5] that significantly helped counteract the vanishing gradient or batch normalization [6] that coordinate the update of multiple layers. Today, it is possible to train models that have thousands of layers and hundreds of billions of parameters [7, 8] on powerful GPU or TPU clusters. Large-scale deep models have achieved great success, but the enormous computational complexity and gigantic storage requirements make it extremely difficult to implement them in real-time applications, especially on devices with limited resources, which is very common in the case of offline inference on edge devices equipped with machine learning accelerators such as Coral Edge TPU [9] or Nvidia Jetson Xavier [10].

## 1.1   Motivation

In machine learning applications, we often have to find a balance between the quality of prediction of a machine learning model and the cost of training and inference. In addition, the size of the dataset is still a real problem in many domains. Data are often missing, too expensive, or impossible to obtain for other reasons. Ensemble learning is partially a solution to the problem of small datasets and overfitting. By training many different models on subsets of the training set, we are able to obtain more accurate and better generalized predictions. However, ensemble learning in its basic version is associated with a linear increase in computational complexity, which is unacceptable in many applications. From an implementation point of view, the motivation to carry out the research presented in this dissertation was the need to develop a mechanism that will improve the trade-off between the quality of prediction and the costs associated with training and maintaining the model, assuming that obtaining a dataset is also not free because it requires the expenditure of time and money to obtain and label examples.

## 1.2   The theses, goals and contributions

In scientific terms, we wanted to check if there are methods based on Ensemble learning that, while maintaining the generalization increase characteristic for ensemble learning, will be immediately more acceptable in terms of computational complexity. As part of this work, we analyze the various aspects that influence ensemble learning. Our activities can be divided into three sub-areas:

- Methods for quick generation of submodels for ensemble learning. There are approaches to obtain multiple checkpoints while training the model once. We examined how the performance of ensembles composed of such submodels compares to ensembles composed of independently trained models.

- We found that the impact of the submodel decision-fusion mechanism was often marginalized. We checked various methods of sharing the decisions of individual models and, as it translated into the quality of the predictions of the entire ensemble, special attention was paid to the use of voting algorithms[11].

- Finally, we decided to use the modified knowledge distillation framework as a decision fusion mechanism, but also to compress the entire ensemble model into a single machine learning model.

The last point is related to the our theses, the goals, and the contributions. Is it possible to use knowledge distillation to aggregate knowledge from multiple *teachers* in a single *student* model and, with the same computational complexity, obtain a better performing model compared to a model of the same computational complexity trained in the standard manner? Our contribution can be summarized in the following points:

- We analyze the decision-fusion methods used in ensemble learning, with particular emphasis on voting algorithms. We have identified limitations that accompany certain methods and potential solutions to these limitations.

- We analyze cyclical methods for generating checkpoints that enable the generation of many submodels during a single training cycle. However, we observed a reduced variety of models obtained in this way, leading to inferior performances compared to ensembles composed of independently trained submodels.

- We have thoroughly analyzed the use of knowledge distillation in ensemble learning. We analyzed $multi - teacher$ architectures in which we have many *teachers* and single *student* model with learns from them simultaneously. This approach enables the aggregation of knowledge accumulated on the weights of several models in a single weight space of a *student* model.

- We have developed our own method for mimicking the responses of $teachers'$ models using a $student$ model, which consists of imitating several $teachers$ at the same time, simultaneously. It is forced by an appropriately modified loss function. Our research has shown that in this way we can obtain better performance $students$ compared to the standard variant when $teachers'$ responses are averaged in the preprocessing stage before actual knowledge distillation.

- For the purposes of the implementation part of this work, a software package was developed that uses the insights obtained in the course of this research. Our software libraries significantly automate the process of generating a $student$ model using the $multi-teacher$ knowledge distillation framework.

- This software library was used in the SmokeFinder project implemented by Neuralbit Technologies. SmokeFinder is a system for the automatic detection of forest fire smoke using machine vision and analysis of images from observation cameras. Thanks to automation and support of the forest fire observation process, it is possible to detect fires faster and thus significantly reduce losses caused by wildfires. We have built a dataset containing photos of smoke and trained on it a light smoke detection model using our $multi-teacher$ knowledge distillation framework. We have obtained the operational capability that enables the early detection of smoke and fumes that are barely visible to the naked eye.

## 1.3   Overview of the dissertation

We divided this work into two main parts. In the first part (Chapters *Methodology* and *Results*), we focused on the scientific aspect, in which we included the theoretical description of our analyzes and the results of experiments obtained on benchmark datasets for the machine learning industry. In the second part, Implementation (Chapter *Implementation*), we describe the practical application of this work. A library using a framework $multi-teacher$ knowledge distillation and examples of implementing this technology in real projects are described. The rest of the document is organized as follows:

- **Chapter Methodology** describes selected theoretical aspects related to ensemble learning and knowledge distillation. We begin with a discussion of classical ensemble learning techniques such as bagging, boosting, and stacking. Then, the connotations between ensemble learning and modern neural networks were discussed, including the concept of stacked neural networks. We then move on to discuss the $decision-fusion$ mechanisms that are currently used. We place special emphasis on voting algorithms and show that the most common voting method, plurality voting, does not generate the best decisions at all; there are other more sophisticated voting methods that are capable of generating better performing ensembles, with the same computational complexity, just by modifying the decision-fusion method. Finally, we move on to the discussion of the knowledge distillation framework, which enables the transfer of knowledge from a larger model to a much lighter model. We show various applications of this method and its incredible flexibility and generality. Finally, we move on to discuss a modification of basic knowledge distillation: $multi-teacher$ knowledge distillation, which uses multiple trained models as $teachers$. We describe several methods to mimic $teachers$ by a $student$ model, including the mechanism we developed to follow all $teachers$ simultaneously.

- **Chapter Results** contains a description of the experiments we performed. In the first part, we showed several experiments on small dense neural networks. We built ensemble models with different numbers of submodels and analyzed how the performance of the entire ensemble is affected by the number of submodels and the decision-fusion method. The next section presents the results of the analysis of cyclic methods of generating submodels by saving checkpoints of model weights from different stages of training; then, we conducted a comparative analysis with independently trained models and their ensembles. The last section describes the $multi-teacher$ knowledge distillation framework and the results we obtained using this methodology on benchmark datasets. The chapter is concluded with a section summarizing the scientific part of this work.

- **Chapter Implementation** contains the utility and application aspects of this work. The first section describes in detail the software library that we have developed that enables the automation of the generation of machine learning models using the $multi-teacher$ knowledge distillation framework. The processes of generating data subsets, preprocessing, training $teachers$ and training the $student$ model using fine-tuning and hyperparameter search are automated. Next, we present the results of the collaboration with the Air Force Institute of Technology (AFIT).While working on this project a system that automates the process of corrosion detection on aircraft fuselage plating was developed. The system is based on the use of DAIS imaging technology. We then describe in detail the SmokeFinder project, which aims to enable automatic detection of wildfire smoke by analyzing images from surveillance cameras in forests. We conclude the chapter with a summary of the implementation aspects of this work.

# 2

# Methodology

## Contents

## 2.1 Ensemble learning

Ensembling models of different types (for example, formal mathematical and data models) often leads to better results, that is, better approximations, predictions, or classification accuracy [12, 13]. The main purpose of the ensemble of models is to increase the accuracy of the predictions. Especially in the case of weak learners [14, 15]. The aggregated predictions of the ensemble are generally more accurate than the single-model predictions. However, ensemble learning can increase the computational complexity of the predictive model, both in terms of training and inference time. In the classic implementation of the ensembling of neural networks, every single submodel is generated in an independent training process. Consequently, $N$ independent models increase the computational complexity of the classifier $N$ times, both in the training and inference phases. Therefore, the high demand for computational resources required by big data models remains a challenging problem.

### 2.1.1 Classic approach

The ensemble learning [12, 13] is based on the idea that combining multiple models can produce more powerful models. The Ensemble Learning technique uses multiple models, also called weak learners, combined to obtain better results, stability, and predictive performance compared to a single model. It is enough that a weak learner will be characterized by an accuracy greater than the random classifier (for the binary classification problem, the accuracy must be greater than $0.5$). Furthermore, weak learners should be as unstable as possible so that they differentiate well. The more differentiated the classifiers, the lower their bias and variance will be in their ensemble. The regression problem was the first issue for which a theoretical description of the effectiveness of the use of ensemble learning was performed [16]. Brown et al. [17] and Krogh and Vedelsby [18] show using ambiguity decomposition that by using an ensemble learning model we get predictions with smaller squared error compared to individual models prediction. The ambiguity decomposition is given as follows:

$$E[o-y]^2 = bias^2 + \frac{1}{M}var + (1 - \frac{1}{M})covar,$$

$$bias = \frac{1}{M}\sum_i (E[o_i] - y),$$

$$var = \frac{1}{M}\sum_i E[o_i - E[o_i]]^2, \tag{2.1}$$

$$covar = \frac{1}{M(M-1))}\sum_i \sum_{j \neq i} E[o_i - E[o_i]][o_j - E[o_j]],$$

where $y$ is target, $o_i$ is the output of $i^{th}$ model and $M$ is the ensemble size. Here, the *bias* term measures the average difference between the base learner and the model output, *var* indicates their average variance, and *covar* is the covariance term that measures the pairwise difference of the base learners. Noise, bias, and variance are one of the main reasons for the prediction and classification errors of machine learning models. These factors can be minimized using ensemble learning techniques that increase the stability of the model. Figure 2.1 shows a schematic, graphical illustration of bias and variance. The bagging [19, 20], boosting [21, 22] and stacking [23] are the three main concepts of ensemble learning implementation.

**Bagging**

Bootstrap aggregation, or bagging, is a technique that can be used with many classification and regression methods to reduce the variance of the prediction and improve the prediction process. This is an ensemble learning method that generates different ensemble learners by modifying the training dataset. Bagging can be described with a relatively simple idea: a subset of samples (bootstrap samples) from training available data are taken many times. In each of these bootstrap samples, some prediction method is applied. Finally, to obtain the overall prediction, the results are combined, by averaging for regression and simple majority voting for classification. These combined predictions tend to have reduced variance due to averaging. Weak models are trained on data subsets and are generated by random sampling with replacement from the original training data. Replacement implies that the sample of the data subset used to train each model may have duplicate data. The final output of an ensemble model is the value generated using decision-fusion of outputs of individual submodels, typically using majority voting for a classification problem or an average for a regression problem. The bagging technique is shown in Figure 2.2. The bagging procedure is as follows:

- Generate multiple training subsets from the entire training dataset, selecting samples with replacements.

- Create and train weak learners (base models) in parallel and independently of each training subset.

- Decision-fusion combines the predictions of all weak learners to get the final ensemble prediction.

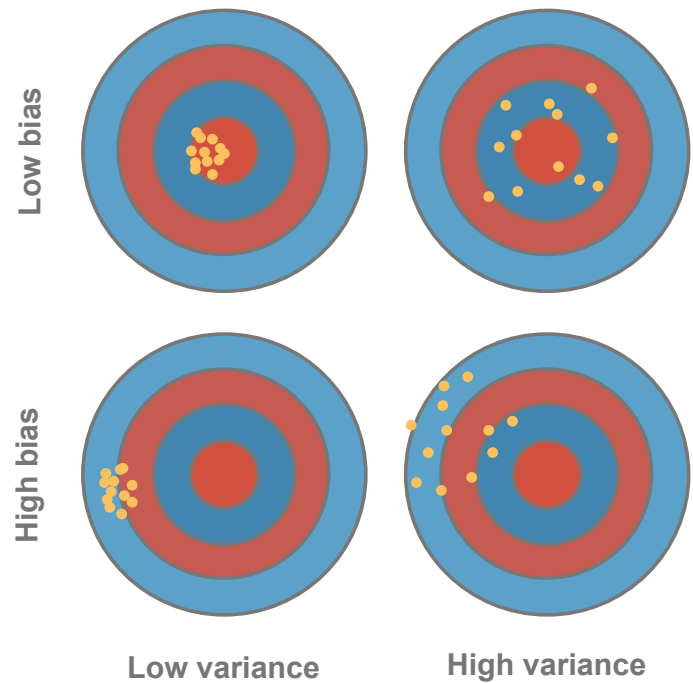

**Figure 2.1:** Graphical illustration of bias and variance.

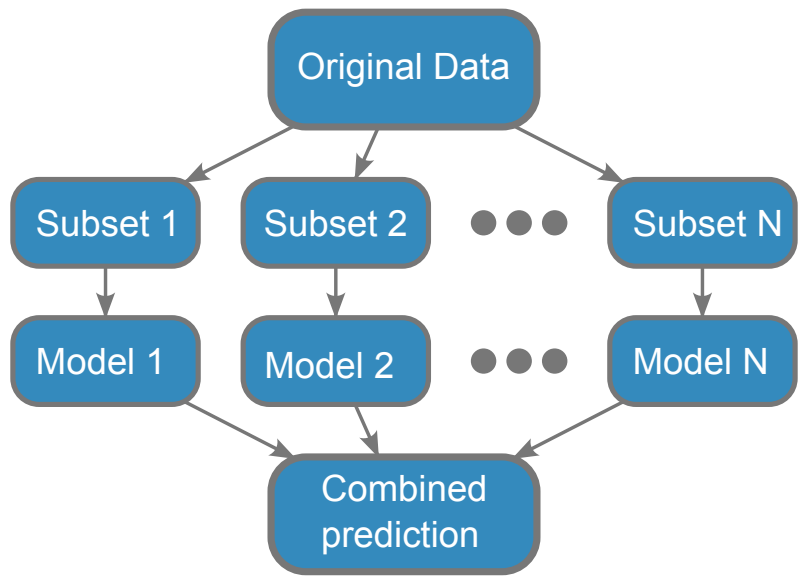

**Figure 2.2:** An illustration of the concept of bagging.

**Boosting**

Boosting, as bagging, is an ensemble-based approach that can be used to improve the accuracy of classification or regression models. Boosting is an ensemble technique that focuses on changing training data and adjusting the weight of observations based on the performance of previous weak learners. Unlike the bagging approach, boosting cannot be done in parallel and involves dependence on weak learners. Weak learners take the results of the previous weak learner into account and adjust the weights of the data points, which converts the weak learner into a strong learner. Also, with boosting, the samples used at each step are not all drawn in the same way from the same population, but rather the incorrectly predicted cases from a given step are given increased weight during the next step. Boosting changes the weight associated with an observation that was classified incorrectly by trying to increase the weight associated with it. Boosting tends to decrease bias error, but can sometimes lead to overfitting of the training dataset. Figure 2.3 shows boosting concept.

The boosting procedure is as follows:

• The training dataset is created where all data points have equal weights.

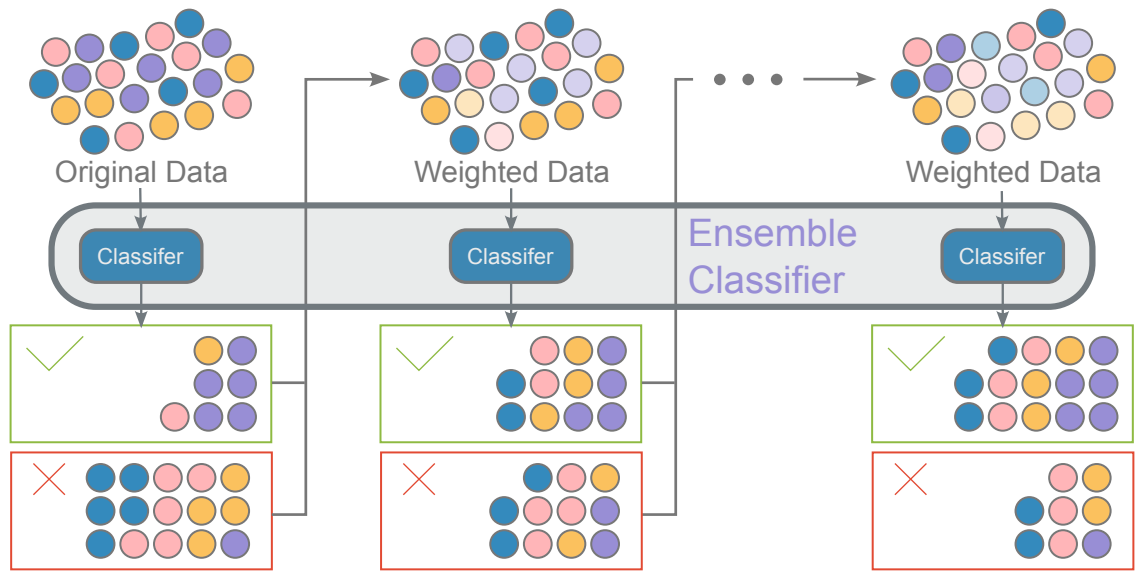

**Figure 2.3:** An illustration of the concept of boosting.

- On the initial dataset, a model is created on the basis of this. This model is used to make predictions on the entire dataset.

- Model errors are calculated using ground truth and model estimations. The observation that was predicted incorrectly will have a greater weight in the next iteration.

- The following model is created, and the boosting attempts to correct the errors of the previous model.

- The process is repeated for multiple models, each correcting the errors of the previous model.

- Finally, the ensemble model is a strong learner and uses the weighted mean of all weak learners.

One of the most powerful boosting techniques used today is GBM [24] (Gradient Boosting Machine) and its modifications. In contrast to random forest [25], with is bagging of decision trees (which combines the output of multiple decision trees to give the prediction, the trees can be trained in parallel, where each tree is trained on a different subsample of the original data), GBM is the boosting technique, and takes a sequential approach. Gradient refers to the error obtained after building a model. Boosting refers to improving prediction performance. Each subsequent tree is trained to take into account the responses of the others. The examples incorrectly tagged by the remaining trees will be weighted higher. In this way, the performance of all trees grows incrementally. GBM often wins kaggle [26] competitions and obtains state-of-the-art results, especially on tabular data [27–29]. Figure 2.4 shows the concept of a gradient boosting machine.

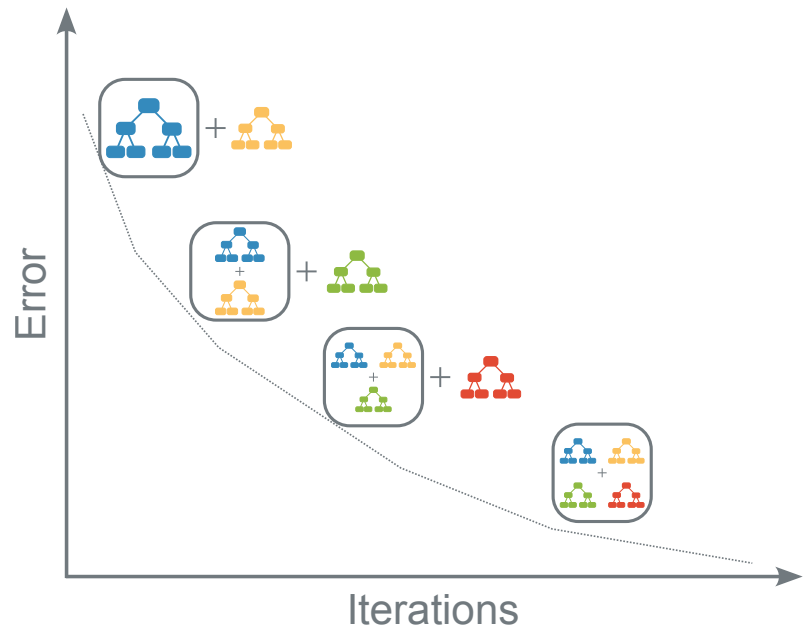

**Figure 2.4:** Illustration of a Gradient Boosting Machine.

**Stacking**

In averaging methods, there are multiple submodels that contribute equally to ensemble prediction in averaging techniques. Model averaging can be improved by weighting each submodel contribution to the ensemble prediction by measuring. Finally, it can improve ensemble prediction. This can be further extended by training a completely new model to learn how to best combine contributions from each submodel. This approach is called stacked generalization [16, 23, 30–32], or stacking. When stacking, the algorithm takes the output of the submodels as input and learns how to best combine the input predictions to achieve better output predictions. Stacking predictive performance generally is better than the predictive performance of individual unstacked models. We can divide the stacking method into two levels:

- Base models are trained to make predictions based on the training dataset.

- The meta-learning model is used to integrate the submodels predictions from the previous step and attempts to learn how to best combine them to get the best results.

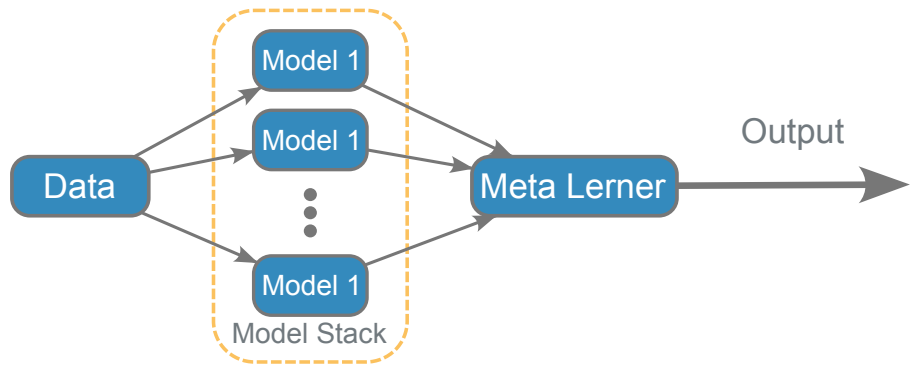

**Figure 2.5:** An illustration of the concept of stacking.

Unlike weighted average ensembles, stacked generalization ensembles can use a set of predictors as a context and conditionally choose to weight the input predictions differently, potentially leading to better performance. It is very important to use different training sets for base model training and meta-learning to avoid overfitting. The stacking technique is shown in Figure 2.5.

### 2.1.2 Decision-fusion schemes

The ensemble learning aggregates the results from the submodels using some decision-fusion rules used to combine the predictions of the submodels is one of the components that determines the effectiveness of the entire ensemble [16]. Most of the time, ensemble designers focus on ensemble architecture and use naive averaging to predict ensemble output. On the other hand, simple naive averaging, used in most ensemble learning applications, leads to non-optimal performance [33] as it is sensitive to the performance of the used submodels. To find better solutions, different decision-fusion approaches have been introduced, such as the Bayes optimal classifier or the super-learner. Here, we present different approaches to combining the submodel's outputs inside ensemble decision-fusion:

- **Unweighted Model Averaging**

  Simple submodels outputs averaging is the most widely used decision-fusion approach. In this baseline method, the outcomes of the base learners are averaged to obtain the final prediction of the ensemble learning model. Deep learning models have a tendency to have low bias and high variance, so an ensemble using simple averaging can increase generalization because it reduces the variance between submodels. Averaging the output of individual submodels can be performed either directly on the output of the base learners or on the predicted classification probabilities via the softmax function.

- **Bayesian Optimal Classifier**

  In this decision-fusion method, with each submodel, there is connected hypothesis $h_i$ which determines the conditional probability distribution $P(y|x)$ of the target class $y$ given the input $x$. Assuming that $h_i$ is a hypothesis generated on the basis of the training set $D$ and evaluated on the test data $(x, y)$ which can be expressed by the formula $h_i(y|x) = P[y|x, h_i, D]$. Using the Bayes rule, we can write:

$$P(y|x, D) \propto \sum_{h_i} P[y|h_i, x, D] P[D|h_i] P[h_i] \qquad (2.2)$$

and the Bayesian Optimal classifier can be written as:

$$\underset{y}{argmax} \sum_{h_i} P[y|h_i, x, D] P[D|h_i] P[h_i] \qquad (2.3)$$

where $P[D|h_i] = \prod_{(y,x) \in D} h_i(y|x)$ is the likelihood of the data under $h_i$.

- **Stacked Generalization**

  Stacked generalization [23] works by finding the bias of generalizers based on the training set provided. To obtain good weights for the linear combination of the outputs of the base learners in regression, cross-validation data and least squares error optimization was used to find the optimal weights for the combination of decisions [34]. Base learners $f_1, f_2, \ldots, f_m$ predictions linear combination can be given as:

  $$f_s(x) = \sum_{i=1}^{m} w_i f_i(x) \tag{2.4}$$

  where $w$ is the optimal weight vector with which the meta-learner is trained.

- **Plurality Voting**

  Plurality (also called majority) voting aggregate submodels outputs similarly to an unweighted averaging decision-fusion method. However, instead of performing probability output averaging, majority voting counts the votes (each submodel votes for the label with the highest probability) of submodels, without taking into account the probability result. During the decision-fusion process, this method counts the votes of all submodels and selects a label with the majority of votes as the final label prediction. Plurality voting is not the only voting algorithm that can be used in decision-fusion process inside ensemble models [35]. There are more sophisticated and better performing voting algorithms. Later in this chapter, a whole section is devoted to the different voting mechanisms, where they are described in more detail.

The choice of the decision sharing method between the submodels included in the ensemble is an important stage in the design of the entire ensemble learning algorithm. Although the basic variants are the most common, an appropriate decision-fusion method can enhance the overall model performance, as we have shown in the following chapters.

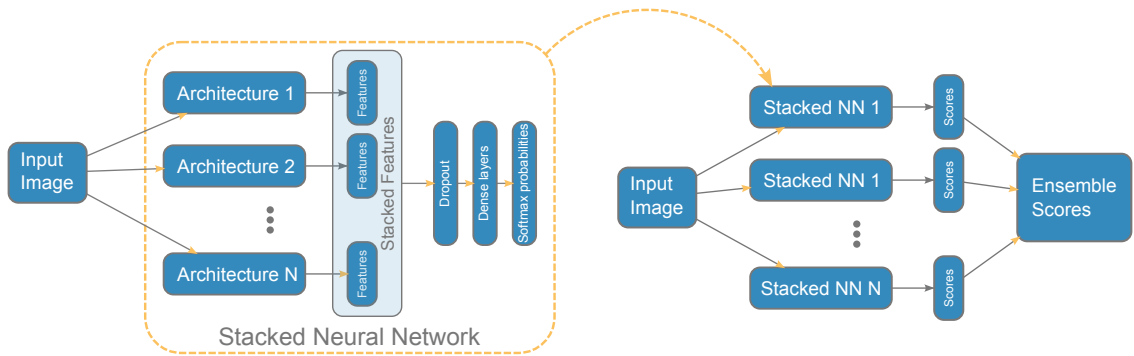

**Figure 2.6: Left**: A stack of few neural network architectures. The features generated from each network are concatenated into a feature vector. **Right**: Ensemble build with few stacked neural networks.

### 2.1.3  Ensemble methods in deep learning

Ensemble learning, as it is a very general technique, can also be used to combine deep neural networks. An interesting method of connecting neural networks with various architectures that utilize the concept of previously discussed stacking was created. Stacked Neural Networks (S-NN) [36] combines different architectures of neural networks, such as GoogLeNet [37], or VGG16 [38], but do not directly aggregate classification probabilities, only the activation of earlier layers, most often the activation of the last convolutional layer. Features from different models are concatenated, then pass through the dropout layer for better generalization, and, using the dense layer, are projected onto the final predictions. The authors also propose ensemble models where the submodels are stacked neural networks, composed of various architectures. On the other hand, this approach is extremely inefficient in terms of inference time. Figure 2.6 shows the scheme of a stacked neural network and the method of their ensembling. Another interesting idea presents the Deep Stacking Network (DSN) [39, 40] architecture. Deep Stacking Network design philosophy is based on the stacking concept. Where simple modules of classifiers or functions are composed and next they are "stacked" on top of each other so as to learn complex classifiers or functions. DSN architecture consists of many stacking modules, each of which takes a simplified form of a shallow multilayer perceptron. Figure 2.7 shows an example of a four-module DSN, each consisting of three sub-layers (input layer, one hidden layer and output layer) and is illustrated with a separate color. The yellow dashed lines denote layer duplication and the gray solid lines denote connections between layers. Stacking is accomplished by concatenating the output predictions of all previous modules with the original input vector to form the new input vector for the next module.

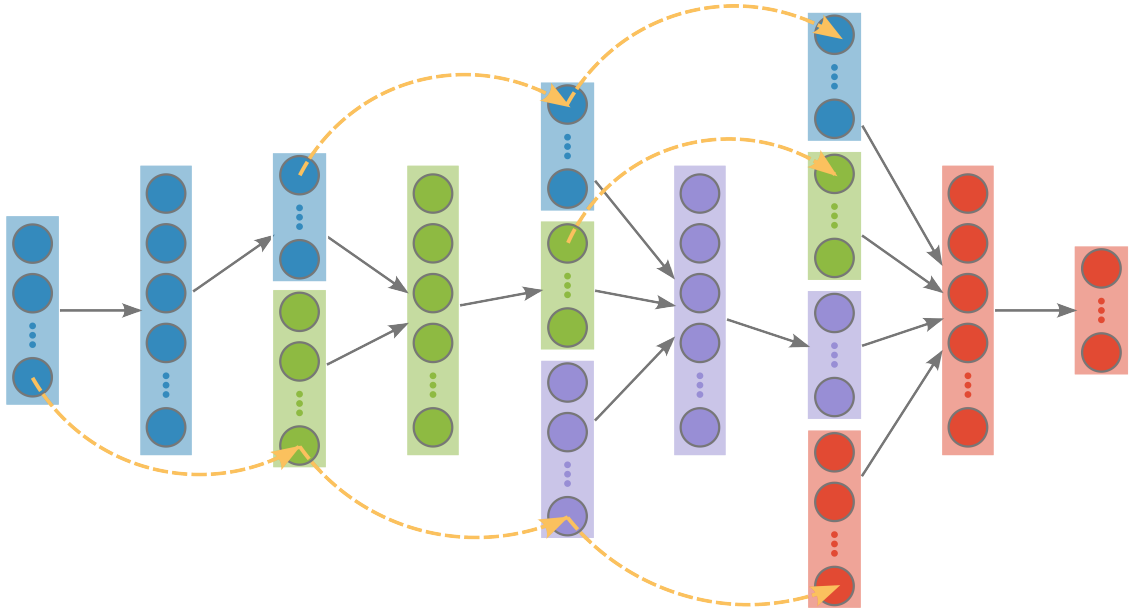

**Figure 2.7:** An example of a four-module DSN, each consisting of three sub-layers (input layer, one hidden layer and output layer) and being illustrated with a separate color. Yellow dashed lines denotes layer duplication's and gray solid lines denotes connections between layers Stacking is accomplished by concatenating all previous modules' output predictions with the original input vector to form the new input vector for next module.

There are also other studies showing attempts to combine ensemble learning and deep neural networks [41–44], and the potential of these applications [45–47]. It is also worth mentioning that dropout [48], a groundbreaking regularization technique in which some neurons are switched off from the network in each iteration during the training phase, can also be considered as an ensemble of different neural networks [49, 50]. Submodels can be represented by a separate set of neurons that are turned on at any given moment, and during the inference phase, all neurons are turned on. Large language models such as BERT [1] or GPT-3 [8] based on the architecture of Transformers [51] somehow use the concept of ensembling. The key element of the transformers is the multi-head attention layer; as the name suggests, this layer creates several identical attention modules (with different, separately initialized weights) that learn to pay attention to other aspects of the input data during training, it is a certain implementation of the concept of bagging. Transformer architecture also draws an analogy to stacking. The basic block creating the transformer architecture consists of a multi-head self-attention mechanism, followed by a simple, positionwise fully connected feed-forward network. The output of both of these elements is summed with the unchanged input and then normalized. This mechanism causes partial preservation of the original input for processing by subsequent transformer layers. Figure 2.8 schematically shows the transformer encoder building blocks.

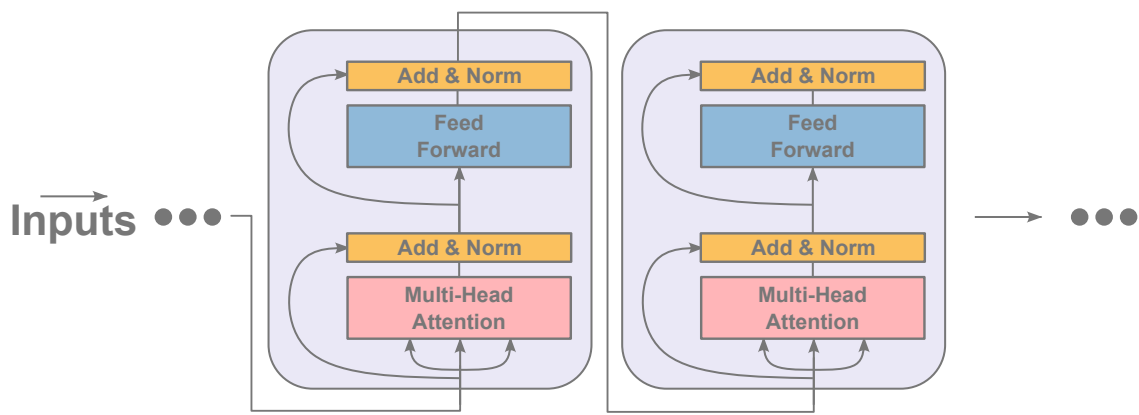

**Figure 2.8:** Building blocks of transformer encoder. Multi-head attention layer creates several identical attention modules that learn to pay attention to other aspects of the input data during training (bagging analogy). The output of self-attention layer and positionwise fully connected feed-forward network is summed with the unaltered input and then normalized (stacking analogy). This mechanism causes partial preservation of the original input for processing by subsequent transformer layers.

Another example also from the field of natural language processing is the Mixture of Experts [52] mechanism. It also uses the concept of training several parallel elements that focus on their specializations in the training process. In this work, we focus on two issues that appear to be less explored:

1. Note that, in general, ensemble learning methods omit the aspect of the decision-making process (in this work, we most often use the term $decision-fusion$). This stage is often marginalized; the simplest method of plurality voting or averaging (simple or weighted) is used. On the other hand, voting theory develops interesting algorithms that optimize the voting process.

During the initial work on this dissertation, we analyzed how to use voting theory for ensemble learning.

2. The work mentioned above does not focus on the problem of computational complexity of ensemble models. Assuming that the ensemble is composed of submodels of the same or similar complexity, the computational complexity of the entire ensemble system increases linearly with increasing number of submodels included in the ensemble model. This increase occurs in both the training and inference phases. Some methods have been developed that shorten the training time of the ensemble (they will be discussed later in this chapter); however, the problem of complexity at the time of inference still remains. Considering that machine learning is a highly iterative process, optimization of the inference time seems to be a more important problem.

## 2.2 Voting rules in decision-fusion

### 2.2.1 Assumptions

In the domain of voting systems [53–55] we assume that voting systems in which the voter can spend only one vote on only one candidate are unable to correctly reflect the entire preferences of the voter. This method omits the order of preferenced candidates. In the domain of voting theory, systems with these limitations are generally called plurality voting. In voting algorithms, we assume that every voter can sort the candidates according to their preferences. At the beginning of the list, the candidate most liked was found, while at the end, the candidate least liked was found. In our approach, we use voting algorithms for the submodels' decision-fusion process inside the ensemble classifier. Individual models are treated as voters, while possible decisions (classifications) are treated as candidates. We would like to mention research on the plurality voting [35, 56] algorithm and other voting methods [11] in pattern recognition and neural networks [57], but these studies are at the moment almost 20 years old, and they do not seem to readily fit into more modern algorithms based on deep learning, such as convolutional neural networks. Therefore, we decided to carry out such analyzes in the scope of this work.

The following two concepts would be useful to better understand the voting algorithms used in our research.

- **Condorcet winner** is the candidate who would like to win the other candidates. A voting system satisfies the Condorcet criterion if it always chooses the Condorcet winner when one exists. Of the voting methods discussed in this section, plurality voting and the Borda method do not meet the Condorcet criterion. Figure 2.9 shows the scheme to choose the winner in the Copeland method and the ideas of the Condorcet winner.

- **preference matrix** $A$ represents voters' preferences between each pair of candidates. Each of its elements $A_{(i,j)}$ represents the number of voters who preferred more candidate $i$ than the candidate $j$, reduced by the number of voters who preferred more candidate $i$ than candidate $j$. Candidate $i$ is preferred when the majority of voters are in a higher position in the preference ranking. In the situation where the candidate $i$ is Condorcet (candidate that wins against each of the other candidates in a direct election), all elements in the $i - th$ row of the matrix are non-negative.

### 2.2.2 Voting schemes selected

Below we provided a brief description of the voting algorithms, which were used for ensemble decision-fusion.

- **Plurality voting** [58] (also called majority voting) is a system in which each voter can vote for one single candidate. The voter cannot divide his vote into more than one candidate, and the candidate who receives the largest number of votes wins the elections. With reference to the preferences list of each voter, we can define a voting vector of weights, which determines part of the vote to pass to candidates at particular places in the list of preferences. In the case of plurality voting, the preference vector has the following structure: $[1, 0, 0, \cdots, 0, 0]$, which means that the candidate in the first place of the preferential list gets $100\%$ of the voter's voting power.

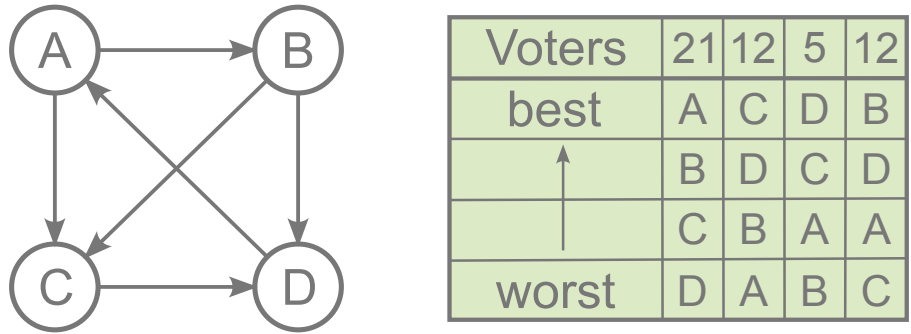

**Figure 2.9:** The Copeland's method and the idea of the Condorcet winner. Graph representing four candidates $A$, $B$, $C$ and $D$. The table on the right presents voters' preferences. For instance 26 voters prefer $A$ more than $B$ (21 + 5) and $B$ is more preferred than $A$ by 24 voters (12 + 12), So $B$ overcomes $A$. By repeating this step for all pairs, we determine the final result of the voting. In this case, the most points are gained by candidates $C$ and $D$. In the presented example there is no Condorcet winner, but after removing candidate $D$, the Condorcet winner would become candidate $C$.

- **Borda count** [59, 60] is a family of single-winner election methods in which voters rank options or candidates in preference order. Candidates may receive points from the voter even if they are not first on the voter preferences list. The method consists of awarding points that represent the preferences of all voters for each candidate. The final ranking results are the sum of points each candidate has received. The one who scores the most points is the winner. This method prefers options that are sufficiently accepted by the majority of voters, as opposed to methods that result in the selection of the option considered the best by the largest group of voters. Therefore, it is described as a consensus-based method and not as a majority method. The number of points assigned to a particular item in the preferences list must form a nongrowing sequence; individual variants of the method differ by precise scoring values.

- **k-Borda** is the most popular variant of the Borda count. The number of points allocated to the first item on the list corresponds to the number of all options and decreases by one each for the next option up to one point for the least preferred option on the list. Referring to the weight vector, this method is represented by the following vector: $[n, n - 1, n - 2, \cdots, 3, 2, 1]$, which are characterized by a better coverage of the electoral preferences of all voters. Figure 2.10 shows a schematic voting situation when, for the same electoral preferences, another candidate wins the voting, depending on the voting algorithm used, the plurality voting and the k-Borda method were compared. The weight vector of the vote for the plurality voting and for the k-Borda method is shown in Figure 2.11.

- **Dowdall system** is another known variant of the Borda method [61]. Voter awards the first-ranked candidate with 1 point, while the second-ranked candidate receives half of a point, the third-ranked candidate receives one-third of a point, et cetera. This variant is used in the electoral system of the third-smallest country in the world, island nation of Nauru [62].

- **Simpson-Kramer method** [63, 64], also known as the successive reversal method or the minimax method. Select the candidate $i$ for whom the highest preference matrix score for another candidate $(j, i)$ is the lowest such score among all candidates. Select the candidate $i$ with the highest minimum value in row $i$ in the preference matrix.

- **Single transferable vote** [65, 66] is a preferential voting algorithm. Each voter presents his list of preferences, in which they can include any number of candidates. Specifying preferences for all candidates is not obligatory. A certain threshold value is set that defines the minimum necessary score. In a situation where none of the candidates has exceeded the electoral threshold, the weakest candidate is removed from the subsequent procedure and the voting procedure is repeated. Votes assigned to the weakest candidate are divided into the remaining candidates, in order of preference of voters. This procedure to reject the weakest candidates is repeated until the winner or the winners of the elections are selected. This method can also be used in multi-mandate voting.

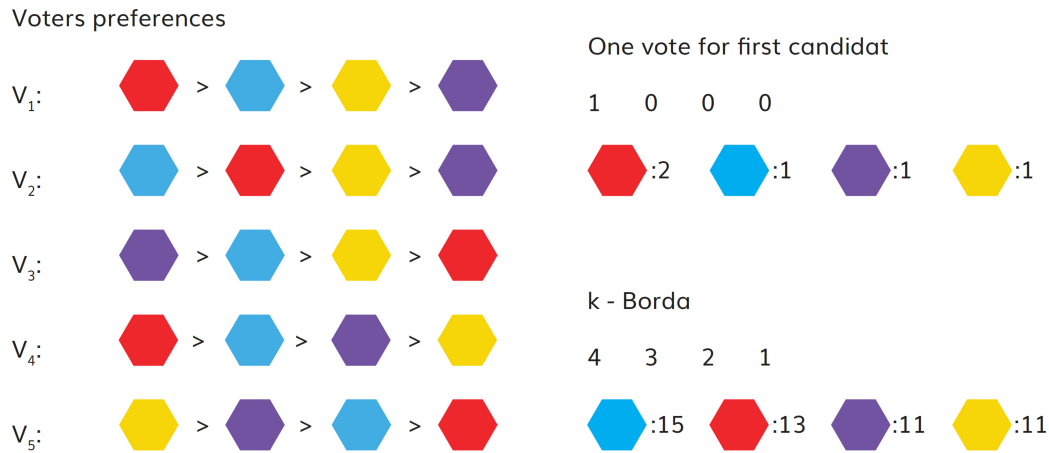

**Figure 2.10:** For the same preferences of the voters, different results may be obtained depending on the voting method used (distribution of the vector of votes).

- **Copeland's method** [67] In this method, a graph is created that defines the electoral preferences. The vertices are the candidates, and the edges, more precisely, their direction, determine which of the two candidates is preferred by the majority. The candidate evaluation is defined as ordered by the number of pairwise victories minus the number of pairwise defeats.

Figure 2.12 shows the distribution of the voting winners. Voters and candidates were randomly chosen over a uniformly distributed 2D space. They set their preferences and choose candidates. Euclidean distance was used as a preference measure. The closer the candidate is to the voter, the higher his position on the list of preferences. Then, the voting algorithms presented above were used to select the voting winners. Although the distribution of voters (and candidates) comes from a uniform distribution, the generated distribution of voting winners is characterized by disturbances specific to a given voting method [68, 69]. Some voting methods reward centrist candidates because they may be relatively high on the preference lists of voters from a large part of the sampling space.

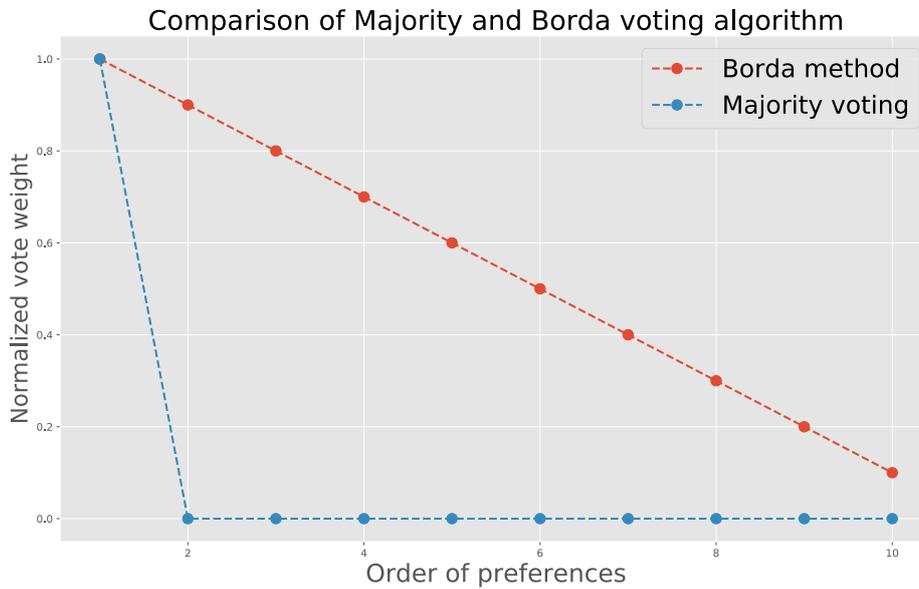

**Figure 2.11:** Comparison of the weight vector of votes for the plurality voting and the Borda method. In the case of plurality voting, only the candidate in the first place in the preferential list receives points. In the Borda method, these votes are distributed among all candidates in accordance with the above pattern.

## 2.3 Cyclical learning rate

Another area of analysis, in addition to the impact of the selected decision-fusion mechanism, was research on accelerating the training of ensemble models. The concept of cyclically modifying the learning rate assumes the possibility of capturing several models during one single training process. There are several techniques focused on models generated by cyclically modifying the learning rate [70, 71], which causes the model weights to settle at various local minima of the loss function. Cyclical methods reduce the time it takes to train the entire ensemble. Submodels come from one training process, they are checkpointed from different epochs (saving the values of trainable weights at a certain point of training), and therefore their weight space differs. Figure 2.15 schematically shows the process of generating checkpoints and the epochs in which they are saved. The ensemble still causes a linear increase in the inference time. In the following section, we describe selected methods for generating checkpoints.

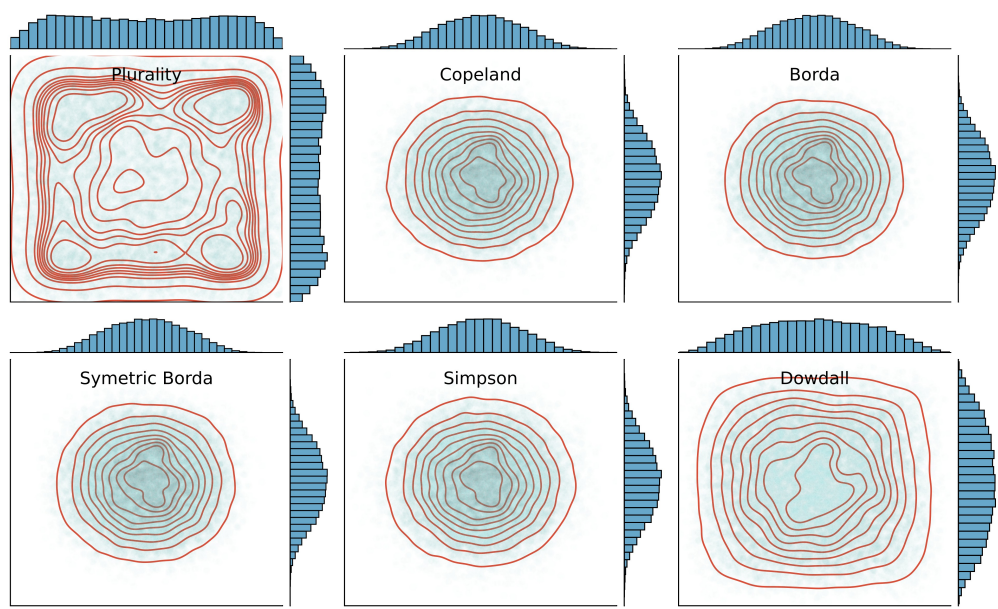

**Figure 2.12:** Distribution of the location of the voting wieners in 2D space, obtained from voting among voters coming from the uniform distribution. Voters and candidates were randomly chosen over a uniformly distributed 2D space. They set their preferences and choose candidates. Euclidean distance was used as a preference measure. The closer the candidate is to the voter, the higher his position on the list of preferences. Although the distribution of voters (and candidates) comes from a uniform distribution, the generated distribution of candidates who won voting is characterized by disturbances specific to a given voting method.

### 2.3.1 Snapshot Ensembles

The first technique we analyzed was the Snapshot Ensembles [72] that use a cosine-cyclical learning rate[73] to obtain multiple model checkpoints during training, in epochs when the learning rate is lowest. The weight states recorded in these epochs are submodels of the ensemble. The motivation behind the methods with a cyclical change in learning rate during the training process draws attention to the fact that the behavior of SGD [74, 75] and other similar optimizers [76–78] is strongly dependent on the learning rate factor and can fall into local minima. Increasing the learning rate leads to a better exploration of loss surface, forcing hopping from one local minimum to other. The combination of several different minima should be more robust and accurate. This method allows one to obtain multiple neural networks which can be combined at no additional training cost. This is achieved by letting a single neural network converge into several local minima along its optimization path and save the model parameters at certain epochs, and therefore the model weights are *snapshots*. Figure 2.13 presents a high-level overview of this method. Rapid convergence repetition is achieved using the shifted cosine function [79] as the learning rate schedule. It can be described as follows:

$$\alpha\left(t\right) = \frac{\alpha_0}{2}\left(cos\left(\frac{\pi\,mod\,(t-1,\,[T/M])}{[T/M]}\right) + 1\right) \qquad (2.5)$$

where $\alpha_0$ is the initial learning rate, $t$ is the number of iterations, $T$ is the total number of training iterations, and $M$ is the number of cycles. In other words, the training process was split into $M$ cycles, each of which starts with a large initial learning rate, which is annealed to a smaller learning rate. From its initial value $\alpha_0$ to $f(T/M) \approx 0$ over the course of a cycle. The higher learning rate $\alpha_0 = f(0)$ provides the model enough energy to escape from a critical point, while the small learning rate $\alpha_0 = f(dT/Me)$ drives the model to a well-behaved local minimum. Figure 2.14 compares the training process using traditional and cyclic learning rate schedules.

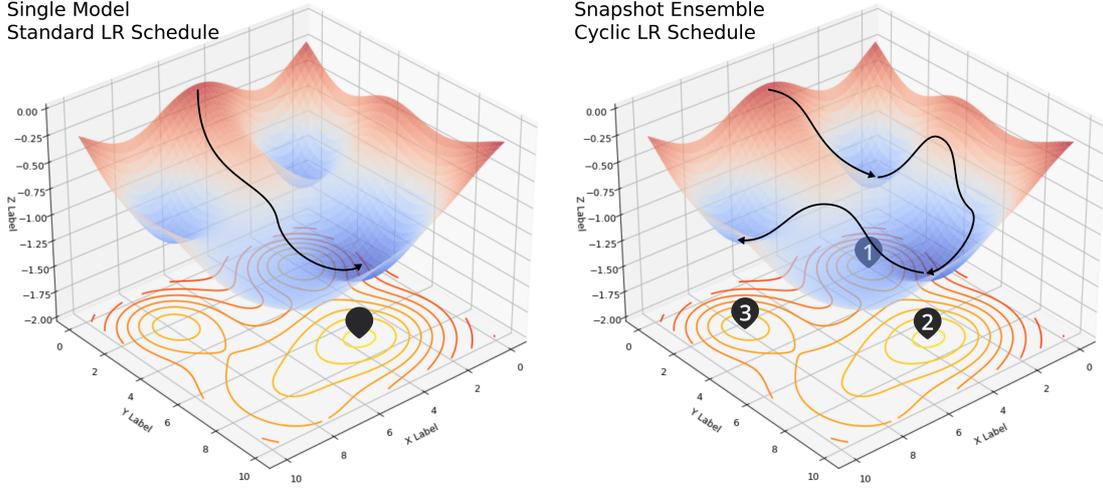

**Figure 2.13: Left**: Illustration of SGD optimization with a typical learning rate schedule. The model converges to a minimum at the end of training. **Right**: Illustration of Snapshot Ensembling. The model undergoes several learning rate annealing cycles, converging to and escaping from multiple local minima. Snapshots are taken at each epoch with the lowest learning rate.

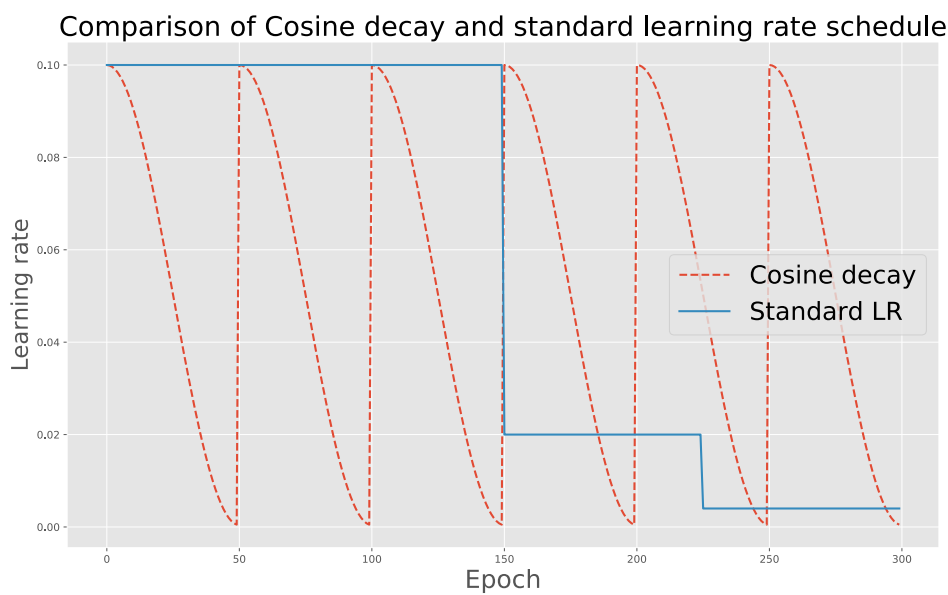

**Figure 2.14:** The Snapshot Ensemble and standard learning rate schedule.

### 2.3.2 Fast Geometric Ensembling

The second method analyzed is the Fast Geometric Ensembling (FGE) [80] whose authors have found that the optima of these complex loss functions are, in fact, connected by simple curves over which training and test accuracy are nearly constant. Based on this observation, the FGE for the first $70\% - 80\%$ of the training budget sets the standard learning rate similarly to the independent training model. In the second phase, cycles of linear increase and decrease in the learning rate are introduced from large $\alpha_1$ to small $\alpha_2$ with an equal cycle $C$, as shown in Figure 2.15. For each cycle, a checkpoint is created when the learning rate is the lowest. Depending on the architecture used, the authors recommend small cycles of lengths of two to four epochs. By cyclically modifying the learning rate, the model weights settle at various local minima. The cyclical methods reduce the time it takes to train the submodels for the ensemble but still cause a linear increase in ensemble inference time. Cyclical learning techniques generates dozens of checkpoints, especially in case of the FGE method.

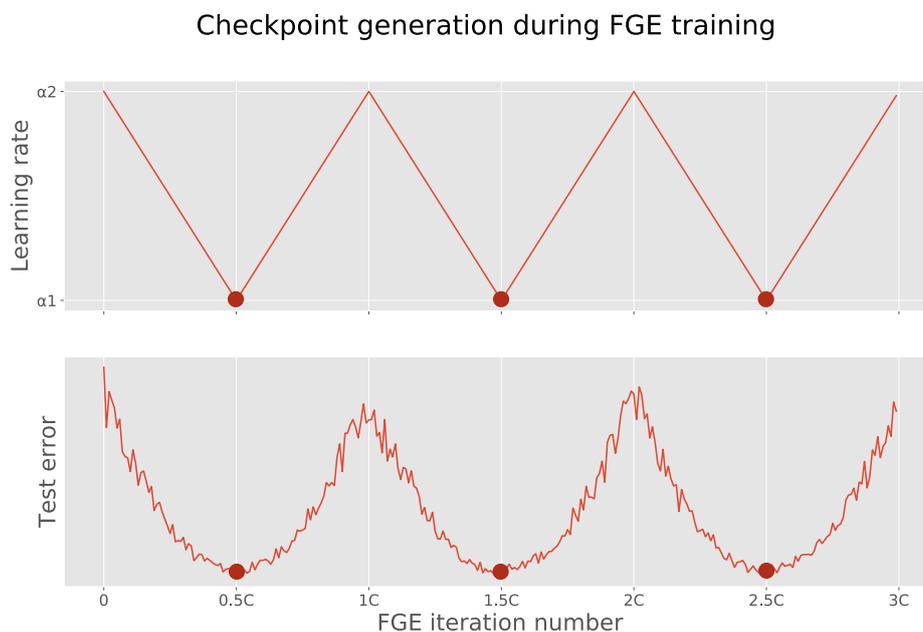

**Figure 2.15:** Schematic drawing of the second phase of the FGE method when we begin to linearly modify the learning rate. Learning rate **(Top)** and test error **(Bottom)** as a function of iteration for Fast Geometrical Ensembling. Models are saved for ensembling in points: $0.5C$, $1.5C$, and $2.5C$ when the learning rate and test error are the smallest (checkpoints are marked with red dots). $C$ denotes the number of epochs equal to the period of the learning rate function.

## 2.4  Knowledge distillation

### 2.4.1  Model compression

To develop modern machine learning models, recent work generally focuses on the development of new efficient building blocks [81–84] but there are also strong research branches that focus on model compression and acceleration techniques [85]. We can divide these methods into the following categories [86]:

- **Parameters pruning and sharing** methods explore redundancy in the model parameters and try to remove redundant and non-critical ones. These methods focus on removing inessential parameters from deep neural networks with no significant effect on performance. This category is further divided into model quantization [87], model binarization [88, 89], structural matrices [90], and parameter sharing [91, 92]. The authors of the Lottery

Ticket Hypothesis [93], make interesting observations. The standard pruning technique naturally uncovers subnetworks whose initializations made them capable of training effectively, and it is possible to train in isolation selected subnetworks that reach test accuracy comparable to the original network in a similar number of iterations.

- **Low-rank factorization** techniques use matrix decomposition to estimate the informative parameters of deep convolutional neural networks [94, 95]. These methods identify redundant parameters of deep neural networks that employ matrix and tensor decomposition [96, 97].

- **Transferred/compact convolutional filters** approaches design special structural convolutional filters to reduce the complexity of storage and computation. These methods remove inessential parameters by transferring or compressing convolutional filters [98].

- **Knowledge distillation** methods learn a distilled model and train a more compact neural network to reproduce the output of a larger network. These methods distill the knowledge from a larger deep neural network into a smaller network [99, 100]. The next section describes this approach in more detail.

## 2.4.2 Knowledge distillation

In the knowledge distillation technique, a small *student* model is generally supervised by a large *teacher* model. The main idea behind knowledge distillation is that the simpler *student* model mimics the complex *teacher* model, resulting in competitive or even superior performance. In this way, the knowledge inscribed in the weights of the *teacher* model is compressed and transferred to the weight space of the *student* model. Knowledge distillation can be thought of as one of the compression methods. However, this framework is much more general and can be widely used in other applications [101–110]. The versatility resulting from the concept of knowledge distillation is primarily due to the lack of requirements for the types of *teacher* and *student* models. This technique is most often used to compress machine learning models based on neural networks whose architectures are very similar, differing only in the number of hidden layers and neurons in each layer. However, there are no formal requirements on the type of machine learning model used. It is possible to distill the knowledge between models of completely different structures, types, and principles of operation. It is sufficient to ensure that both models have the same output and input structure. This diversity was used in the interpretability of neural networks, with the use of knowledge distillation, a decision tree was trained where *teacher* was a convolutional network [111], which allowed insight into the decision-making mechanism of the neural network itself.

In the standard training process of a classifier in supervised learning, the loss function is closely related to the data labels. In the case of knowledge distillation, the loss function has a second component related to the distance between the outputs (or/and hidden states) of the *teacher* and *student* models. More details and variants of knowledge distillation are presented in the survey paper by Gou et al. [112]. In the knowledge distillation framework, we can distinguish three main concepts: knowledge, distillation strategies, and the architecture of *teacher* and *student* [100]. A basic framework uses the logits of a large deep model as knowledge of *teacher* [99, 113]. The activations and characteristics of the hidden layers can also be used as a source of knowledge to guide the *student* model [114, 115]. Relationships between different neurons or activations [116] or parameters of the *teacher* model [117] also contain knowledge and can be used in the distillation process. In general, knowledge can be separated into three different categories [100]: response-based knowledge, feature-based knowledge, and relation-based knowledge. Figure 2.16 shows different types of *teacher* knowledge.

In our research, we focus on response-based knowledge. This type of knowledge seems to be the most general; this makes it possible to treat the *teacher* and *student* models as a black box. We can use any different statistical models, the only formal condition being that they must have the same input and output shape. The loss determines how well the *student's* output mimics the *teacher's* output conditioned by the same input. Figure 2.17 shows the general scheme of response-based knowledge.

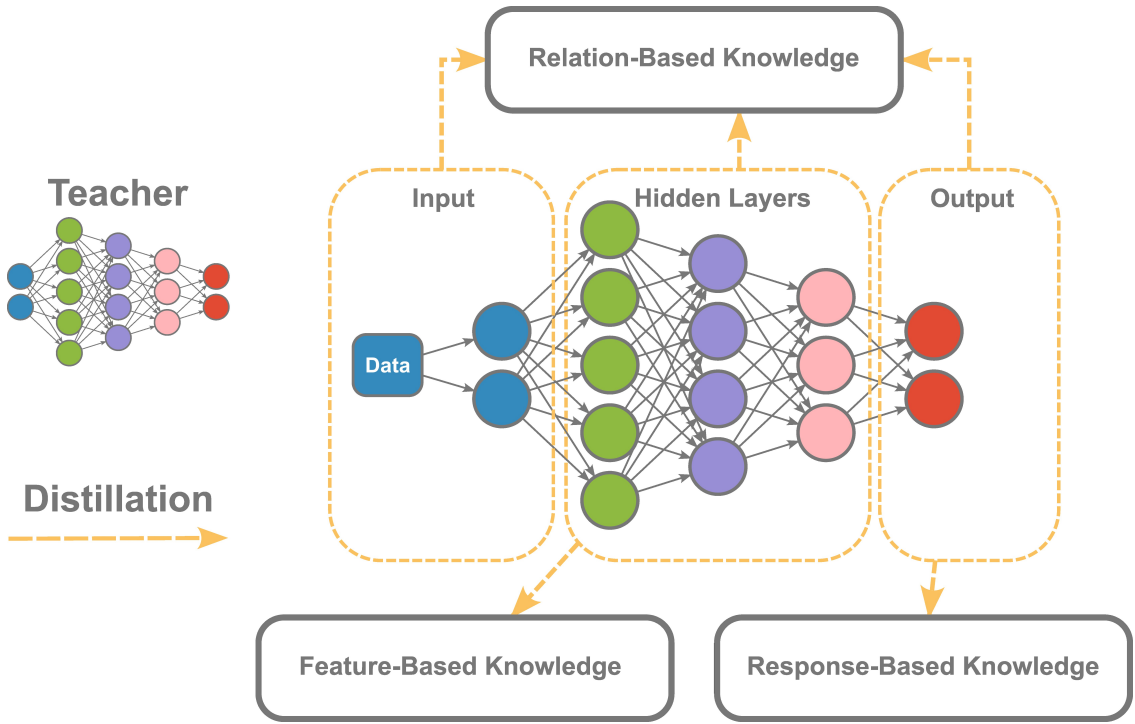

**Figure 2.16:** The schematic illustrations of sources of response-based knowledge, feature-based knowledge and relation-based knowledge in a deep *teacher* network.

This flexibility opens many possibilities for applications. For example, freedom to choose the *student's* architecture allows performing a hyperparameter search [118] to find the optimal architecture and meta-parameters for a given problem [119]. Thanks to this approach, we can find a *student* that is not just a simple slimmed version of the *teacher* (fewer neurons, fewer layers); it may be a completely different architecture that turned out to optimize the knowledge distillation process. Knowledge distillation can also be combined with quantization or compression. The *student* model can be compressed, quantized, or binarized [120–122], and the distillation of knowledge can also use the quantization-aware training strategy [123].

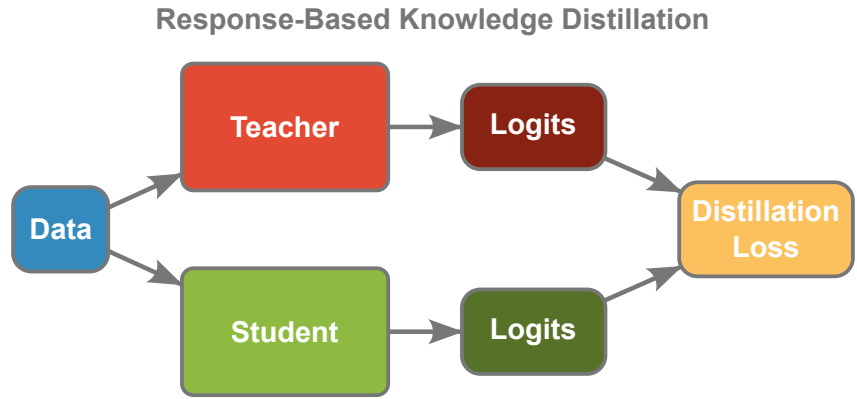

**Figure 2.17:** The generic response-based knowledge distillation.

## 2.5 Multi-teacher knowledge distillation

The versatility of knowledge distillation not only allows the freedom to create a *student* model, as shown in the previous section, but also gives great flexibility in defining the *teacher* model. In the particular case, the ensemble model acts as *teacher*. *Teacher* models can be used individually and integrally for distillation during the training period of a *student* network. To transfer knowledge from *Teachers*, the simplest way is to use the average response of all *teachers* as the supervision signal [99]. Several *multi − teacher* knowledge distillation methods have recently been proposed [124–131]. A generic framework for *multi − teacher* knowledge distillation is shown in Figure 2.18.

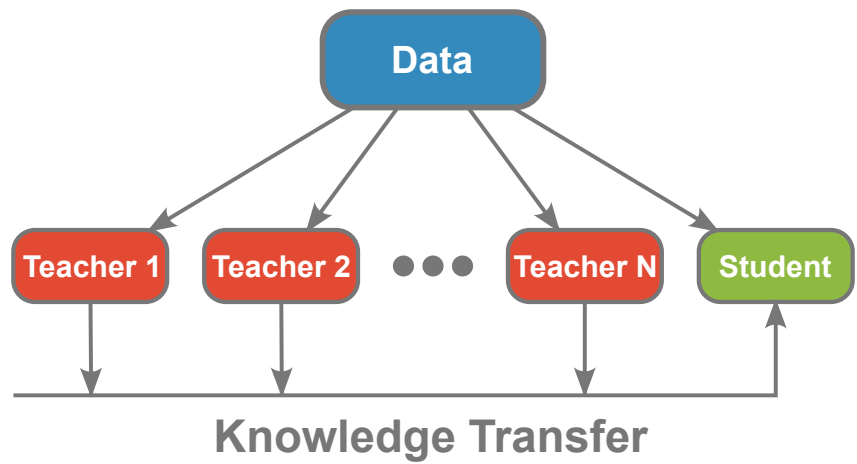

**Figure 2.18:** The generic framework for *multi − teacher* knowledge distillation.

As in the case of general ensemble learning, the aspect of decision-fusion is marginalized in the case of *multi − teacher* knowledge distillation. Averaging the output signal of the *teachers'* models makes the knowledge that can be transferred to the *student* model shallower. Instead of averaging, we have developed a technique that forces the *student* model to simultaneously mimic all *teacher* models, increasing its robustness and generality.

### 2.5.1   Framework

In our approach, response-based knowledge was used, so we assumed a lack of formal restrictions on knowledge distillation (comparing hidden layer activation requires consistency between the *teacher* and *student* architectures). We treat the entire ensemble as the *teacher* model. Each subnet *teacher* was trained on unique and randomly generated subsets of the training data set in bagging manner. The procedure to generate the subsets is shown in Figure 2.19. As a result, we can use knowledge distillation as an ensemble decision-fusion scheme. The *student* model learns to mimic the predictions of the entire ensemble of the *teacher* submodels.

There have been several studies that utilize knowledge distillation as ensemble aggregation, as mentioned in the previous section. However, there are some important differences between these approaches. The major modification assumes transfer of the decision-fusion part (about the aggregation of individual sub-models) from the stage before knowledge distillation to the *student* model itself. To the best of our knowledge, this is the first work that incorporates this concept. The main task of the *student* model is not to imitate some aggregation of *teachers'* outputs, e.g. by averaging them, but all individual *teachers'* outputs. We also decided not to include in the loss function the factor that represents the similarity between *teacher* and *student* in hidden layers. Our loss function forces the *student* model to mimic only the output of *teachers'*. This approach is more flexible in terms of knowledge distillation between models of a different type (a similar architecture is not required).

The softmax function converts a vector of $K$ real numbers into a probability distribution of $K$ possible outcomes. Softmax is the most used activation function located in the last layer in the case of a classification problem. It converts logits into classification likelihood. The softmax function is given by the following equation:

$$P_i = \frac{e^{y_i}}{\sum_{k=1}^{K} e^{y_k}} \tag{2.6}$$

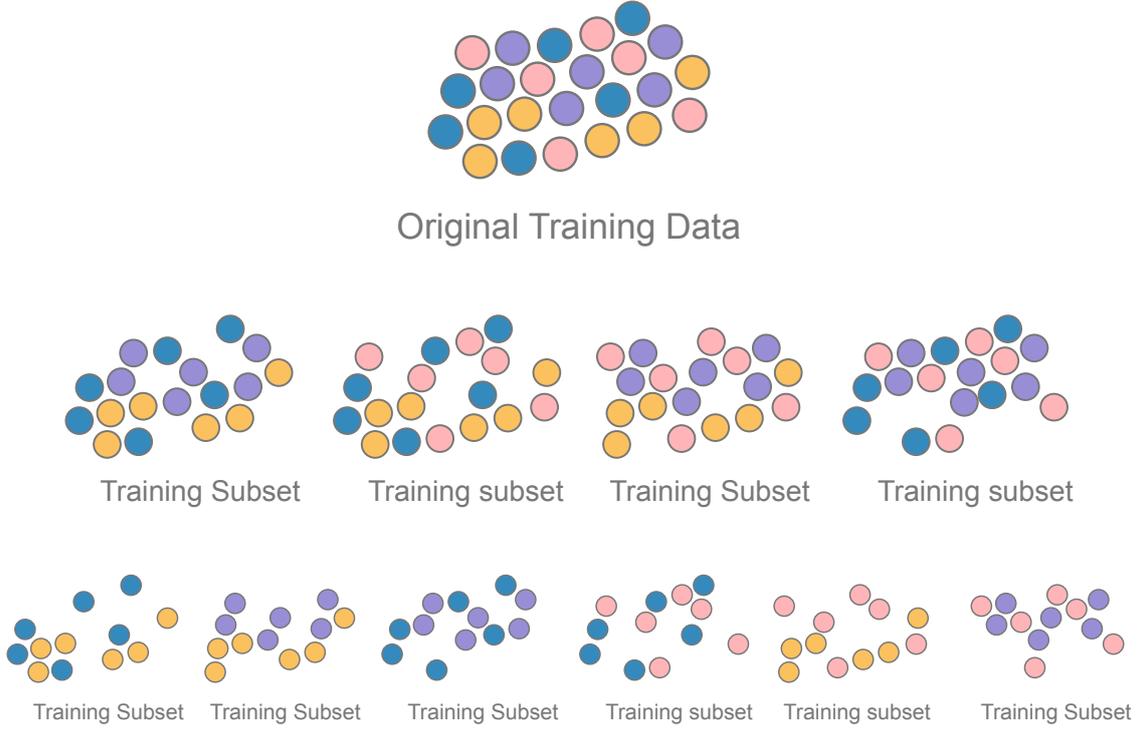

**Figure 2.19:** Procedure to generate training samples for each *teacher*. The top row symbolically represents the complete training dataset. 4 training subsets containing 75% of the training samples are shown in the middle row. In the bottom row, we generate 6 various subsets containing 50% of the training samples.

Where $P_i$ denotes the probability of belonging to the $i-th$ class and $y_i$ denotes the output value (before activation) corresponding to the $i-th$ class. In knowledge distillation, the *teacher* outputs obtained using the softmax activation function are a quantity that the *student* model tries to imitate by minimizing the Kullback-Leibel divergence [132] between *teacher* and *student* predictions. However, an additional temperature parameter is introduced, the purpose of which is to introduce an additional regularization mechanism. The softmax function extended by the temperature parameter $T$ is as follows:

$$P_i = \frac{e^{y_i/T}}{\sum_{k=1}^{K} e^{y_k/T}} \tag{2.7}$$

When the parameter $T$ is greater than 1, which is standard practice in knowledge distillation, the probabilities are softened. That is, the highest probability value in the returned probability vector decreases and the lowest values increase. In our $multi-teacher$ approach, we have decided to leave the unchanged parameter $T$ equal to 1. This decision was made intentionally. although label smoothing leads to better results in terms of knowledge distillation, however, in the case of learning from many $teachers$, just imitating many outputs at the same time has a regularizing effect.

We analyzed three variants of the $multi-teacher\ single-student$ architecture, with different $teachers$ mimicking schemes.

1. **Prediction averaging** - Currently used approach [133] consists in averaging the ensemble predictions before the $teacher$ output is included in the $student's$ loss function. The $student$ model learns to mimic the average response of the $multi-teacher$ ensemble (Figure 2.20 upper).

2. **Mimic of prediction geometric center** - In the training process, the output of the $student$ model is compared with the predictions of all $N\ teachers$ individually. The $student$ model learns to mimic the predictions of several $teachers$ simultaneously. However, since bringing the prediction too closely to a single $teacher$ output increases part of the loss function responsible for mimicking other $teachers$, the output of the $student$ model settles in the geometric center of all the $teachers'$ predictions (Fig. 2.20 center).

3. **Independent mimicking of all the teachers** - In contrast to the method presented above, the $student$ model does not produce a single output, but $N$ outputs, where $N$ is equal to the number of $teachers$, each is characterized by an independent set of trainable weights, see Fig. 2.20 lower. The last layer or the last few layers may be separated. Each of these independent outputs in the training process is compared with its assigned *teacher's* output. Note that the convolution part responsible for feature extraction is shared. However, the weights of the last layer (or the last few layers) responsible for the classifications are specific to each *teacher*. In this way, the model does not learn to mimic the aggregation of all *teacher's* outputs, but rather generates $N$ independent predictions linked to each *teacher*. The total loss function is made up of the ground-truth related loss part and the sum of loss related to mimicking each *teacher*. In the inference phase, making a prediction consists of aggregating all the outputs of the *student* model, by averaging softmax probabilities. The *student* model has the last additional layer that averages independent $N$ outputs that represent a single *teacher*.

In our opinion, which is part of the hypothesis being verified, averaging *teachers'* responses leads to the loss of some information. Therefore, methods that allow the *student* model to access all *teachers* outputs should lead to better results. As shown in [133] there are many definitions of loss functions, different distance matrices, and distillation strategies. We decided to use hard *ground true* labels and hard (meaning that the temperature in formula 2.7 is set to 1, the *teachers'* output probabilities are not smoothed; instead, we obtain a similar regularization effect by analyzing the probabilities of several *teachers* simultaneously), rather than light labels (in which the labels do not have the entire probability assigned to one class, but it is partially "fuzzified" to other classes). We also used the Kullback-Leibler divergence to determine the distance between the $teacher-$ and $student-model$ outputs. Below we present the respective equations that determine the *student* loss function for different variants of mimicking described above.

$$Loss_{avg} = \alpha \sum_{i=1}^{D} \bar{y}_i \cdot log(\frac{\bar{y}_i}{\tilde{y}_i}) \ - \ (1-\alpha) \cdot \sum_{i=1}^{D} y_i log(\tilde{y}_i) \qquad (2.8)$$

$$Loss_{geo} = \alpha \frac{1}{N} \sum_{i=1}^{D} \sum_{j=1}^{N} y_{ij} \cdot log(\frac{y_{ij}}{\tilde{y}_i}) \ - \ (1-\alpha) \cdot \sum_{i=1}^{D} y_i log(\tilde{y}_i) \qquad (2.9)$$

$$Loss_{ind} = \frac{1}{N} \sum_{j=1}^{N} \cdot (\alpha \sum_{i=1}^{D} y_{ij} \cdot log(\frac{y_{ij}}{\tilde{y_{ij}}}) - (1-\alpha) \cdot \sum_{i=1}^{D} y_i log(\tilde{y_{ij}})) \quad \text{(2.10)}$$

Where $D$ is the *student* output size (number of classes), $N$ is the number of *teachers*, $\tilde{y_i}$ is the $i - th$ scalar value in the *student* model output, $y_i$ is the corresponding target value, $\bar{y_i}$ is the corresponding average of the *teachers* model output, $y_{ij}$ is the corresponding $j-th\ teacher$ output, and $\tilde{y_{ij}}$ is the $i$-th scalar value in $j - th$ output of *student* model output (independent mimicking variant). The $\alpha$ parameter set proportion between the expression associated with knowledge distillation (first sum in the equations) and the standard loss function connected to the data ground truth (second sum) is the process controlling parameter. Increasing this parameter increases the *student* imitation loss part in the total loss function. Figure 2.20 demonstrates the block diagram of all the $multi-teacher$ $single-student$ networks described above in the order presented in the text.

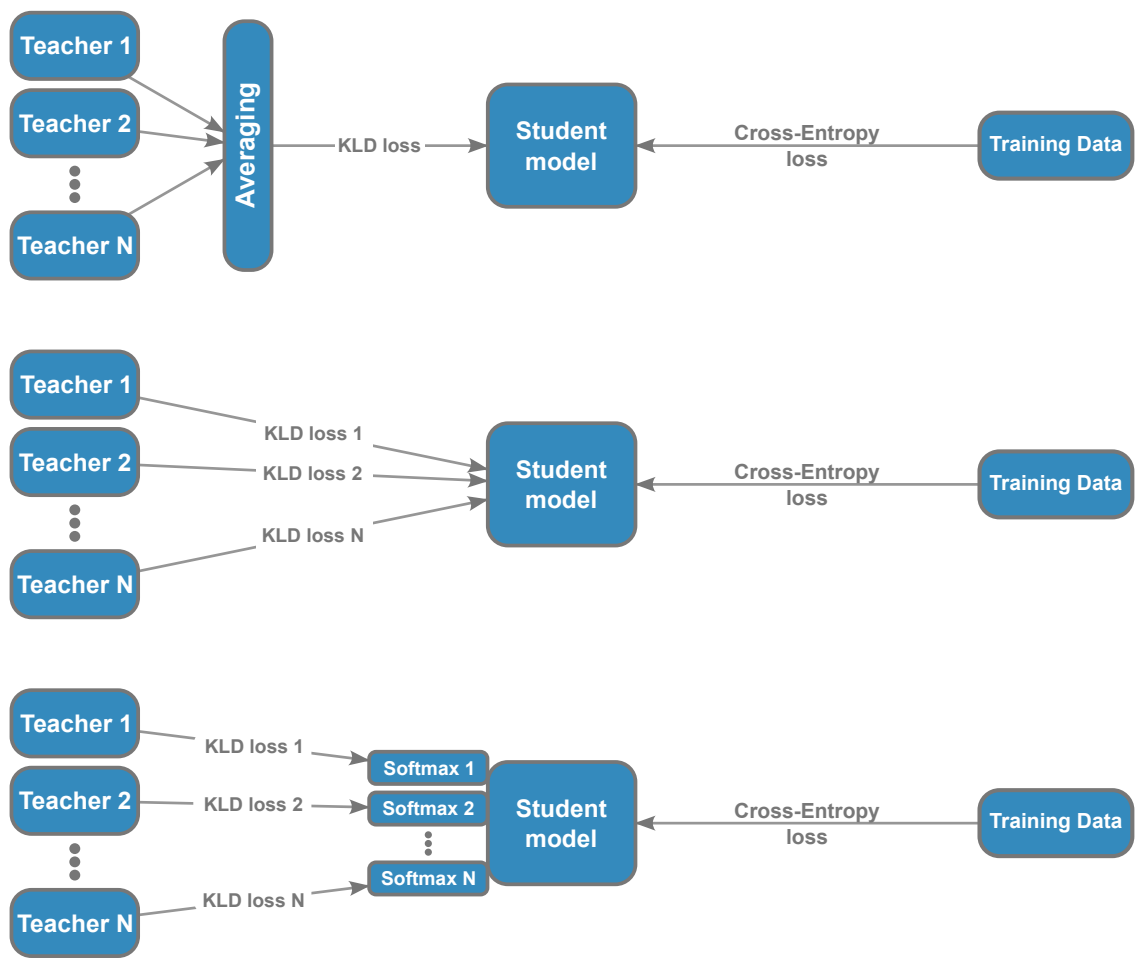

**Figure 2.20:** The schemes of the *multi − teacher single − student* models employed in this paper: *prediction averaging* model (**top**), the model *mimic of prediction geometric center* (**center**), *independent mimicking of all the N teachers* (**bottom**).

<div style="text-align: right; font-size: 3em;">3</div>

# Results

## Contents

## 3.1  Hardware and software setup

The computations were performed on the Prometheus supercomputer (475th on the top 500 list (June 2022) [134]; HP Apollo 8000, Xeon E5-2680v3 12C 2.5GHz, Infiniband FDR, HPE Cyfronet Poland). We used one node (Intel Xeon E5-2680 v3, 2.5 GHz) with two Nvidia V100 GPU accelerators in the cluster dedicated to deep learning. In the computation, we used the TensorFlow framework [135] with Keras [136]. Furthermore, we used the pandas [137] and NumPy [138] libraries for data processing. For efficient parallel data processing and computing in Python, we used the Dask library [139]. Matplotlib[140] and seaborn [141] packages were used for visualization[1].

---

[1]Code to reproduce reserch results is available at `https://github.com/ZuchniakK/MTKD`, also `https://github.com/ZuchniakK/VotingAlgorithms` contains an implementation of the voting algorithms used in my research along with examples of use.

## 3.2 Voting in ensembles

To investigate the influence of voting methods (see Section 2.2) on decision-fusion in ensemble models, we trained 3000 simple fully connected deep neural networks containing only two hidden layers with 50 neurons each. MNIST dataset [142] was used due to the low complexity of these data and, consequently, the fast training time, which is crucial for such a large number of individual classifiers that were trained. The MNIST dataset contains $60,000$ training samples and $10,000$ test samples. To make the models sufficiently diverse, each of them was trained on a randomly generated set of samples from the original training set. The models were intentionally trained with a limited time budget so that the weights did not converge. *Batch size* was set at 100 examples and 100 iterations of batch training were performed. Finally, each of the models, during the short training phase, analyzed $10,000$ training images. This training procedure lasted only one-sixth of an epoch (in relation to the full training set), which is a very short period. We used Adam [76] optimizer with *learning rate* $= 0.001$, $\beta_1 = 0.9$, $\beta_2 = 0.999$, and $\epsilon = 1e - 07$. We deliberately wanted to obtain poorly trained classifiers to emphasize the benefits of ensemble learning when several poor-quality models work together. To this end, we have randomly chosen $N$ $(2 - 60)$ among the trained models and formed an ensemble classifier. This random selection was performed many times for each $N$ to obtain statistical distributions of the ensemble classification accuracy. We measured the classification accuracy of the ensemble classifier using several voting methods described previously as a decision-fusion mechanism of individual submodels. In Figure 3.1 we present the classification accuracy histograms based on the number of basic models in the ensemble system. In this case, the decision-making process is based on the basic plurality voting algorithm. When the size of the ensemble increases, the classification accuracy increases, but it also comes with a linear increase in computational cost. However, we can improve the classification accuracy without changing computational costs by using different decision-fusion schemes (voting algorithms). Figure 3.2 presents the histograms of the classification accuracy obtained by the ensemble classifiers built from the same number of submodels. The only modification was to replace plurality voting with the Borda method. Without modifying computational costs, the quality of the classification has improved noticeably. In the presented case, the size of the ensemble group is 15, however, for each $N$ tested in the range of 2 to 60, an improvement in classification quality was observed.

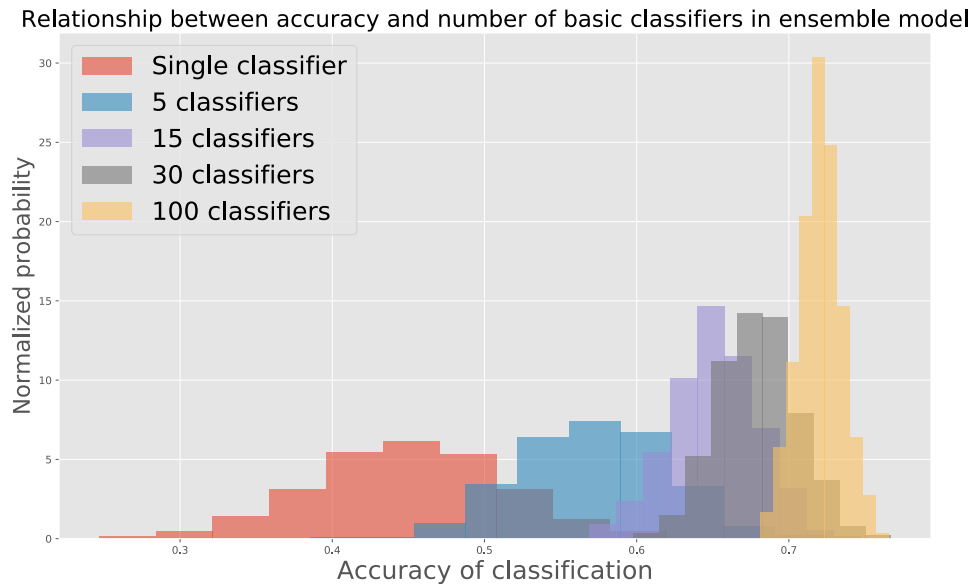

**Figure 3.1:** Accuracy of classification on MNIST dataset depending on the number of basic models in the ensemble system. Ensemble submodels are simple fully-connected deep neural networks containing only two hidden layers with 50 neurons. Each model was trained on a small fraction of the training dataset.

As shown in Figure 3.3, various voting schemes produce different dependencies between classification accuracy and the size of the ensemble. Plurality voting gives the worst results among all the methods tested. The single transferable vote and Copeland's method give similar and slightly better results. The similarity of the two obtained results comes from the similar algorithm structure of these methods. Both, in contrast to the Borda method, meet the Condorcet criterion. Borda method generates the highest quality of classification. Compared to the summing weights of the last layer of the neural network (softmax layer), which has much more complete information on the predictions of the submodels (complete vector containing the probabilities of belonging to each class vs. preferences list), the performance of the Borda method is comparable to the summation of softmax layers for larger values of $N$ ($> 20$). However, for smaller values of $N$, summing the probabilities (from the softmax layer) is the most efficient method. This is mainly due to the fact that the preference list loses some of the information returned by individual submodels. Table 3.1 contains several sample values of the classification accuracy for various voting algorithms.

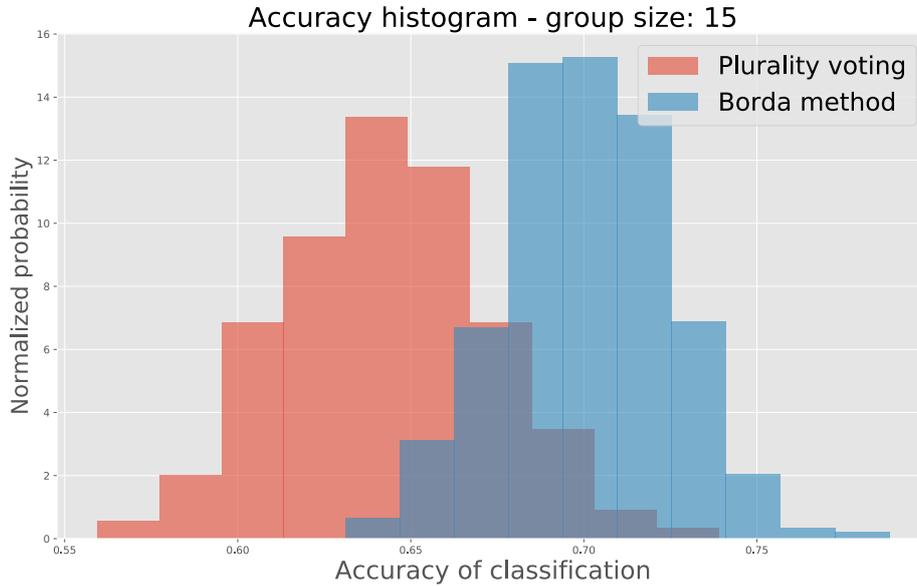

**Figure 3.2:** Accuracy of classification depending on the used voting algorithm, ensemble group size is 15. Ensemble submodels are simple fully-connected deep neural networks containing only two hidden layers with 50 neurons. Each model was trained on a small fraction of the training dataset.

| Ensemble size | Plurality | Borda | STV | Copeland's | Softmax |
|---|---|---|---|---|---|
| 5 | 58.6% | 61.1% | 58.3% | 60.0% | 62.7% |
| 25 | 66.1% | 69.8% | 67.8% | 68.1% | 69.7% |
| 55 | 67.7% | 71.5% | 69.6% | 69.9% | 71.1% |

**Table 3.1:** Classification accuracy depends on the size of the ensemble model and the voting algorithm used.

The results presented in this chapter are behind the SOTA results. Note that MNIST is a very simple dataset, its current SOTA is $99.91\%$ classification accuracy [143]. The analyzes performed were rather for modeling purposes. We have shown the impact of applying ensembling that if we have a group of classifiers of known effectiveness and there is non-zero variance between them, we can obtain an ensemble model that will perform better than any of its submodels. The second fundamental observation shows that by changing the decision-fusion algorithm, which has a marginal impact on the level of computational complexity, we can also improve the performance of the ensemble model.

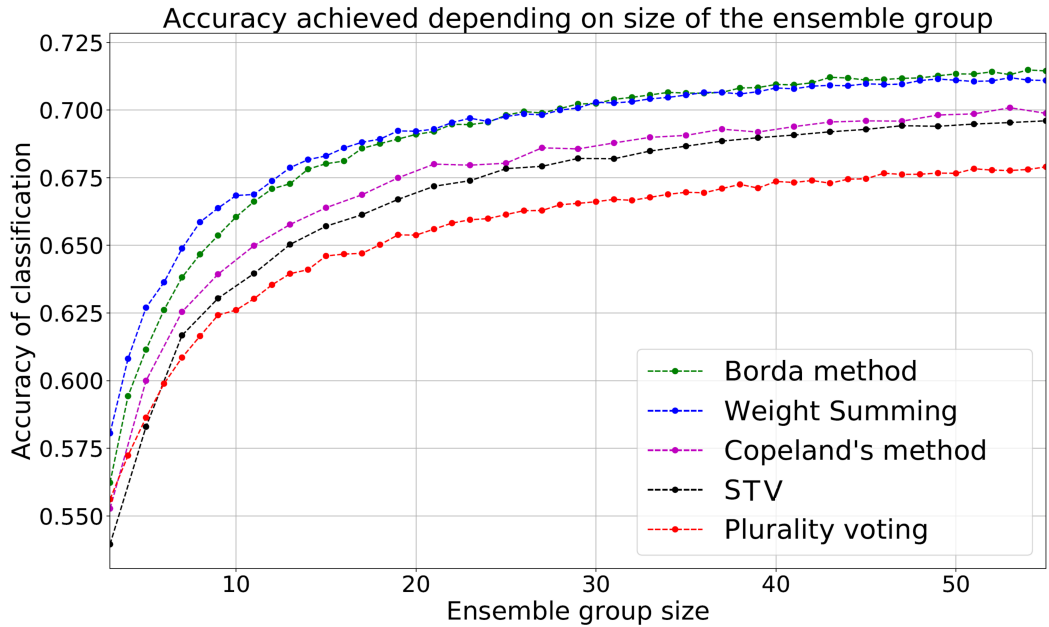

**Figure 3.3:** Dependence between classification accuracy and size of ensemble groups for different decision-fusion methods. Ensemble submodels are simple fully-connected deep neural networks containing only two hidden layers with 50 neurons. Each model was trained on a small fraction of the training dataset.

## 3.3 Cyclical learning rate

Here, we conduct experiments on ensembles with submodels that were trained with a cyclic learning rate schedule; see Section 2.3 During this research, an additional element was the verification of cyclical learning rate methods, allowing the acquisition of multiple weights checkpoints during a single training process. We train DenseNet [144] model on the CIFAR10 [145] dataset in three different ways: using the standard learning rate schedule, using the Snapshot Ensemble method and the Fast Geometric Ensemble method. Furthermore, we use the advanced AutoAugment [146] data augmentation technique. The following analysis focuses more on a qualitative explanation of the consequences arising from the use of cyclical learning rate methods.

For the baseline training, we used Adam optimizer with $learning\ rate = 0.0001$, $\beta_1 = 0.9$, $\beta_2 = 0.999$, and $\epsilon = 1e-07$, and $batch\ size = 32$. For Snapshot Ensemble we also used the same $batch\ size$ and Adam optimizer with the same parameters but with different learning rate during training, $\alpha_0$ was set to $0.1$. In the case of Fast Geometric Ensembling, $\alpha_1$ was set to $0.01$ and $\alpha_2$ was set to $0.0005$. As shown in Figure 3.4, an ensemble built from independently trained models gives better results but requires much more computational time in the training process. The results of the snapshot ensemble are very similar, regardless of the voting algorithm used. The analyzes presented here were obtained on the CIFAR10 dataset and the DenseNet121 neural network architecture. Furthermore, the performance of a single model trained independently for 300 epochs (the red dot in Figure 3.4) is comparable to the ensemble of all snapshots. The effectiveness of the ensemble learning procedure depends mainly on two factors. The first is the quality of individual models, while the second is the diversity between submodels. If we concatenate a set of submodels trained on the same training dataset, with the same overfitting and bias, the ensemble model that will combine them will also have these disadvantages. Therefore, it is important for individual models to be varied; in this comparison, cyclically generated models perform worse compared to the ensemble for which the submodels were trained independently.

Figure 3.5 shows a similarity matrix of the classification output for the same input data for independently trained models, Snapshot Ensemble, and Fast Geometric Ensembling methods. Pairwise similarity comparison illustrates how often the classifications of two models converged. It is obvious that the high

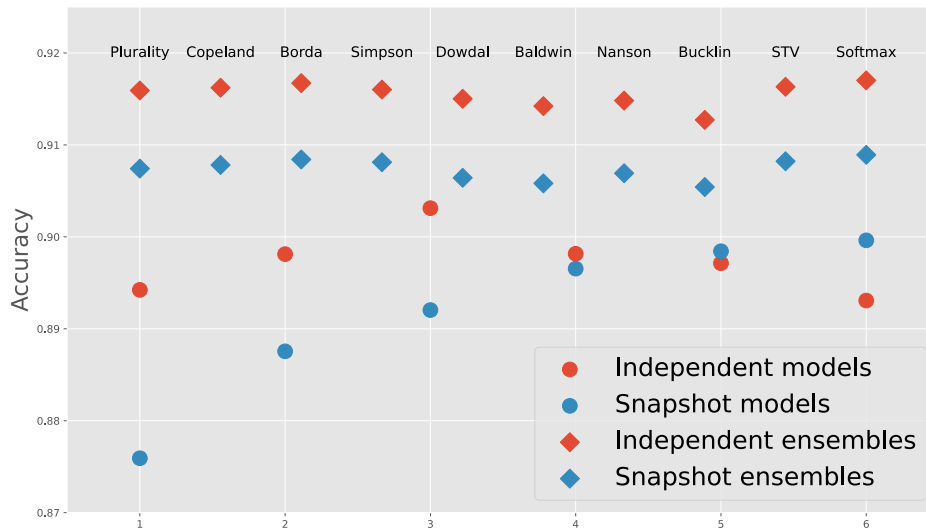

**Figure 3.4:** Blue dots represent accuracy obtained by DenseNet121 model on CIFAR10 dataset trained with cyclical learning rate in local minimum phase (weights are saved on epochs 50, 100, 150, 200, 250, 300 - when the learning rate was smallest). Blue diamonds represent the accuracy of snapshot ensemble possessed by different voting methods. Red dots represent 6 independently trained models, and their accuracy after finishing full training process according to the standard learning rate schedule (epoch 300). Red diamonds represent the accuracy of the independent ensemble possessing by different voting methods, it required 6 times more computational time (it is linearly dependent on the number of models, while cyclic methods, like snapshot ensemble, generate $N$ models during one training process)

prediction accuracy leads to high pairwise similarity (two classifiers that always indicate the correct class will give identical predictions). However, cyclically generated models also tend to increase this similarity. In other words, cyclically generated models are less varied than models trained independently. We can see that the checkpoints recorded in the 50 and 300 epochs when training with the Snapshot Ensemble method differ more than the average difference of the two independently trained models. However, this is because the 50-epoch checkpoint is a classifier with low classification accuracy.

To better analyze and illustrate how model weights evolve during Snapshot Ensemble training in each epoch (300 data points), we checked the output value for all $10,000$ testing examples. Then, using t-SNE [147, 148], we visualize the output of the model in each training epoch, as shown in Figure 3.6). The same analysis was carried out for models with a standard learning rate schedule, as shown in Figure 3.7.

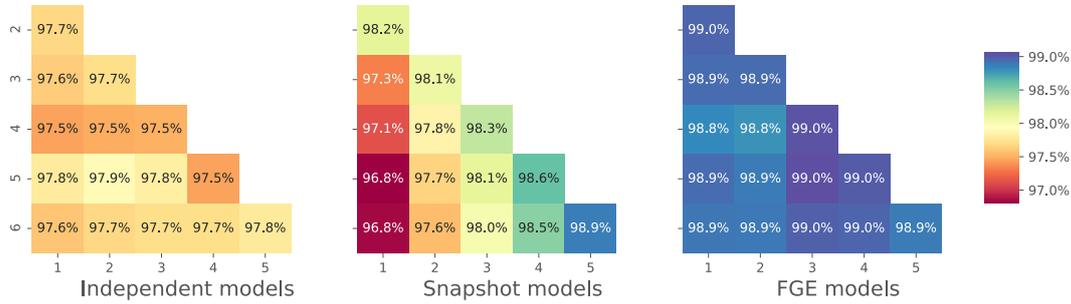

**Figure 3.5:** Similarity matrix of the classification output for the same input data for independently trained **(left)**, Snapshot Ensemble **(center)**, and Fast Geometric Ensembling **(right)** methods. Presented data are obtained on training DenseNet121 neural network architecture on CIFAR10 dataset.

It also provides interesting conclusions. The locations of individual epochs have much smoother transitions. Throughout the training period, a relatively uniform change in the embedding position can be observed. A similar analysis was also performed for the model trained using FGE. However, due to its specificity, the visualization does not differ much from the t-SNE for the baseline learning rate schedule. The first phase of training, which includes the vast majority of epochs, runs the same way as in the case of baseline training. In the second phase of training, the embeddings of the last epochs from the phase of linear learning rate changes in the learning rate are located close to each other.

## 3.4 Knowledge distillation

The most important part of this dissertation (excluding the implementation aspect) is the use of knowledge distillation as a mechanism for decision-fusion and aggregating model predictions. This section presents the results of the knowledge distillation efficiency for the $multi-teachers \ single-student$ approach. Previously, we discussed the problems associated with the use of ensemble learning.

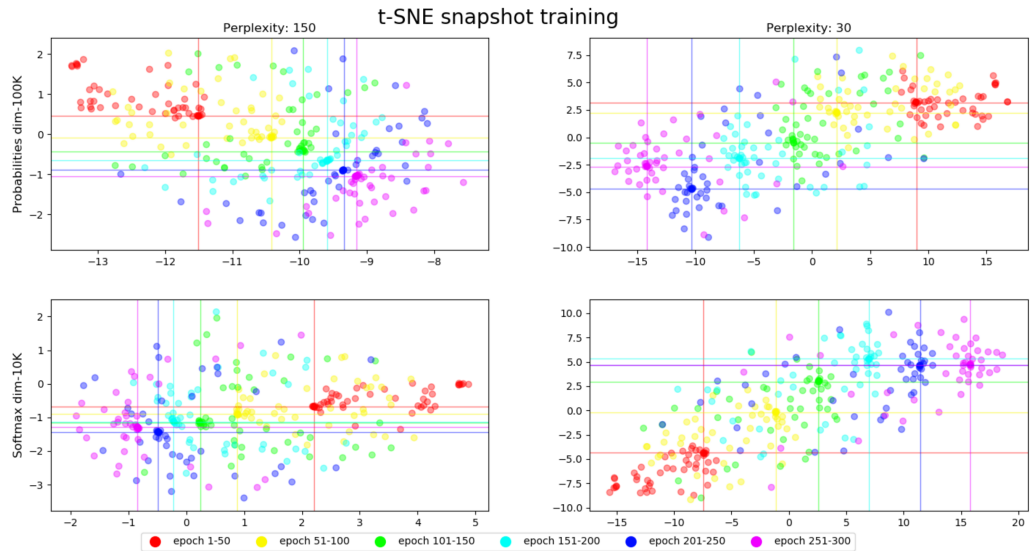

**Figure 3.6:** t-SNE for Snapshot Ensemble model output. Two output representation have been considered, in first variant **(upper)**, we used all returned probabilities ($dimension = number\ of\ examples * number\ of\ classes = 100.000$). In the second variant **(bottom)**, dimensionality was only represented by correct class probability. This method reduces output dimension to $10,000$. We used two values of the perplexity parameter, 30 **(right)** and 150 **(left)**. Perplexity a measure for information that is defined as 2 to the power of the Shannon entropy. Which says how to balance attention between local and global aspects of data. The parameter is, in a sense, a guess about the number of close neighbors each point has. However, the perplexity value has a complex effect on the resulting embedding and therefore the optimal value is often chosen empirically. It is clearly shown that that output converges around the local minimum. The lines intersect points corresponding to the epoch with the smallest learning rate ending each cycle. Presented data are obtained on training DenseNet121 neural network architecture on CIFAR10 dataset.

Cyclic learning rate methods partially solve these problems, but only in terms of reducing the time needed to train the entire ensemble model. The computational complexity of inference remains the same; it is the sum of the inference complexity of the submodels that make up the entire ensemble model. The problem is also the poor quality of the classifiers obtained in this way. Independently trained submodels can be differentiated in many ways: architectures can be modified, the dataset can be changed, or the training hyperparameters can be modified. Cyclical checkpoints do not provide such flexibility. However, knowledge distillation is an extremely general and flexible framework. As shown in previous chapters, there are not many formal constraints that define the characteristics of the *teacher* and *student* models. To obtain diverse *teachers*, we randomly generated subsets of the training dataset. The examples with probability $p$ were part of the training subset. Training the model on the full training set corresponds to a $p - value$ of 1. Decreasing $p - value$ leads to slightly weaker performance of individual models (they are trained on a smaller number of training examples). However, this increases the diversity between individual submodels. The number of training examples, which is not common to individual models, is increasing. A schematic diagram of the generation of training subsets for different $p - values$ was shown in the chapter 2, Figure 2.19. The $teachers'$ models were trained in the manner described above, on incomplete data sets, and constitute the basis for the evaluation of various knowledge distillation mechanisms. We wanted to examine four factors that influence the quality of the prediction of the $student$ model obtained.

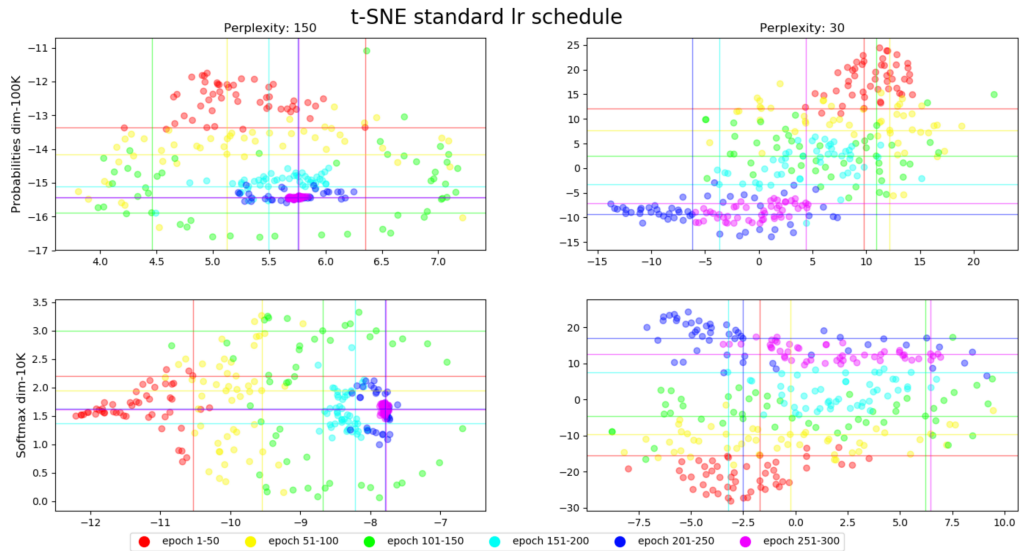

**Figure 3.7:** t-SNE for standard learning rate schedule. Two output representation have been considered, in first variant (**upper**), we used all returned probabilities ($dimension = number\ of\ examples * number\ of\ classes = 100.000$). In the second variant (**bottom**), dimensionality was only represented by correct class probability. This method reduces output dimension to $10,000$. We used two values of the perplexity parameter, 30 (**right**) and 150 (**left**). The lines intersect points corresponding to the epoch: 50, 100, 150, 200, 250, 300 (for analogy and easy comparison with the example from the Snapshot Ensemble t-SNE - Figure 3.6). Presented data are obtained on training DenseNet121 neural network architecture on CIFAR10 dataset.

- $p - value$ - Reducing the size of the training subset leads to a decrease in the performance of a single classifier; on the other hand, the differentiation between classifiers increases, which, in turn, should positively affect the performance of the ensemble. Therefore, it should also positively affect the performance of the $student$ obtained in the $multi-teacher$ knowledge distillation process. The purpose of the research was, inter alia, to determine where the optimal $p-value$ is located and what it depends on.

- **Type of mimicking mechanism used** - we tested three methods described in Chapter 2 to mimic the responses of the $teachers'$ models by the $student$ model. The baseline method is output averaging, which consists of averaging the probabilities of belonging to classes returned by each of the $teachers'$ models before sending them to the $student$ model in the knowledge distillation process. The next method is to mimic the geometric center of the $teachers'$ predictions, in the $student$ training process, the output of the $student$ model is compared to the predictions of all $N$ $teachers$ individually. The $student$ model learns to mimic the predictions of several $teachers$ simultaneously. The last method tested is an independent mimicking of all the $N$ $teachers$. In contrast to the method presented above, the $student$ model does not produce a single output, but $N$ outputs, where $N$ is equal to the number of $teachers$, each is characterized by an independent set of trainable weights. The last layer or the last few layers may be separated. Each of these independent outputs in the training process is compared with its assigned $teacher's$ output.

- *alpha* **parameter** - during the training of the $student$ model, we have two main components in the loss function. The first is the error associated with imitating the $teacher's$ (or $teachers'$) responses; the most commonly used is the Kulbacka-Leibel divergence. Another component is the error related to the prediction of ground truth. This contribution is identical to the basic training of the model without knowledge distillation machinery; most often in the case of classification, the cross-entropy loss is used. The parameter $alpha$ affects the proportion between these two components. An *alpha* equal to one is equivalent to standard training without knowledge distillation because the loss function is only related to real data. When this parameter is equal to $0$, the *student* loss function is entirely related to following the *teacher's* output. We conducted the distillation process for different values of the *alpha* parameter $(0, 0.25, 0.5, 0.75)$ to determine the impact of this value on the final quality of the *student's* classifier.

- **Number of** *teachers* - The number of submodels in a standard ensemble model increases the performance of the entire ensemble, leading also to a linear increase in computational complexity. In previous studies, we observed a saturating characteristic, that is, the curve representing the dependence of the classification accuracy on the number of submodels had a certain asymptote, which means that increasing the number of submodels at some point ceases to affect the performance of the team classifier. We wanted to observe this relationship in the case of $multi-teacher$ knowledge distillation.

The results presented here apply to the ResNet50 [4, 149] baseline model and the CIFAR10 dataset. Supplementary materials A include the results of experiments performed on the remaining datasets and other base models. For training *teachers* and *students*, we used the same setup: Adam optimizer with $learning\ rate = 0.0001$, $\beta_1 = 0.9$, $\beta_2 = 0.999$, and $\epsilon = 1e - 07$, and $batch\ size = 32$. We deliberately did not use data augmentation methods in this experiment. It worsened the results of the obtained classification accuracy. However, it allowed us to investigate the analyzed factors more precisely and to be closer to the simulated conditions of a highly limited data set. Table A.3 shows the accuracy of the *student* classification according to the knowledge distillation method used for different $p-values$. The *mimicking all* method gives the best results. Knowledge distillation generates a model that is slightly weaker than full ensembling with $N$ times larger computational overhead. Compared to a single model, we observed increases in classification accuracy. The following data are presented for the ResNet50 baseline model and the CIFAR10 dataset. Figure 3.8 shows the accuracy of the *student* classification, depending on the number of *teachers* used during the knowledge distillation process. Individual data series correspond to different methods of *teachers'* output mimicking by *student* model and different $p - value$. We observe the saturation characteristics. Adding more *teachers* increases the *student's* performance but to a lesser extent. Similarly, interesting observations can be obtained when we swap the $y$-axis and data series from the previous chart. Figure 3.9 shows the precision of the *student* classification, depending on $p - value$. This time, individual data series correspond to the number of *teachers* used during different methods of *teachers* mimicking the knowledge distillation process. In the case of the dataset analyzed and the neural network used, the best results are obtained using the complete training dataset ($p - value = 1$). This is because although the submodels trained on the full dataset will be more similar to each other, their accuracy is higher, which ultimately translates into a better-performing *student* model.

Table A.3 presents the aggregated values for all *alpha* parameters we tested. On the other hand, Table 3.3 presents a similar comparison, but with separate cases for different values of the *alpha* parameter. Due to the significant increase in the various values to be shown, the table 3.3 contains information only for the *mimick all* method and the *number of teachers* equal to 3. The observed relationship is parabolic. middle *alpha* values perform the best. Let us remind the reader that the value of the *alpha* parameter equal to 0 means that the *student's* model loss during training is related only to the *teacher's* imitation loss function. The loss factor related to the ground truth is zeroed and increases with the increase of the *alpha* parameter. When this value reaches 1, we have a case of classical training in which the factor of knowledge distillation and *teacher* imitation is zeroed. The table 3.3 shows an interesting relationship between the *alpha* parameter and $p - value$. The optimal value of *alpha* depends on $p - value$ and the optimal $p - value$ depends on the parameter *alpha*. For each column (*alpha* parameter const), we have marked the best classification accuracy in bold.

| Teachers number | Fraction | Single accuracy [%] | Ensemble accuracy [%] | Multi-Teacher KD accuracy [%] | | |
|---|---|---|---|---|---|---|
| | | | | *Output avg.* | *Loss avg.* | *Mimick all* |
| 2 | 1 | 82.18 (0.55) | 83.35 (0.47) | 81.92 (0.39) | 82.08 (0.54) | **82.72 (0.25)** |
| | 0.9 | 81.29 (0.55) | 82.67 (0.49) | 81.46 (0.41) | 82.07 (0.34) | **82.68 (0.38)** |
| | 0.8 | 80.52 (0.65) | 81.42 (0.52) | 81.55 (0.52) | 81.03 (0.62) | **82.09 (0.24)** |
| | 0.7 | 79.52 (0.53) | 80.43 (0.43) | **81.47 (0.29)** | 80.78 (0.34) | 81.28 (0.22) |
| | 0.6 | 78.17 (0.91) | 79.21 (0.66) | 79.25 (0.66) | 79.94 (0.66) | **81.1 (0.5)** |
| 3 | 1 | 82.18 (0.55) | 84.04 (0.43) | 81.97 (0.61) | 81.18 (0.89) | **83.11 (0.22)** |
| | 0.9 | 81.29 (0.55) | 83.14 (0.51) | 78.76 (0.46) | 81.89 (0.25) | **82.76 (0.31)** |
| | 0.8 | 80.52 (0.65) | 81.89 (0.47) | 81.53 (0.5) | 81.48 (0.4) | **83.08 (0.31)** |
| | 0.7 | 79.52 (0.53) | 80.93 (0.39) | 81.82 (0.28) | 81.64 (0.4) | **82.14 (0.47)** |
| | 0.6 | 78.17 (0.91) | 80.14 (0.53) | 80.31 (0.46) | 80.61 (0.64) | **81.2 (0.51)** |
| 4 | 1 | 82.18 (0.55) | 84.21 (0.35) | 80.77 (0.94) | 82.84 (0.36) | **83.22 (0.21)** |
| | 0.9 | 81.29 (0.55) | 83.78 (0.42) | 80.42 (0.6) | 81.92 (0.43) | **82.73 (0.19)** |
| | 0.8 | 80.52 (0.65) | 82.17 (0.38) | 82.09 (0.48) | 81.83 (0.39) | **82.48 (0.23)** |
| | 0.7 | 79.52 (0.53) | 81.24 (0.45) | 81.02 (0.73) | 81.44 (0.36) | **82.33 (0.25)** |
| | 0.6 | 78.17 (0.91) | 80.61 (0.44) | 80.59 (0.45) | 79.94 (1.05) | **81.68 (0.67)** |
| 5 | 1 | 82.18 (0.55) | 84.35 (0.31) | 83.39 (0.3) | 82.82 (0.45) | **83.66 (0.31)** |
| | 0.9 | 81.29 (0.55) | 84.42 (0.39) | 82.71 (0.48) | 82.17 (0.49) | **83.0 (0.32)** |
| | 0.8 | 80.52 (0.65) | 82.39 (0.41) | 82.31 (0.43) | 81.26 (0.48) | **83.55 (0.22)** |
| | 0.7 | 79.52 (0.53) | 81.52 (0.42) | 81.41 (0.78) | 82.01 (0.5) | **82.19 (0.49)** |
| | 0.6 | 78.17 (0.91) | 80.83 (0.38) | 81.28 (0.56) | 80.91 (0.66) | **81.86 (0.41)** |

**Table 3.2:** Classification accuracy comparison for a single model, ensemble model, and *student* model obtained using various *multi − teacher* knowledge distillation techniques of *teachers* mimicking. The following data are presented for the ResNet50 baseline model and the CIFAR10 dataset.

We can see that with the decrease of the $p − value$, the *alpha* parameter, giving the highest classification accuracy increases. The higher value of *alpha*, the closer to standard training (a minor contribution to the loss function comes from imitating *teachers*). When we have weaker *teachers*, access to the ground truth improves the performance of the *student* model. Similarly, with the decrease of the *alpha* parameter, the $p − value$ that gives the highest classification accuracy decreases. An interesting view on the influence of the *alpha* parameter can also be obtained from the presentation of these relationships in a graphic form. Figure 3.10 shows the dependence of the classification accuracy obtained on the *alpha* parameter used. Different data series mean that a different number of *teachers* is used in the training *student* model, the $p − value$ is set to 1. Figure 3.11 shows a twin relationship, but here the data series show different $p − values$, the number of *teachers* used for training is set to 3.

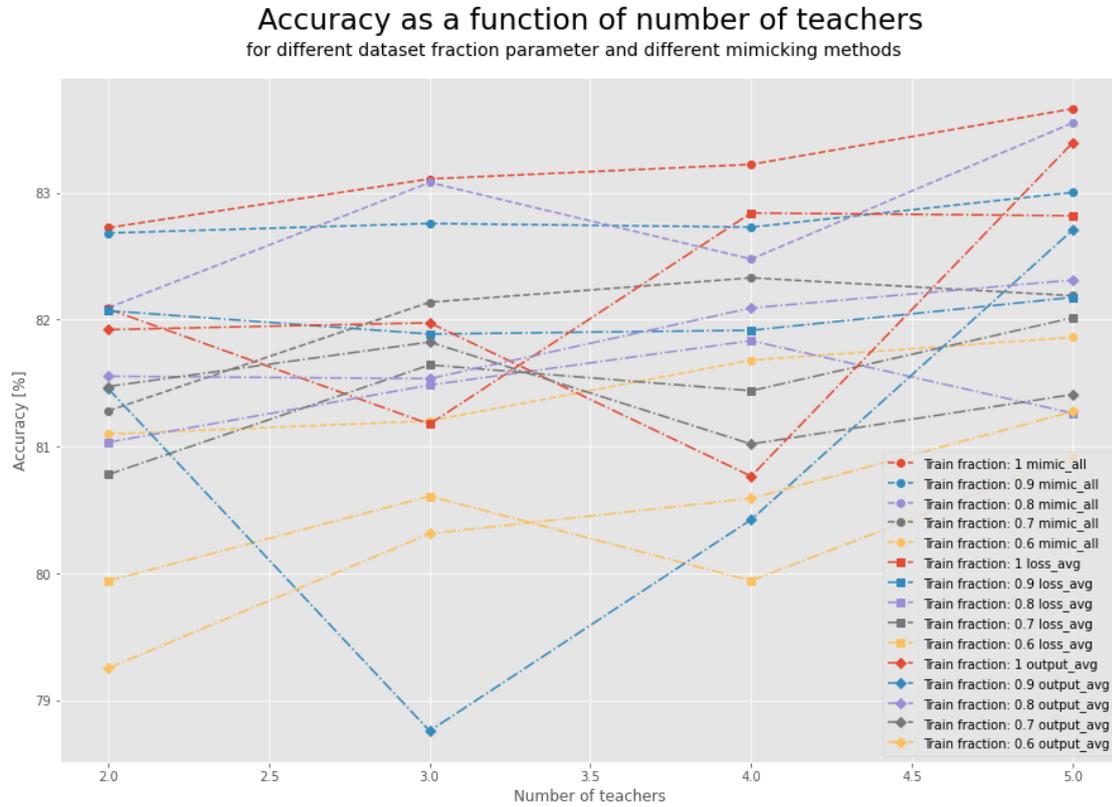

**Figure 3.8:** *Student* classification accuracy depending on *teachers'* number for different mimicking methods and $p - value$. The following data are presented for the ResNet50 baseline model and the CIFAR10 dataset.

| Fraction | *Alpha* parameter | | | |
|---|---|---|---|---|
| | 0 | 0.25 | 0.5 | 0.75 |
| | Accuracy [%] | | | |
| 1 | *83.16 (0.31)* | **83.87 (0.33)** | *83 (0.49)* | 82.4 (0.28) |
| 0.9 | 82.08 (0.52) | **83.95 (0.33)** | 83.33 (0.22) | 82.06 (0.53) |
| 0.8 | 82.93 (0.42) | **83.88 (0.07)** | *83.54 (0.64)* | 81.97 (0.19) |
| 0.7 | 80.66 (0.75) | 82.62 (0.84) | **83.09 (0.31)** | *82.92 (0.56)* |
| 0.6 | 80.26 (0.34) | 81.43 (0.27) | **82.14 (0.57)** | 81.92 (0.31) |

**Table 3.3:** Classification accuracy for *student* models obtained using various *alpha* and $p - value$ parameter. The number of *teachers* was set to 3 and *teachers'* mimicking method was *mimick all*. The following data are presented for the ResNet50 baseline model and the CIFAR10 dataset.

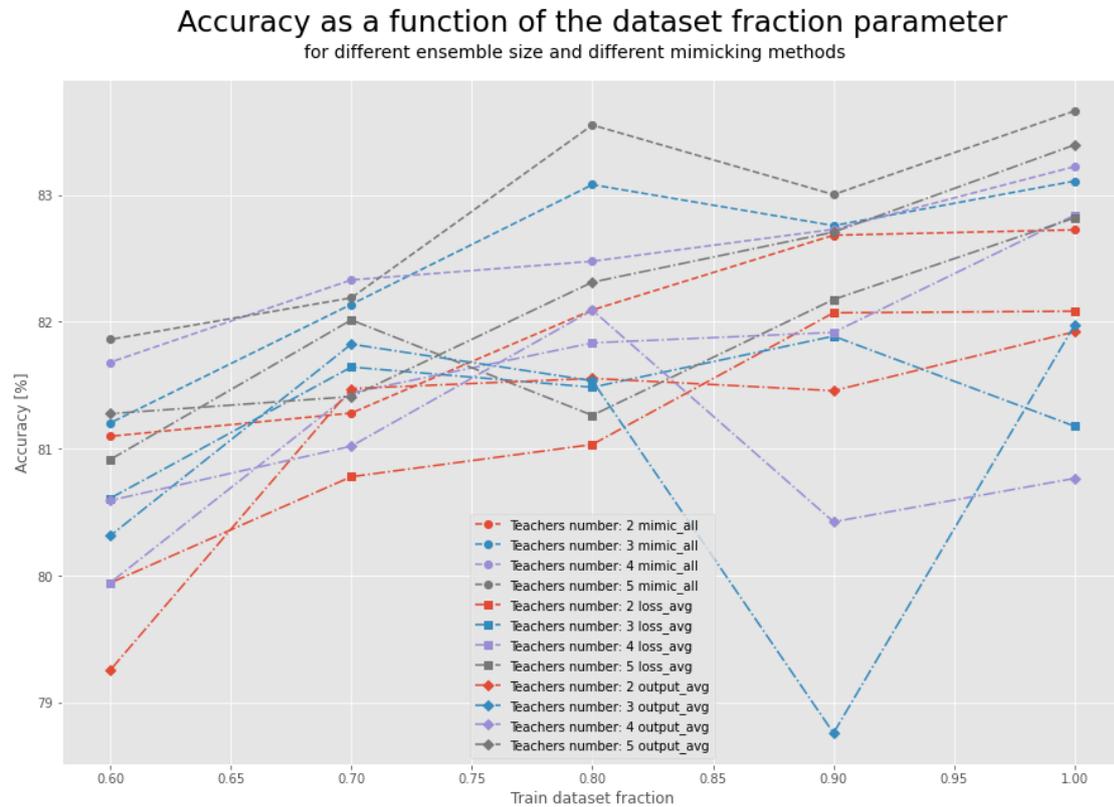

**Figure 3.9:** *Student* classification accuracy depending on $p-value$ for different *teachers* number and used mimicking method. The following data are presented for the ResNet50 baseline model and the CIFAR10 dataset.

## 3.5 Conclusions and future research

Comparing the efficiency that can be achieved by using the same submodels in the same ensemble, where the only variable is a different voting algorithm as a decision-fusion mechanism, gives us revealing insight into the voting algorithms themselves. The experiments presented above show a situation in which the ground truth is known. We know which classification is the correct one. In many situations where voting algorithms are used, this assumption is not met. We can measure the efficiency of the voting algorithms themselves as the ability to make the right decisions based on biased and uninformed voters' decisions. In the case of the above experiments, the voters were weak neural networks trained on the MNIST dataset. We have shown that there are voting methods that for the same data make better decisions compared to voting baseline, i.e. plurality voting.

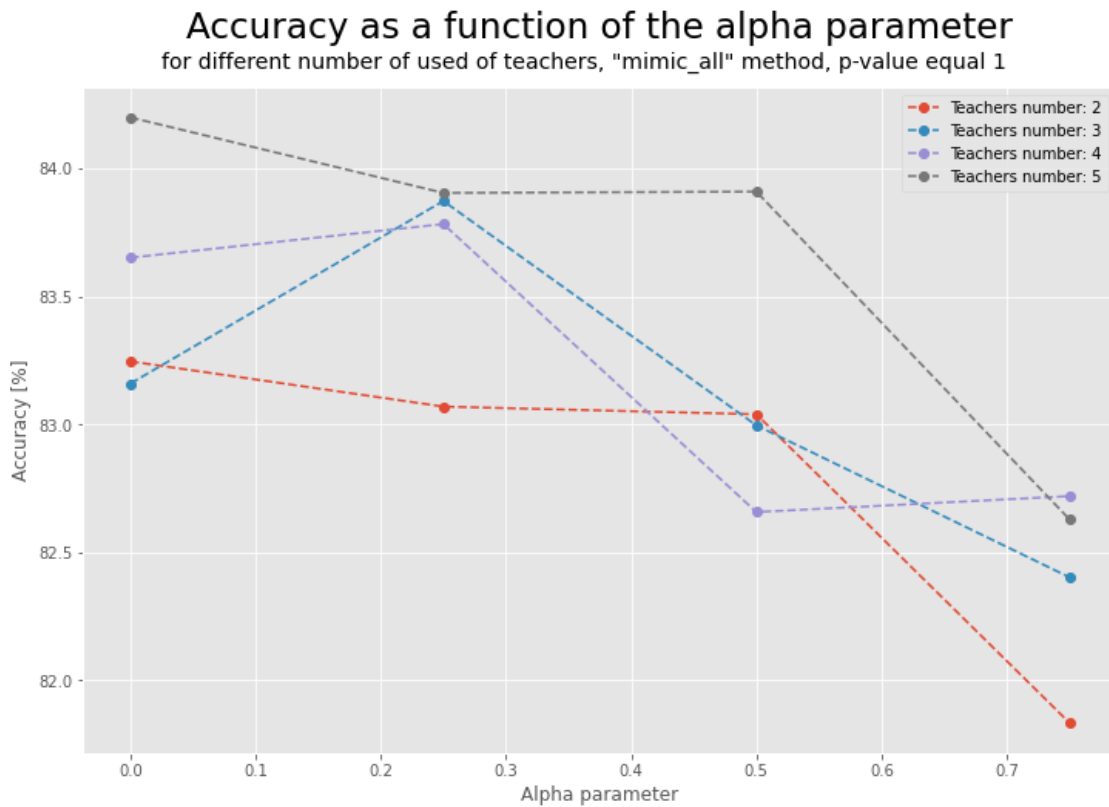

**Figure 3.10:** Obtained classification accuracy depending on the *alpha* parameter used, different data series for a different number of *teachers* used in training *student* model, the $p - value$ is set to 1. The following data are presented for the ResNet50 baseline model and the CIFAR10 dataset.

The main reason for this is that plurality voting limits data usage the most. Only the first item on the list of voting preferences is taken into account. However, the same problem undermines the legitimacy of using voting algorithms as a decision-fusion mechanism in ensemble learning. There are many applications where we have nothing more than a preference list. Then the use of voting methods seems to make the most sense. In the case of classifiers based on neural networks, at the output of a single submodel, we obtain a vector of probabilities belonging to a particular class. This information is much more detailed than just the preference list, which in this case determines the order of the sorted probability values without specifying the actual values. When deciding to use voting algorithms based on the lists of preferences, we lose some information, which leads to a worse effectiveness of the classification compared to the basic

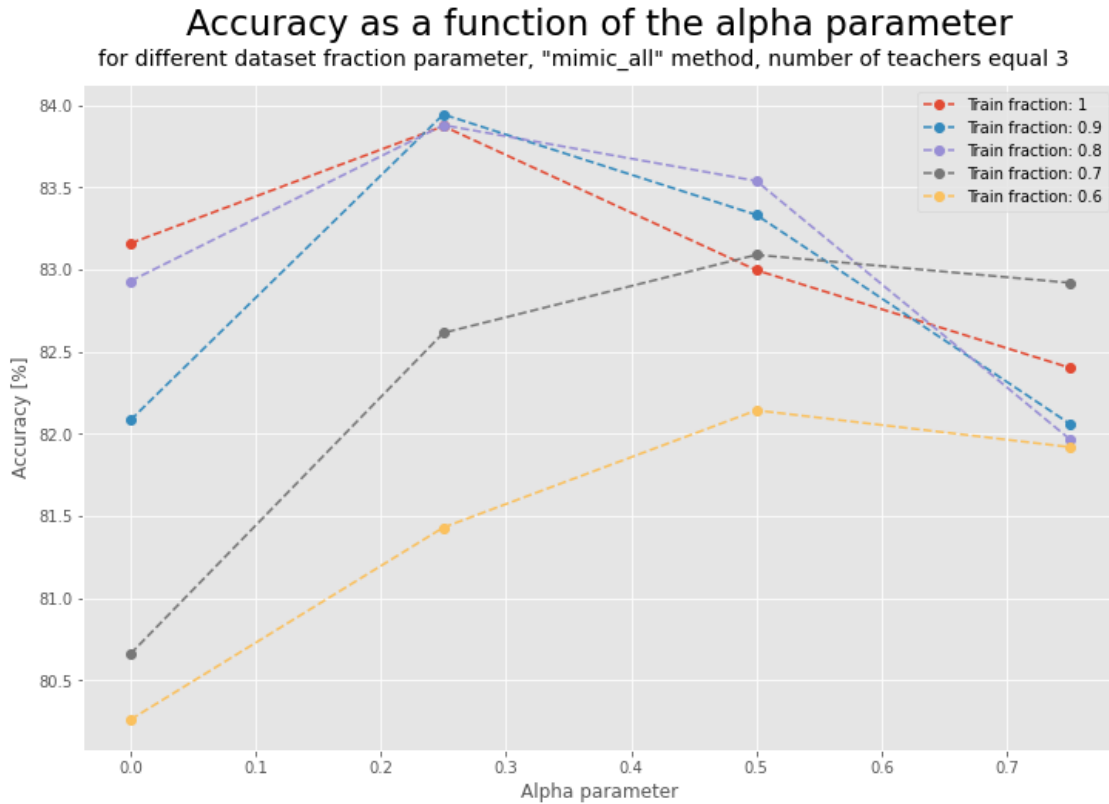

**Figure 3.11:** *Student* classification accuracy depending on *alpha* parameter used, different data series for different *p−value*, the number of *teachers* used for training *student* are set to 3. The following data are presented for the ResNet50 baseline model and the CIFAR10 dataset.

method of averaging probabilities (from the last softmax layers).

We described the concept of generating models using cyclical learning rate schedules. Then, we did a detailed analysis of the resulting checkpoints. We have shown that, in relation to a given input, the generated outputs are very similar. This leads to some issues in creating efficient ensembles from these models. The very high similarity between the checkpoints of the models leads to less generalizing ensemble models. Methods used to generate multiple checkpoints during one training process shorten the ensemble model training time significantly. However, they do not affect the inference time. The disadvantage of cyclical methods implies that a smaller variety of submodels are obtained in relation to each other than in the case of training them in independent processes. This results in higher classification accuracy with ensemble models built from independently trained classifiers.

$Multi-teacher$ knowledge distillation enables the aggregation of knowledge from many $teachers$ only into the single-weight space of a single $student$ model. It can, therefore, be considered an extremely efficient ensemble learning algorithm that eliminates the biggest problem of these mechanics - the linear increase in inference computational complexity when the number of submodels is growing. The $student$ model is characterized by a slightly less precise classification than the ensemble of all $teachers$ by averaging the probabilities they return in the softmax layer. However, the prediction of the $student$ model requires the inference of one base model, while the ensemble inference requires the inference of all $N$ $teachers$. Compared to a single model trained without knowledge distillation, we observe a significant increase in classification efficiency, with the unchanged weight size of the model and the computational complexity needed for inference. We have tested three methods of mimicking in $multi-teacher$ knowledge distillation: $averaging\ mimicking$, $geometric\ center\ mimicking$, and simultaneously $mimicking\ all\ teachers$. The last-mentioned method performs the best. Averaging is an operation that leads to the loss of some information. When we create a knowledge distillation pipeline based on the $teachers'$ average output response, part of the information is irretrievably lost (in the sense that the operation of calculating the mean is a one-way operation, we cannot recreate the individual values knowing only the average value). We decided that providing the $student$ model with more complete feedback from the ensemble of $teachers$ will improve the knowledge transfer process, which was confirmed in the experiments we presented. An error function constructed as the sum of the following individual $teachers'$ responses performs better than the error function associated with reproducing only the mean value. We have some ideas for further research to expand on the topic presented in this dissertation. Looking for places where it is possible and justified to use voting algorithms in deep learning, we analyzed the problems related to creating rankings. The problem of creating an appropriate ranking can be presented for many applications, such as information retrieval or recommendation systems [150, 151]. For a given query, a list of documents sorted by level of relevance to the query should be returned. The ranking is a data structure on which voting algorithms naturally operate. Here, it is very easy to apply the translation that the query is the voter and the documents are possible candidates. What's more, in the learn-to-rank systems, we can also use team learning, as a result of which we will receive several rankings and on their basis, we will want to create the final ranking. The election algorithms seem to fit very well here. In addition, we can use this mechanism for multi-criteria ranking, creating several models with different objectives and then using the voting procedure to find the ranking that best meets all the assumptions. In the context of the main topic presented in this dissertation, knowledge distillation, we have several concepts that we want to expand, examine, and verify in future research. We focused on the domain of machine vision, but knowledge distillation has a much wider application. Equally, or even more so, is the use of smaller distilled *student* models in natural language processing [152–154], where the underlying models have billions of parameters. Also in this domain, it is possible to use several models as *teachers* and try to transfer knowledge to one *student* model in three. As mentioned earlier, knowledge transfer can be divided into three main types: response-based knowledge, feature-based knowledge, and relation-based knowledge. While developing the multi-*teacher* approach, our focus was on response-based knowledge. This was due to the fact that this method of knowledge transfer allows for great flexibility in terms of the sub-models included in the team model which is the *teacher*. However, with adequate effort and appropriate constraints, it is possible to use other types of knowledge transfer in a $multi-teacher$ approach, which may also be one of the issues for future research. Recently, techniques have been developed to introduce an intermediary between the *teacher* and the *student*, the assistant. Establishing such an intermediary increases the efficiency of the knowledge distillation process [155, 156]. We could also test this modification in conjunction with our $multi-teacher$ knowledge distillation framework.

In our experiments, we used the $multi-teacher\ single-student$ approach, where multiple $teacher$ models were trained on different subsets of the training data and then distilled their knowledge to a single $student$ model. We found that this approach was effective in reducing the computational complexity of the ensemble almost without sacrificing accuracy. Overall, our results demonstrate the effectiveness of using knowledge distillation as a mechanism for decision-fusion and aggregation of model predictions in ensemble learning. This approach has the potential to significantly improve the efficiency of ensemble learning, making it more practical for use in real-world applications.

# Implementation

## Contents

In this chapter, the implementation aspect of this doctoral dissertation will be presented. From the very beginning, the entire research was conducted with the potential application in mind. From the implementation point of view, the problem was to train models with limited data resources. We checked ensembling as a mechanism to increase the generality of the machine learning model with limited data constraints, but there was a problem of increasing computational complexity, which was also unacceptable. $Multi-teacher$ knowledge distillation turned out to be a great solution. The application results are presented below.

In the first part, we will present the main result of the implementation, which is a software package developed by us on the basis of $multi-teacher$ knowledge distillation framework analysis, that allows to a large extent the automation of the process of training machine learning models with the use of the $multi-teacher$ knowledge distillation method. The software is used by Neuralbit Technologies in machine learning projects. The second section presents our cooperation with the Air Force Institute of Technology (AFIT, pol. Instytut Techniczny Wojsk Lotniczych, Warsaw, Poland) and its results. Using our software, we have presented a prototype of a system that automates the assessment of the condition of an aircraft fuselage. In the last section, we present one of the projects implemented by Neuralbit Technologies company, i.e., SmokeFinder. The aim of this project is to enable automatic smoke detection in the recording of observation cameras and, ultimately, to increase wildfire safety in forest areas. It is an example of a full implementation of the technology developed by us. The heart of SmokeFinder is the smoke detection algorithm, which was based on a $multi-teacher$ knowledge distillation framework, described in the previous chapters.

## 4.1 Multi-teacher knowledge distillation framework

We have developed a highly automated tool to generate optimal models using the $multi-teacher$ knowledge distillation framework. The library is written in Python3 [157] programming language. The machine learning framework is TensorFlow [158] with Keras [136]. Additionally, we use the pandas [137] and NumPy [138] libraries for data processing. For efficient parallel data processing and computing in Python, we use Dask library [139]. We used matplotlib [140] and seaborn [141] packages for visualization. In the following, we present the various stages of $student$ model preparation with our software. Figure 4.1 shows a diagram of the following steps:

- **Data acquisition and preparation** - The data set is automatically divided into the optimal subsets (in this context, such data splits that give the greatest differentiation between the individual subsets) of the dataset. As shown in Figure 4.1, the user can add additional restrictions that define how the data are to be allocated to: training, validation, and test parts; which examples must or cannot be in a common subset. We also use feature extraction [159–161] by embedding images from the convolutional parts of pre-trained models [162], we also use CLIP (Contrastive Language-Image Pre-Training) [163] which generates embedding with an extremely high level of understanding of the full semantics in the image. Feature extraction is performed to determine the semantic similarity between the examples and to detect outliers. The user can quickly see examples of the dataset that differ from the others. Data statistics are calculated and then used to standardize the samples during training.

- **Training** *teacher* **models** - After the data has been preprocessed, the *teacher* model training phase begins. The *teachers* may differ in the training subset. They can also differ in the architecture used and can be pre-trained on various large datasets such as ImageNet [164], SVHN [165] or COCO [166]. The use of pre-trained models and transfer learning [167, 168] reduces the problem of overfitting on small data sets and generally creates models that perform better than training from scratch. In the scientific part, we intentionally did not introduce $multi-teacher$ knowledge distillation in which *teacher* models are pretrained and use transfer learning or have different architectures. We wanted to carefully examine the impact of the knowledge distillation itself, especially the influence of different variants of *teachers* mimicking by *students*. Additional factors would significantly obscure the phenomenon under study. In applications on the other hand, we want *teacher* models as general as possible to get the best results and the best performance.

- *Student* **training** - When the data and *teachers'* models are prepared, the process of $multi-teacher$ knowledge distillation and training the final *student* model can begin. We use the flexibility of the knowledge distillation framework and enable tuning of hyperparameters of the *student* model. Both the parameters of the *student* model itself and the knowledge distillation process are tuned. e.g. *teacher* training level or *alpha* parameter. For hyperparameter tuning, we use Ray Tune [169] with Optuna [170] integration. The library above is framework-agnostic, which means that we can optimize models developed in TensorFlow, but it is also possible to easily use other frameworks such as PyTorch [171] or LightGBM [172]. Various methods can be used to search the hyperparameter space, such as grid search, random search, or Bayesian optimization [173–176]. Figure 4.2 shows a schematic comparison of the grid search, random search, and Bayesian optimization. In order to reduce hyperparameter search costs, we use pruners that evaluate trial performances during model training and decide whether to stop a given trial. Recently several high-performance trial pruning algorithms have been developed, such as, for example, Async Successive Halving [177], Median Stopping Rule [178] or Hyperband [179]. The early stopping mechanism is also used, which interrupts the training in a situation where the improvement of the validation loss is not observed for a certain number of epochs. We can also apply constraints to the trained model. The most common case is quantization-aware training, which we use when we know that the model will ultimately work on the edge, i.e. at the data acquisition site. The objective function can also be precisely defined, the most common case is when the false negatives are either more, or less important than the false positives, and the standard accuracy metric needs to be adjusted.

- **Model validation** - the trained *student* model is validated in the test set, a report is automatically generated containing the full characteristics of the *student* model, taking into account metrics such as accuracy, recall, precision, true positive rate, false positive rate receiver operating characteristic curve (ROC) and area under the ROC curve (AUC).

- **Model export** - The finished model is stored in its original format, which enables inference on clusters equipped with GPUs. It is also possible to export the model to TensorFlow Lite format, which enables model inference on mobile and edge devices. The model can also be exported to TensorFlow.js format, which enables inference directly in the browser.

In the next two sections, we will present specific applications in which we have used our software to solve real-world problems.

## 4.2   Aircraft inspections

Corrosion, fatigue, and corrosion-fatigue cracking are the most common types of structural problems experienced in the aerospace industry [180, 181]. These damage modes can potentially lead to an unanticipated out-of-service time or to catastrophic failure. To ensure the safety of aircraft structures, regular maintenance is necessary using non-destructive visual inspection methods (NDI) [182]. Traditionally, visual inspections are performed by human operators who scan the fuselage of the aircraft to find corrosion, cracks, and incidental damage. However, this is a costly and time-consuming procedure that is likely to be subject to human errors caused by mental fatigue and boredom. In our latest work [183] in cooperation with a team led by prof. Krzysztof Dragan, from the Air Force Institute of Technology (AFIT, Warsaw, Poland) and being consulted by prof dr Stan Matwin from Dalhousie University, Halifax, Canada, and dr Jerzy Komorowski, Warsaw Technical University, Poland, we present a broad coverage of the application of machine learning in non-destructive inspection in the aerospace industry. We have developed a convolutional neural network that automatically detects small spots of corrosion on the fuselage surface and rivets. We have developed software to support the operator in recognizing damaged fragments [184]. Due to the domain of operation and, above all, security issues, image analysis must take place on the edge, which is associated with the problems of limited computing resources. We also presented the utility, and legitimacy, and verified the possibility of implementing our $multi-teacher$ knowledge distillation mechanism. In this section, we present the most important results of this project.

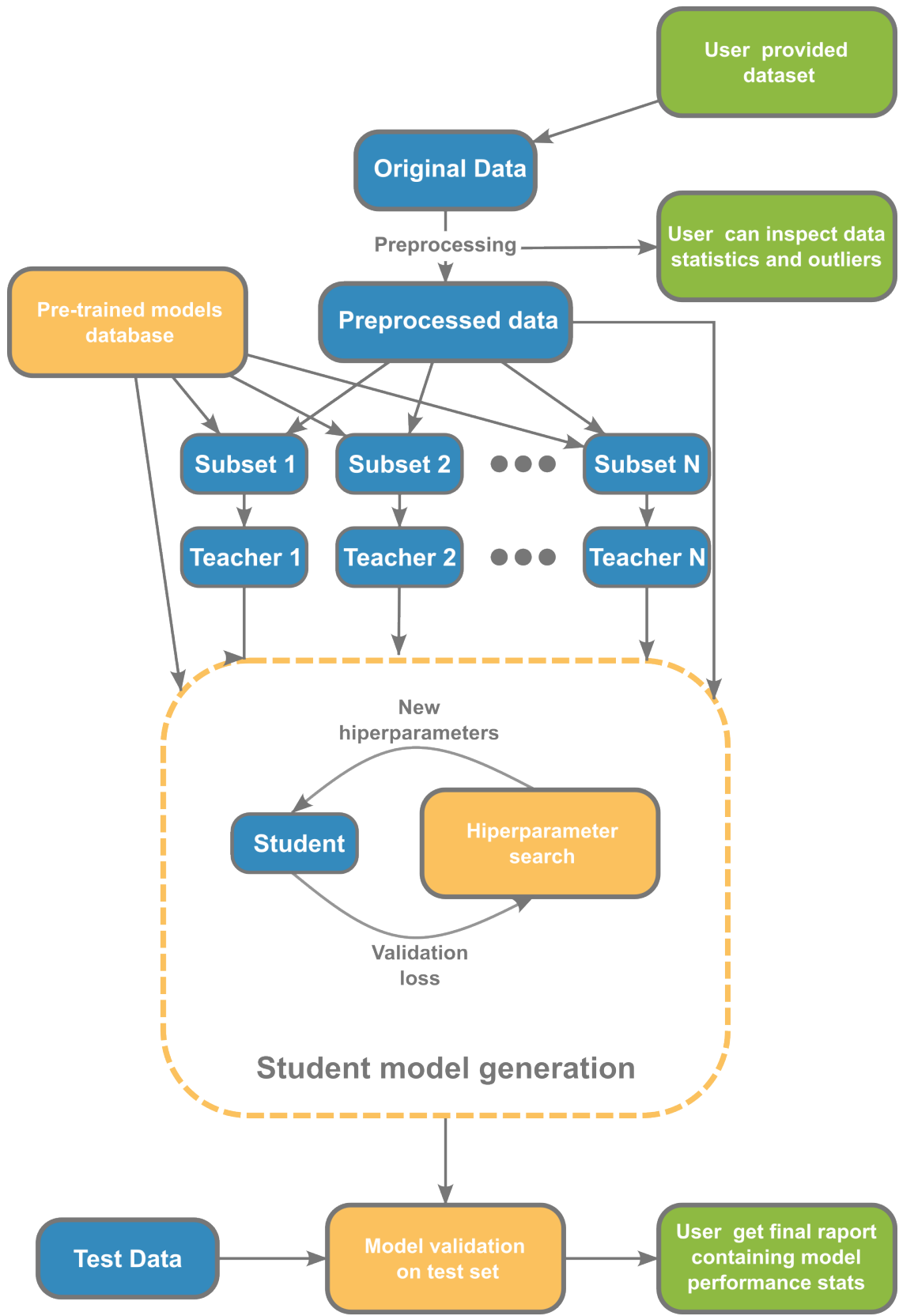

**Figure 4.1:** Operational diagram of our library to automate the generation of machine learning models using $multi - teacher$ knowledge distillation. The subsequent stages of creating the final model are presented.

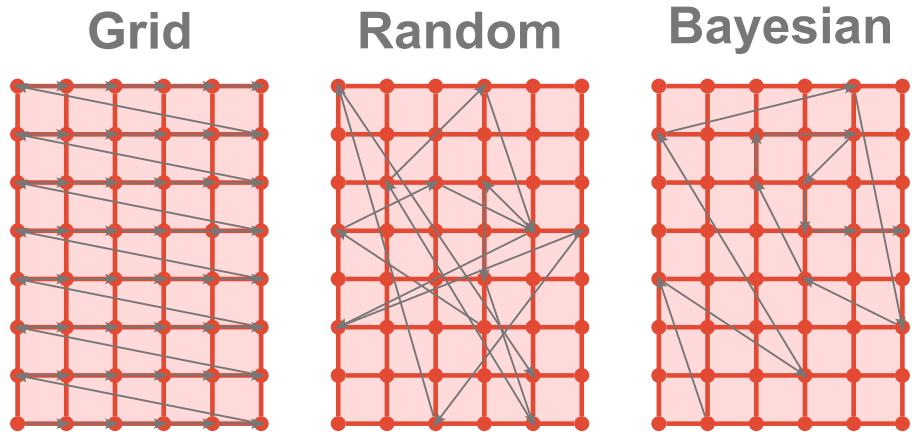

**Figure 4.2:** Schematic comparison of grid search **(left)**, random search **(center)** and Bayesian optimization **(right)**. The grid search is based on the even distribution of test points in the hyperparameter space. The random search draws a random set of parameters each time; previous draws do not affect the next draws. Bayesian optimization uses the knowledge of already performed trials to adjust the generation of subsequent trials; in other words, the metaparameter generator focuses on searches that, based on the current experience, seem to be more promising.

### 4.2.1 DAIS (image acquisition system)

Data for our study come from the DAIS (D-Sight Aircraft Inspection System) image acquisition system designed by dr Jerzy Komorowski, Warsaw Technical University, Poland [185] widely used by the Polish air force. DAIS images can enhance hidden corrosion spots that are invisible to the naked eye under similar lighting conditions. In this way, the digital image preprocessing stage can be skipped. Currently, trained technicians perform a visual inspection of the images captured by DAIS. To decrease the costs while simultaneously increasing the reliability of this time-consuming procedure, herein we propose supporting it with an autonomous system based on advanced neural network architectures. From an application point of view, the main objective of this research was to partially automate and improve aircraft fuselage inspections. D-Sight [186, 187] is an optical double-pass retroreflection surface inspection technique created by Diffracto Ltd. from Canada. It is a patented method to visualize very small surface distortions outside the plane, such as dents and corrosion. The D-Sight optical system consists of a retroreflective screen, camera, a light source, and a fuselage fragment tested (Figure 4.3). Light from a standard divergent source is reflected off the sample. The surface of the sample must be reflective. The reflected light is then shone onto a reflective screen, which consists of many semi-silvered glass spheres (typical diameter $60\ m$). This screen tries to redirect all incident light rays at the same angle to the starting point of reflection on the sample surface. However, the screen is not perfectly reflective and actually returns a divergent cone of light rather than a single beam at the same angle. This imperfection of the reflective screen creates the D-Sight effect. The light is reflected again by the sample and collected by a camera slightly away from the light source. When the examined sample is perfectly flat, the camera sees a sample with uniformly distributed light intensity on the surface. However, an out-of-plane surface distortion will result in local intensity differences. DAIS system [188] uses this imaging technology for damage detection that is not visible to the naked eye. Figure 4.3 presents an overview drawing that contains the operation principle of the DAIS imaging system and a photo showing the fuselage image acquisition process. The corrosion detection system in the aircraft fuselage consists of many modern non-invasive visual inspection techniques presented in [189], also including a highly modernized imaging tool based on the D-Sight methodology.

Thanks to the Air Force Institute of Technology (Warsaw, Poland), we got access to data representing the images acquired using D-Sight technology. We received approximately $1.3 \times 10^4$ labeled images ($640 \times 480$ pixels). Labels include the year of testing, the anonymized $id$ of the aircraft, and the label that represents the extent of corrosion damage. For security reasons, the real $id$ of the type of aircraft has been removed, but due to its identification number, we were sure that the test, validation, and training data were fully separated, that is, when the machine samples with $id = n$ were included in the training set, then no sample from the inspection of this aircraft was included in the validation or test sets.

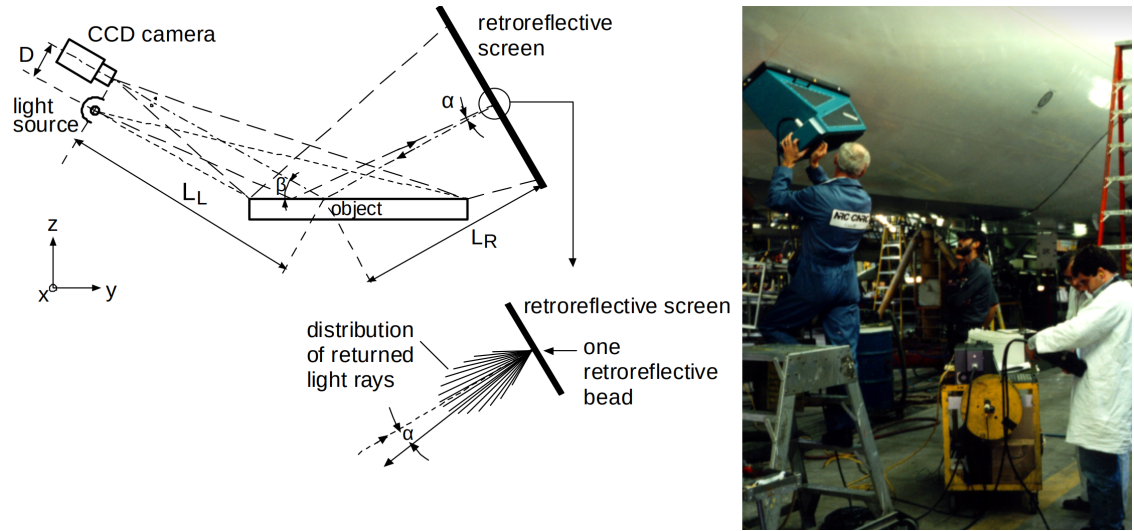

**Figure 4.3: Left**: a scheme of operation of the DAIS imaging system, from [187]. **Right**: the image that demonstrates the process of captioning aircraft images.

Figure 4.4 shows the details of the data, that is, the frequency distribution of the samples according to the year of technical examination and an aircraft *id*. Sample images of the DAIS system are also shown. Our aim was to classify the images according to the strength of the identified damage. Due to the imbalanced data set, we decided to consider this problem as a binary classification: *no damage* and *damage detected*. The original images are in $640 \times 480$ resolution; however, following the guidelines of the authors of the models used, we reduced the resolution to $320 \times 240$ for training and inference speed up. The tests, carried out while training the models in full resolution, showed a minimal decrease in classification accuracy [190]. Furthermore, the training and inference time was many times longer than that obtained for lower-resolution images.

### 4.2.2 Experiment description

Using previous analyzes, we have selected ResNet50 [4] as the convolutional neural network baseline architecture. We show in the supplementary material [190] that this architecture produces the best and most stable results compared to the several others. ResNet50 training setup is as follows: Adam optimizer[76] with *learning rate* $= 0.001$, $\beta_1 = 0.9$, $\beta_2 = 0.999$, and $\epsilon = 1e - 07$ and *batch size* $= 128$, *number of epochs* $= 150$. The data set was divided into training, validation, and test parts, according to the aircraft *id*. Machine samples with *id* between 1 and 30 were assigned to the training data set (10, 534 examples), from *id* = between 31 and 34 were assigned to the validation set (1, 463 examples), while samples from aircraft with *id* between 35 and 37 were assigned to the test set (1297 examples). We used a simple augmentation that involved a random flip on the horizontal and vertical axes. We consciously resigned from other augmentation methods that disturb the levels of contrast, saturation, and brightness because the physics behind the DAIS method is based on generating local disturbances of these values, in particular brightness. Augmentation in these areas would distort the measurement values that are the target of the analysis. The ensemble classifier consisting of ResNet50 subnets (*teachers*) was trained on different training sub-sets. We generated many ResNet50 subnets, each trained on different randomly generated subsets of training data. The examples were generated in such a way that the percentage of common examples for any two selected subsets was the same. Due to this approach, we obtain the maximum diversity of the subsets of the generated data. In the next step, we applied $multi-teacher$ knowledge distillation to an already trained set of $teachers$ to train a single $student$ model. We chose a number of submodels $N = 5$ and $p-value$ equal to $0.7$ (defined as the size of the training subset relative to the entire training set). Our research has shown that this is the optimal value at which we obtained the best results in this case. The experiments carried out also demonstrated that this value of the factor $p-value$ generates the best (well-diversified) ensembles. A further increase in $N$ did not substantially improved the performance of the ensemble. Smaller values of this parameter led to overfitting, whereas larger values produced submodels with very low diversity levels in relation to each other. We tested and compared three $teachers'$ mimicking variants described in the "Methodology" chapter.

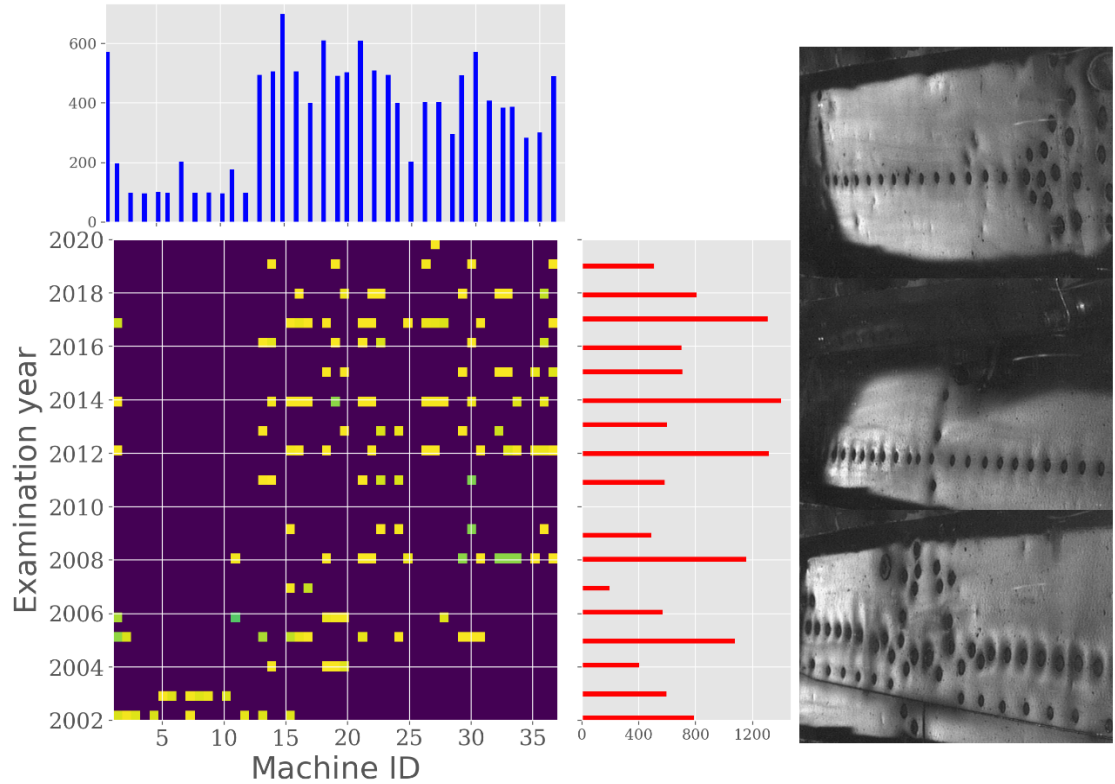

**Figure 4.4: Left**: Distribution of the image examples in the DAIS dataset with machine *id* and inspection year, **right**: DAIS samples. *Top*: *no corrosion*, *Center*: *light corrosion*, *Bottom*: *moderate corrosion*. Total number of examples in the data set by class: *no corrosion and no damage*: 6431, *light corrosion*: 6040, *moderate corrosion*: 578, *strong corrosion*: 0, *minor damage*: 26. Histograms show the marginal distribution according to machine *id* and examination year dimension.

### 4.2.3 Corrosion detection

Depending on the threshold level, we can modify the trade-off between the number of false negatives and false positives. Figure 4.5-left shows the precision and recall metrics depending on the threshold value. We define the threshold as the minimum value of the probability of assigning a sample to the *corrosion detected* class. Images labeled as *corrosion detected* come from several more specific classes that represent various degrees of material failure. We performed the analysis for these specific corrosion classes by comparing how the models treated the samples labeled as *light corrosion* and *moderate corrosion*. The results appeared to be very promising. In the test set, our models were able to recognize $100\%$ *moderate corrosion* samples. Unfortunately, the test set of examples with *moderate corrosion* is limited to only 79 examples. However, from an application point of view, the detection of stronger corrosion examples is the most important and can be a positive test of the usefulness and reliability of our detection algorithm. For safety reasons, in the operation of the autonomous corrosion detection system, the detection of stronger corrosion samples is crucial. It should also be remembered that the entire dataset was manually labeled by experts, and this may be the reason for the existence of some bias (incorrect markings for pairs of *no corrosion - light corrosion*). Figure 4.5-right shows different recall curves for specific corrosion levels.

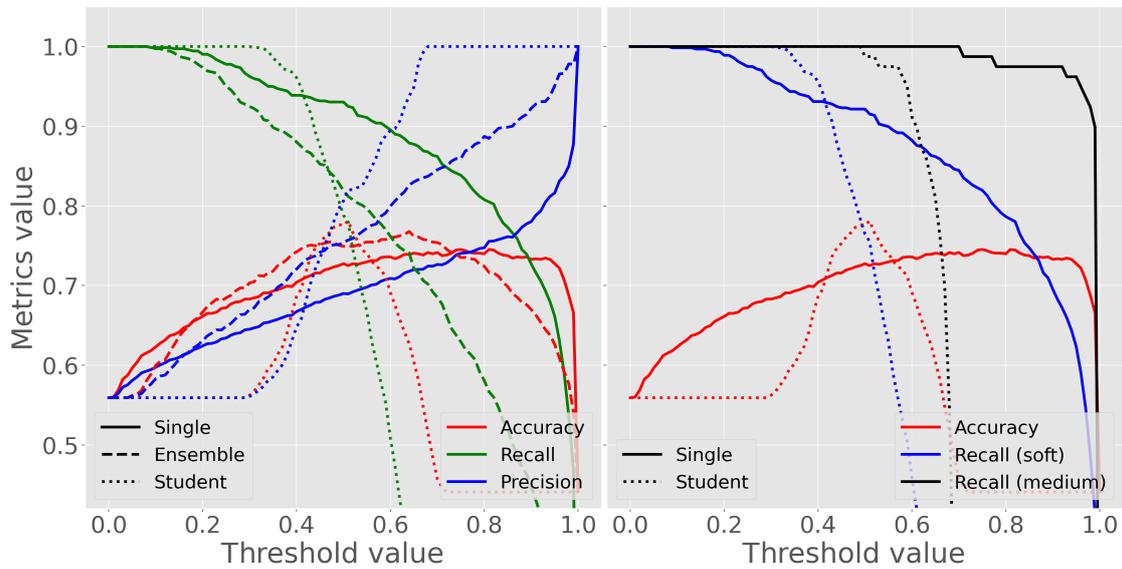

**Figure 4.5: Left**: Characteristics of precision and recall of the models considered. The intersection of the recall and precision lines is at the highest point for the *student* model. **Right**: Precision and recall characteristic with separation for *light* and *moderate* corrosion levels. The *moderate corrosion* samples are much better recognized by models.

To determine the appropriate threshold values for a fair comparison of the various methods, we independently assessed them for each model. The maximum classification accuracy achieved in the validation set was the selection criterion for the thresholds. Then we calculate the remaining metrics in the test set. For the thresholds selected in this way, the *student* model achieves the detection of $100\%$ *moderate corrosion* class while the ensemble and single models get $97.5\%$. It also gives better results on the other metrics. The results are collected in Table 4.1.

To visualize which areas of the images analyzed influence the decision on corrosion classification, we use the Grad-CAM method [191]. The algorithm uses the cumulative gradients calculated in the backpropagation, which are treated as weights to explain network decisions. It can be seen that the highest activations are generated on the riveting line (see Figure 4.6). As hidden corrosion occurs in the rivets, this behavior of the model shows a good level of understanding of the data. Figure 4.6 compares the Grad-CAM activation generated by a single model and by the ensemble. The ensemble activation is more blurred, which shows that most of the image is being analyzed. The use of these heat maps is considered to have high application potential, as it can help operators determine which sites should be examined more thoroughly.

| Used model | Threshold | Accuracy [%] | Recall [%] | Precision [%] | F1 Score [%] | Com* |
|------------|-----------|--------------|------------|---------------|--------------|------|
| Single | 0.89 | 73.6 | 73.96 | 77.75 | 75.81 | 1 |
| Ensemble | 0.62 | 76.25 | 74.81 | 81.24 | 77.89 | 5 |
| *Output avg.* | 0.54 | 74.14 | 69.63 | 81.43 | 75.07 | 1 |
| *Loss avg.* | 0.47 | **76.63** | **84.26** | 76.39 | **80.13** | 1 |
| *Mimic all* | 0.51 | 74.3 | 69.78 | **81.6** | 75.23 | 1 |

**Table 4.1:** Accuracy, recall, precision, and F1 score matrices obtained by tested classifiers on test part of DAIS dataset. Complexity* is expressed as a relative value, where one means the complexity level of a single ResNet50 base model.

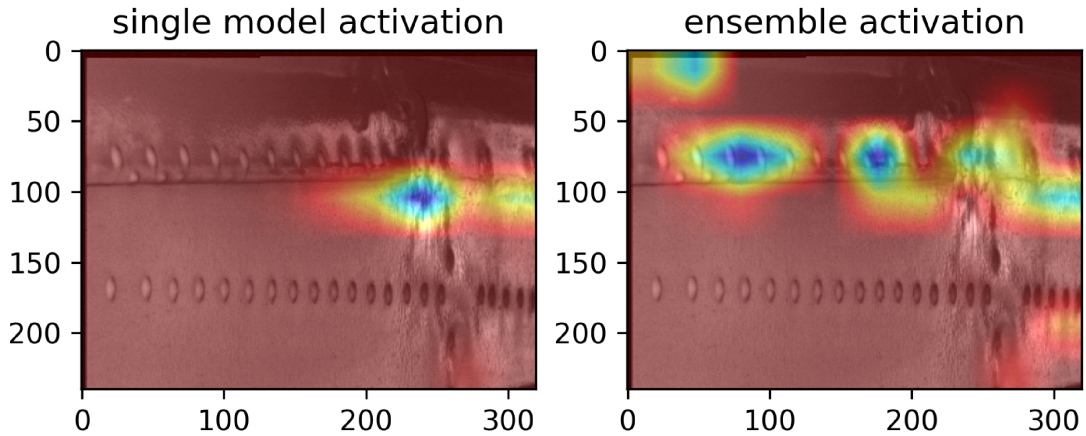

**Figure 4.6:** Grad-CAM activations for a single baseline model (**left**) and the ensemble (**right**) model. The activation map for the ensemble is much wider. From the explainability point of machine learning models, we can determine that ensemble takes more factors into account when generating predictions.

Furthermore, we use the t-SNE [147] data embedding method to visualize the location of samples from the test set in a 2D space. The single model (independently trained, without knowledge distillation) and the *student* model were compared. The feature vectors are collected from the output of the global max-pooling layer, which follows the last convolutional layer. The resulting vector of characteristics had $2,448$ dimensions. Figure 4.7 shows this vector of characteristics embedded in a 2D space for visualization purposes. It is easy to observe a strong separation between the *moderate corrosion* and *no corrosion* classes. The *light corrosion* class lies in the middle area and partially overlaps the *no corrosion* class. This result coincides with the classification metrics achieved by the model for individual corrosion classes. The data points were normalized to better cover the plot canvas. We calculate the Silhouette coefficient [192] (single: $0.0015$, *student*: $0.0359$) to quantitatively show that *student* produces better clustering (a higher coefficient score means better class separation of groups).

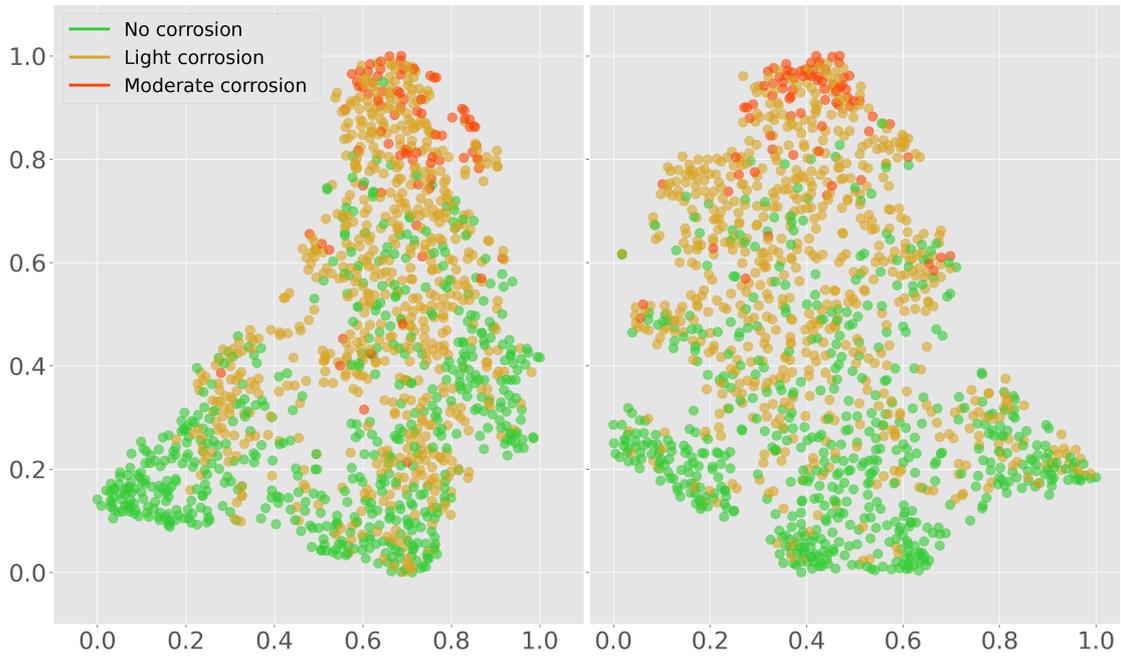

**Figure 4.7:** DAIS samples embedding using t-SNE. The single model and the *multi−teacher* knowledge distillation *student* model were compared.

## 4.3   SmokeFinder

The main objective of the SmokeFinder project[1] is to develop an early smoke detection system using video surveillance systems. The innovation in the approach to the matter is the decentralized operation of machine learning algorithms that analyze images from surveillance cameras. The end device is equipped with a machine learning accelerator that will work directly at the data source, the vision camera. This approach enables access to the highest quality of uncompressed video streaming, which provides better algorithm performance. Due to edge computing, we can achieve a significant reduction in the requirements for the telecommunications infrastructure necessary for video streaming, making the solution technologically viable and economically reasonable. When edge computing is applied, the system will benefit from its accessibility and versatility. It could be integrated in a plug-and-play manner with any existing video surveillance system.

### 4.3.1   Analysis of the occurrence of forest fires and the system of wildfire detection in Poland

The data in this subsection come from an internal report prepared by the Forest Research Institute (pol. Instytut Badawczy Leśnictwa, Raszyn, Poland) at the request of Neuralbit Technologies[193]. Climate changes, manifested by high air temperatures, long-term atmospheric droughts, and the consequent hydrological droughts, pose a significant threat to forest areas from fires. To limit losses caused by them, the observation of forest areas is carried out, the aim of which is to detect a fire in the early stage of development, which will allow effective extinguishing activities. According to the applicable legal status in Poland, the obligation to observe covers forests classified as forest fire risk categories I and II (Polish state forests divide forests into 3 categories of fire risk, the first category being the highest risk). Figure 4.8 shows the distribution of all watch towers in Poland. Currently, there are 700 observation towers, more than half of which are equipped with TV equipment in the vast majority of the FULL HD standard. Figure 4.9 shows an example of the observation towers used to detect wildfires.

---
[1]https://smokefinder.pl/

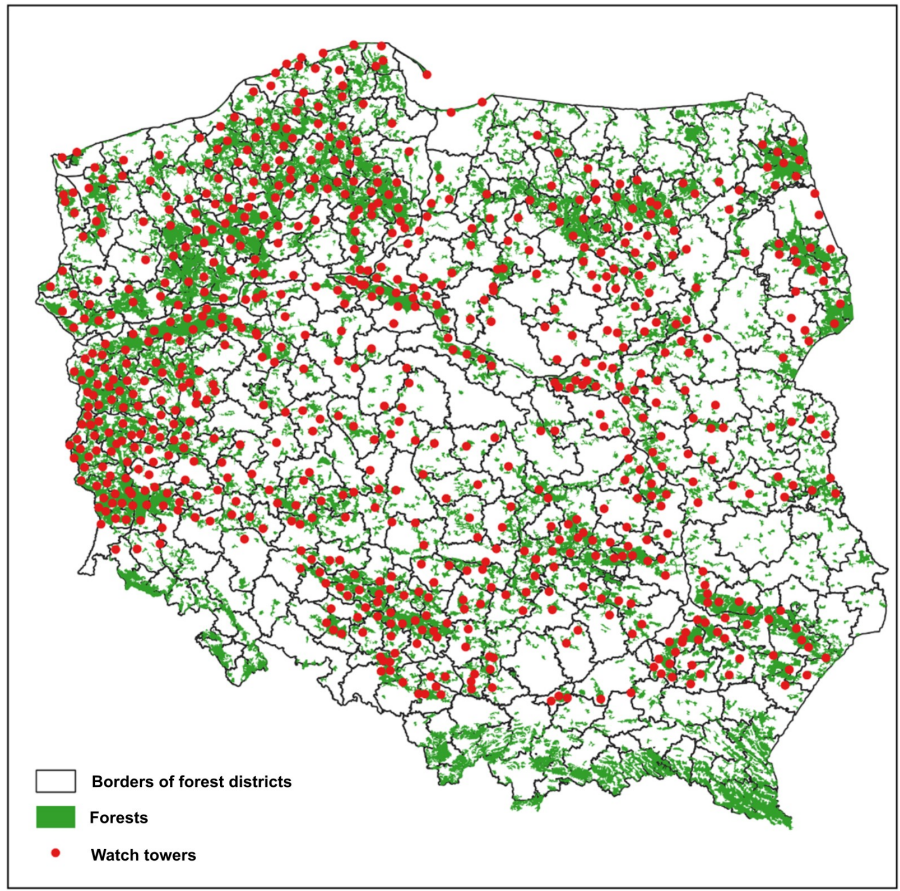

**Figure 4.8:** Location of the observation towers in Poland.

Analysis of the occurrence of fires was carried out based on data collected in the National Forest Fire Information System, which includes data on all forest fires in Poland obtained from the Information System of the State Forests. The analysis covered the years $2012 - 2021$ and included $23,652$ fires. The spatial distribution of the fires is shown in Figure 4.10. The total area of these fires was $8,940.65$ $ha$, of which $6,981.28$ were soil cover fires and $1,959.37$ total fires in stands. Most of the fires, up to 61%, occurred in forest districts classified as category I fire risk, which constitutes only 31% of the total number of forest areas. Another 32% of the number of fires occurred in forest districts of the II fire risk category, which constitutes 42% of their total area, and the lowest in forest districts of the III fire risk category - only 7%, while these forest districts

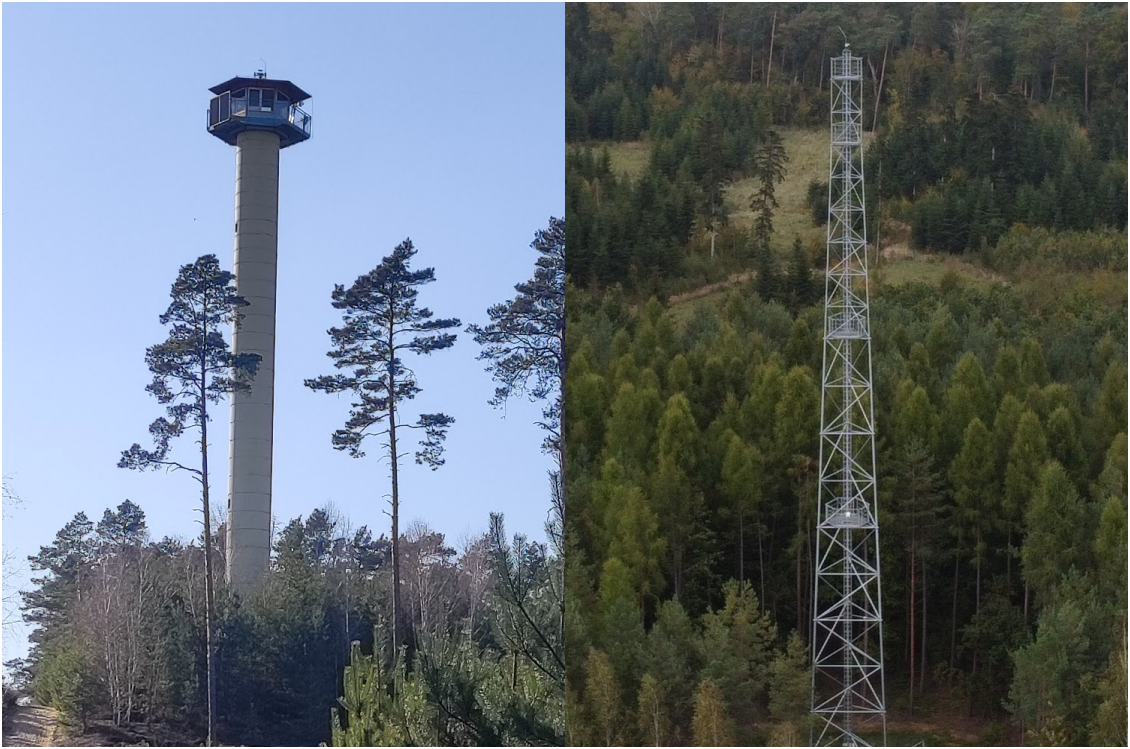

**Figure 4.9: Left**: Older concrete tower. At the top, there is an observation point from which the observer looked for smoke through binoculars. Currently, it is equipped with a remote camera system. **Right**: A newer steel tower built from the ground up with a camera system in mind, it is only accessible for servicing the cameras, without a covered observation point for a human observer.

constitute 27% of the total forest area.

At air temperatures below $14C$, 20.8% of the fires occurred, with their average area the largest and reaching 50.7 ares. The most fires of 43.3% occurred at temperatures between 14 and 24 $C$, and their average area was 42.6 ares. At the highest temperatures, above $24C$, there were slightly fewer fires of 35.9% and their average area was the lowest, which may be due to the highest degree of readiness of the fire protection system in these conditions. A detailed distribution of forest fires according to air temperature is shown in Figure 4.11.

During the period of high relative air humidity above 72%, 14.7% of the fires occurred and their area was the lowest and amounted to 24.7 ares, the most fires of 43.3% occurred at humidity between 40 and 72%, with their area being the largest 42.7 ares. At the lowest humidity below 40%, there were slightly fewer fires, that is, 42.0%, and at the same time their average area was slightly smaller - 37.3 ares, which is most likely due to the higher readiness of the fire protection system. A detailed distribution of the occurrence of fires depending on the relative humidity of the air is shown in Figure 4.12.

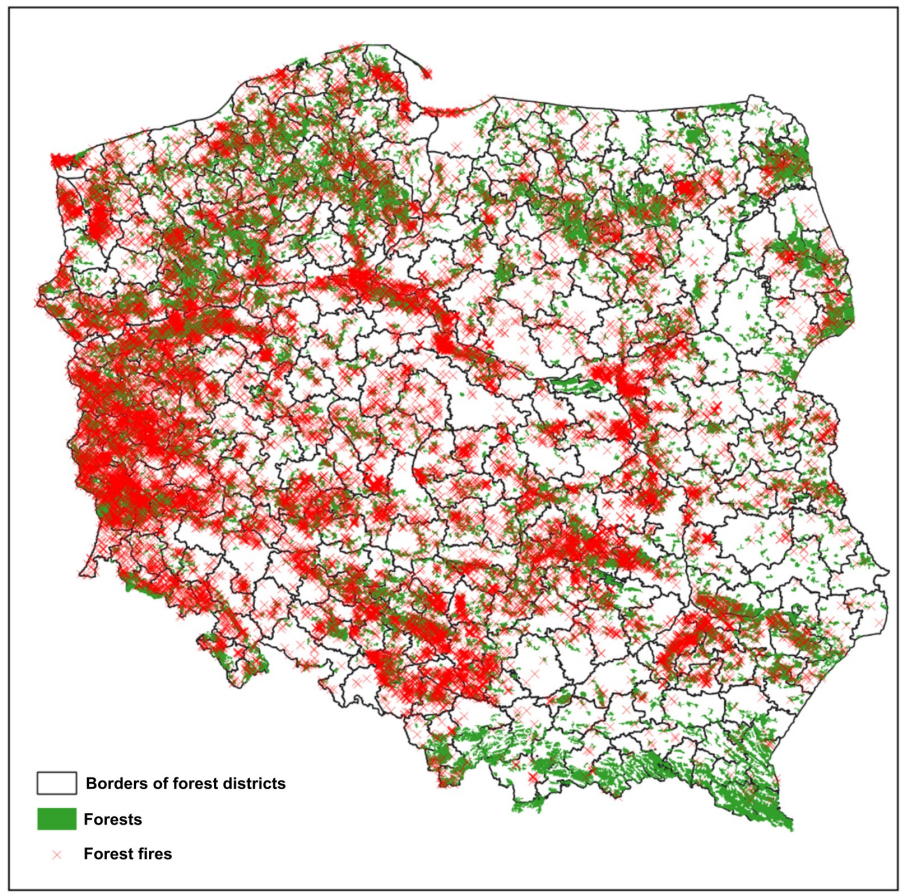

**Figure 4.10:** Spatial distribution of forest fires in Poland in 2012 − 2021

At high values of litter humidity, above 30%,8.5% of all fires occurred and their area was 30.2 ares. This quite large area is due to the fact that under such conditions no protective measures are taken. Slightly more fires of 12.2% occurred with litter humidity between 20% and 30% and their average area was the lowest, only 24.2 ares. This is because, on the one hand, the conditions are not conducive to the intensive spread of fires and, on the other hand, most of the planned protective measures are taken. Most of the fires occurred with humidity between 12% and 20%, and their area was the largest and amounted to 44.0 ares. At the lowest humidity, below 12%, there were slightly fewer fires of 39.3% and their area was clearly lower than for the previous compartment. This is most likely due to the fact that in such conditions all planned protective measures are taken and the fire protection system operates at the highest level. A detailed distribution of the occurrence of fires depending on the humidity of the litter is presented in Figure 4.13.

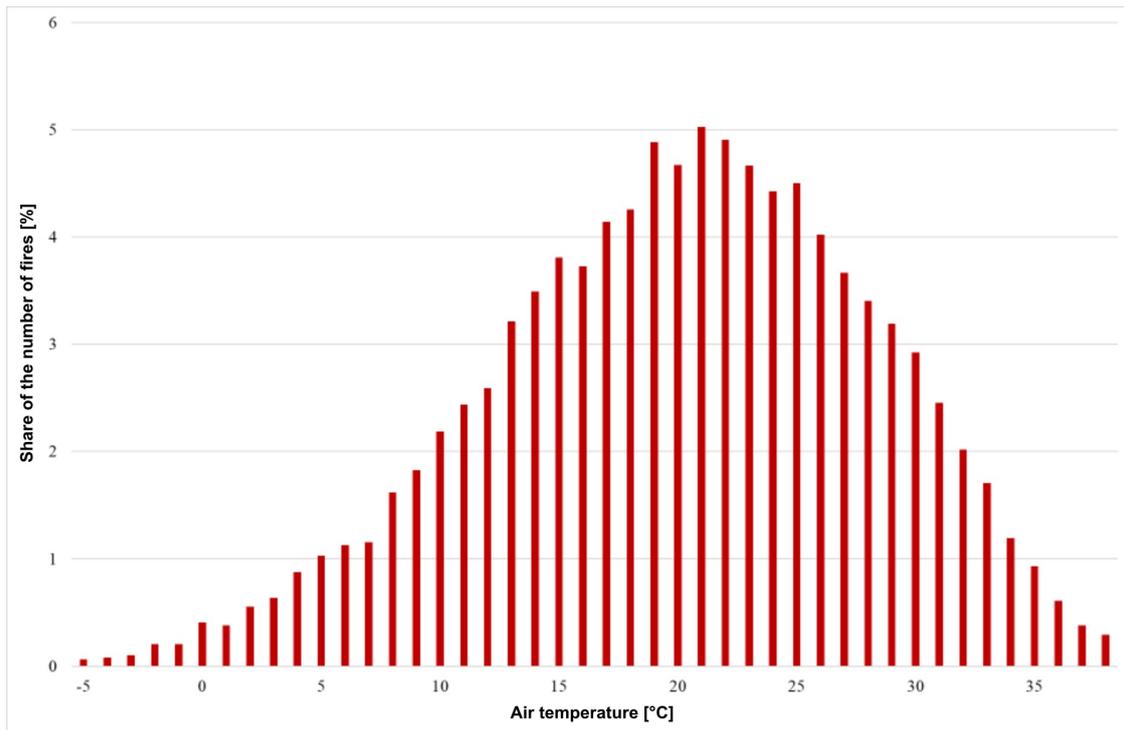

**Figure 4.11:** Distribution of forest fires according to air temperature

This information is very useful in the process of creating automatic smoke detection. We use the smoke probability distribution depending on the meteorological conditions to determine the smoke detection threshold above which we consider the system to have detected smoke. For example, on days with high air humidity (low fire risk), the model must return a higher probability of classifying the sample as smoke to alert the user of the system.

In the period $2012 - 2021$, the highest number of fires, nearly a quarter, occurred in April ($23.05\%$). The number of fires that occurred between May and August was from $12.5\%$ to $15.6\%$. The distribution of medium-sized fires is interesting, which confirms its dependence on the readiness of the fire protection system. The area of fires is clearly larger outside the flammability season or at its beginning or end. Although fires in the winter months from November to

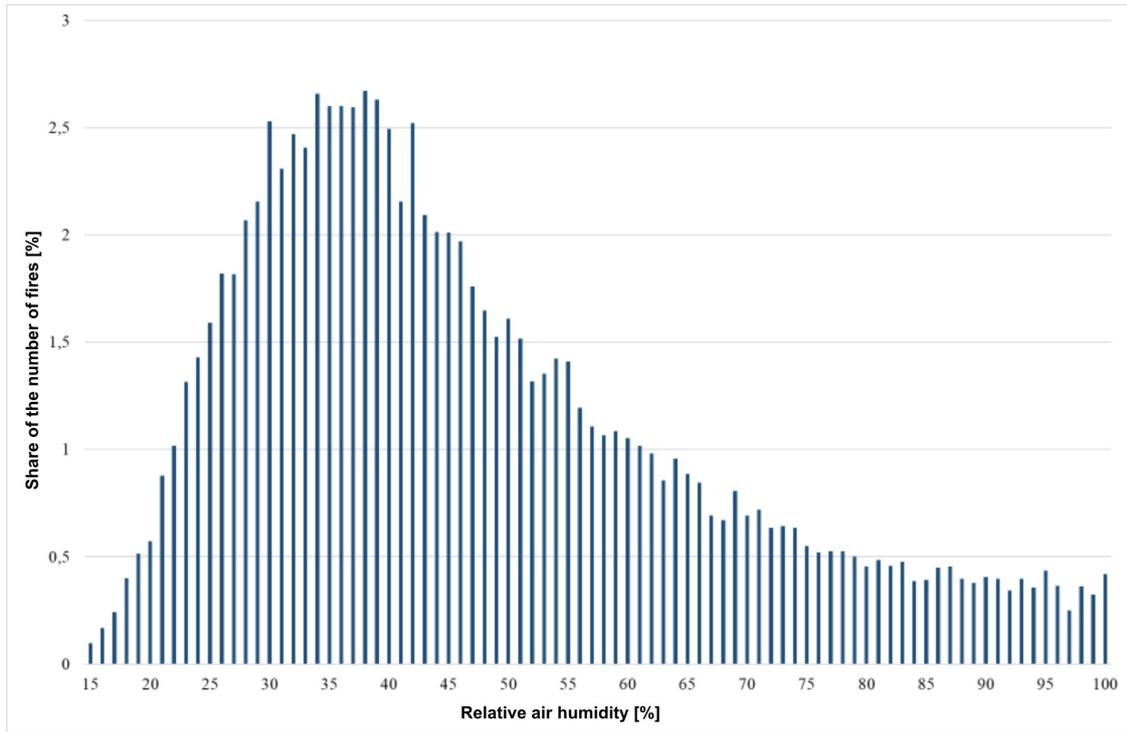

**Figure 4.12:** Distribution of forest fires depending on the relative air humidity

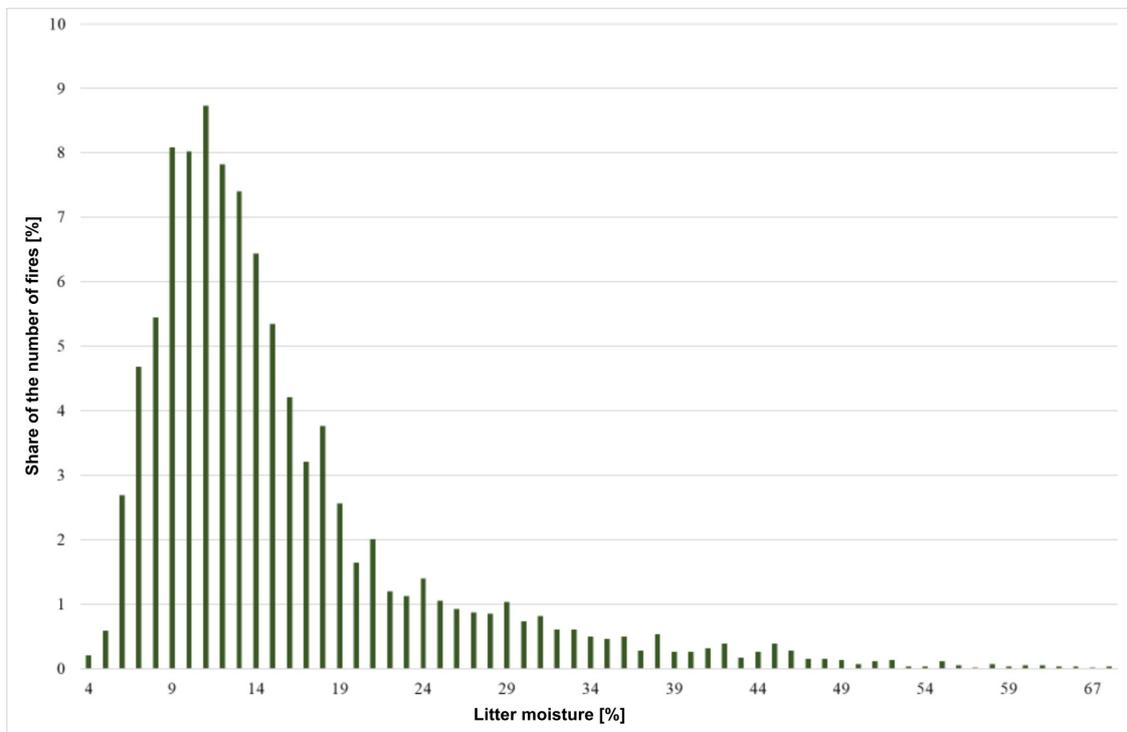

**Figure 4.13:** Distribution of forest fires according to litter humidity

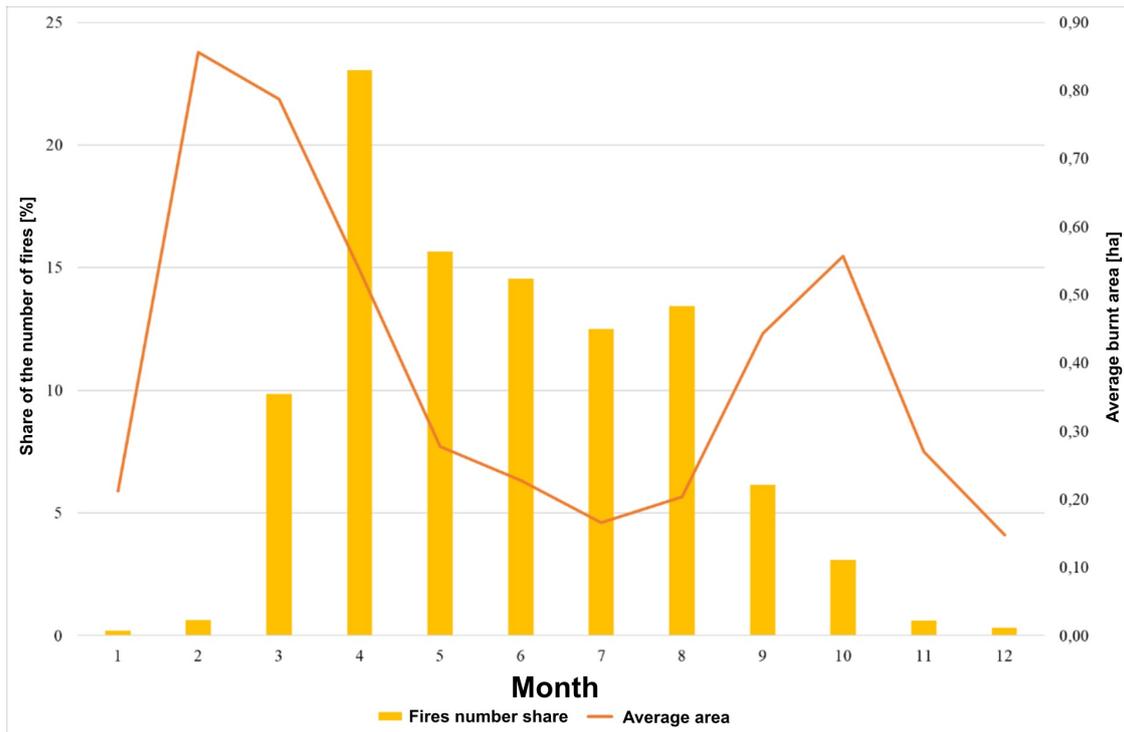

**Figure 4.14:** Distribution of forest fires according to the month of the event

February are not a major problem, despite the large average area, their number is small, which in turn leads to the fact that the total burned area is also not very large. The case of the fires that occurred in March and April, as well as in September and October, is different. So are fires with a large average area that are numerous at the same time, which is associated with a significant total burnt area, and this, in turn, proves the need to develop a fire protection system to make it more effective in these months. The distribution of the number of forest fires and the size of the burned area according to the months of the fire is shown in Figure 4.14.

When analyzing the occurrence of fires by hours, a clear dominance of the noon and early afternoon hours (from 12.00 to 17.00) is visible. On the other hand, the distribution of the average fire area is very characteristic, as long as it is fairly even throughout the day; the fires that started at 9 and 10 o'clock clearly distinguish in this respect, i.e. when the weather conditions are already conducive to the intensive development of a fire and, at the same time, the fire protection system on a given day it is still at the start-up stage. The distribution of the number of forest fires and the size of the burnt area depending on the hours of the occurrence of the fire is shown in Figure 4.15. It also shows how important it is to introduce solutions to support people who work at emergency points on a daily basis. Looking at the monitors for many hours is very boring and it is easy to make a mistake; the automatic smoke detection system is a great support.

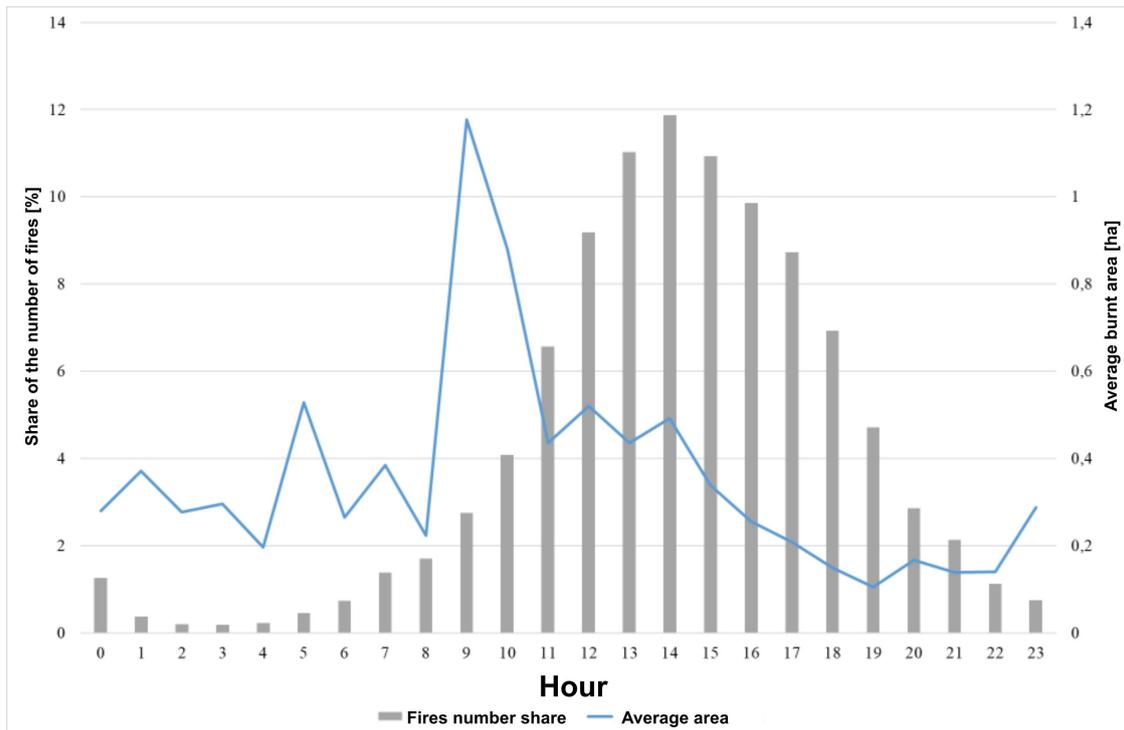

**Figure 4.15:** Distribution of forest fires according to the hour of occurrence.

### 4.3.2    Machine vision empowered smoke detection

Several research studies are focused on computer vision wildfire smoke detection. Maximally stable extremely regions [194, 195] were proposed as a smoke detection method [196]. The spatiotemporal bag-of-features are also used as a smoke detection mechanism [197]. M Bugarić at. all used adaptive estimation of visual smoke detection parameters based on spatial data and fire risk index [198] by tuning the threshold parameters of the system based on the current fire hazard level. There are several studies that utilize wavelets analysis for smoke detection [199, 200], Markov models [201, 202], fluid dynamics [203, 204], random forests [205] and neural networks [206]. These publications can be described as attempts to apply classical machine learning tools to the problem of autonomous smoke detection. Today, more and more attention is paid to deep learning application, in particular convolutional neural networks, for the wildfire smoke detection task [207–214].

In our opinion, this is also the most promising approach at the moment, which is why our research on automatic smoke detection has been focused in this direction. In our case, the challenging problems were limited computing power. SmokeFinder works completely offline. Observation towers are often located far from civilization, and thus from the high bandwidth telecommunications infrastructure, so we decided to conduct model inference on the edge. For energy reasons, we chose Google Coral TPU as a machine learning accelerator. With the appropriate settings, an inference time of a few milliseconds can be achieved [215]. However, the requirements for the size of the models are very restrictive. The use of our $multi-teacher$ knowledge distillation framework under these conditions has brought a huge increase in performance.

### 4.3.3 Smoke Dataset

As part of our cooperation with the Polish Forest Research Institute, we acquired recordings collected by observation cameras. In selected observation towers, we installed recorders that collected video recordings. Then, using custom photo labeling software (based on the open source LabelImg program [216], we identified the frames in which the smoke is visible and determined the location of the smoke in the photos. Additionally, as a data source, we used two smaller publicly available data sets containing photos and videos of smoke.

1. The smoke detection dataset was acquired in July 2012 at the University of Salerno. It consists of 149 videos, each lasting approximately 15 minutes, resulting in more than 35 hours of recording [217, 218].

2. The smoke database collected by the Center for Wildfire Research (Faculty of Electrical Engineering, Mechanical Engineering, and Naval Architecture,

University of Split) contains a collection of nonsegmented wildfire smoke images, photographed both from the ground and from the air. [219, 220]

We use the type of image annotation of bounding boxes and the Pascal VOC [221] image annotation format. we have decided to consider this problem as a classification problem rather than an object detection problem. The decision was mainly due to the higher speed of the classification algorithms, but we also encountered the fundamental problem of correct smoke annotation. It is not an object with sharp edges; therefore, the bounding boxes drawn by the annotators had a relatively large error. Since it is impossible to accurately determine the edges of the smoke, the data set for object detection would be very blurry. To solve this problem, we developed a different approach. By having one bounding box containing smoke, we can generate multiple image slices that share some fragment with the annotation. In this way, we can generate many samples from one photo; we are clipping different fragments of the same photo showing smoke. We imposed several boundary conditions on the generated slices. The maximum and minimum percentages of the smoke area have been determined. The maximum and minimum percentages of the annotation area that must be included in the slice have been defined. The maximum and minimum ratio of the long side to the shorter side was defined, as well as the maximum and minimum area of the sample cut. Figure 4.16 shows a scheme for generating multiple samples from a smoke annotation. Then the selected sliced rectangles are scaled to the target size, e.g. $224 \times 224$. An additional benefit is that this approach allows us to generate the final training dataset with different distributions. By modifying the boundary conditions of generating slices and modifying the filtering of annotations that we want to include in the data source. Figure 4.17 shows some examples from our final dataset representing the class in which the smoke appears.

### 4.3.4 Knowledge distillation framework application

We used the $multi - teacher$ knowledge distillation framework presented earlier to prepare smoke classification models. Table 4.2 shows a summary of the classification accuracy achieved in our test set. We tested various variants of the number of *teachers* and a different level of *teacher* training (training on a training subset of different sizes). For all models, *teachers*, and *students*, we used the EfficientNet-EdgeTPU [222–224] pretrained on the ImageNet dataset as the base model. Also, for all models, *teachers* and *students*, we use Adam optimizer with $\beta_1 = 0.9$, $\beta_2 = 0.999$, and $\epsilon = 1e - 07$ with *learning rate* $= 0.0001$ for the transfer learning phase and *learning rate* $= 0.00001$ for the fine-tuning phase. *Batch size* was set to 64. The aggregation method was the *mimick all* variant, and the *alpha* parameter was 0.5. The use of the *multi − teacher* knowledge distillation framework allowed one to increase the classification accuracy from 93.3% (model trained in a standard way, on a full training set) to 96.3% (the best performing *student* model trained with knowledge distillation) while maintaining the same computational complexity of inference - these are models with the same architecture and number of weights. To further illustrate the performance gains by using the *multi − teacher* knowledge distillation framework, Figure 4.18 shows the precision-recall characteristics for the best obtained baseline model (trained independently) and the best obtained *student* model. For the *student* model, the recall and precision curves are located noticeably higher than for the baseline model. The resulting *student* model is characterized by a smaller number of generated false positives and false negatives. This means that the use of such a model will contribute to increasing the reliability of the fire protection system while minimizing the number of generated false alarms.

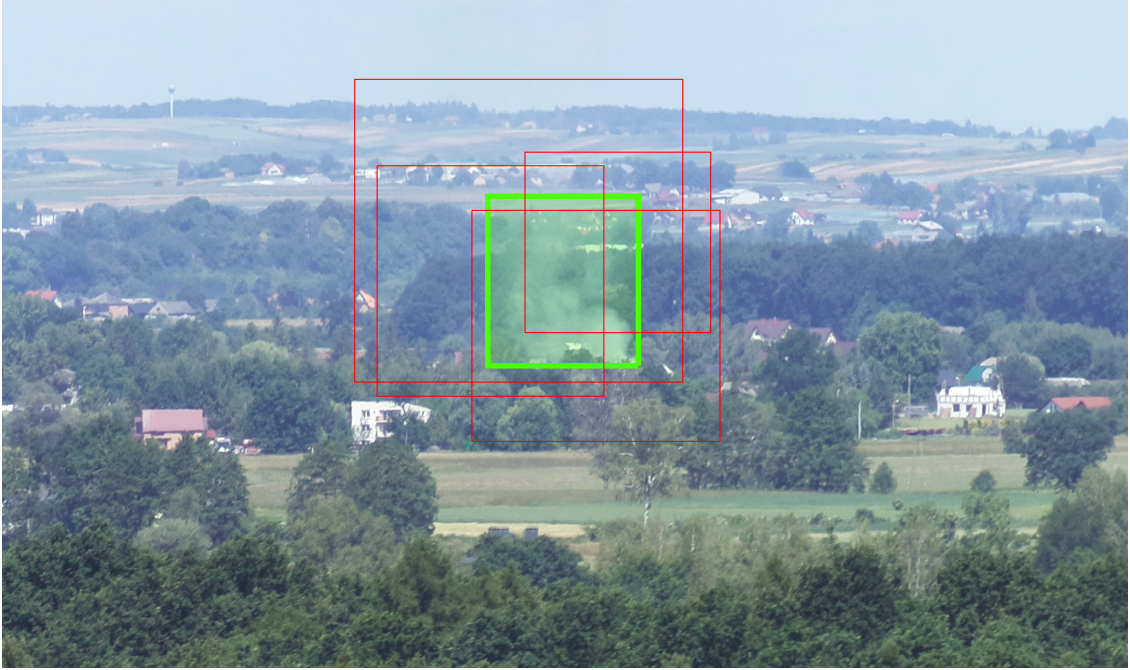

**Figure 4.16:** Photo of the smoke observed from our surveillance camera. The smoke bounding box is marked with a green rectangle. Based on this annotation, we can generate multiple smoke clippings. The proposals for such slices are marked with red rectangles.

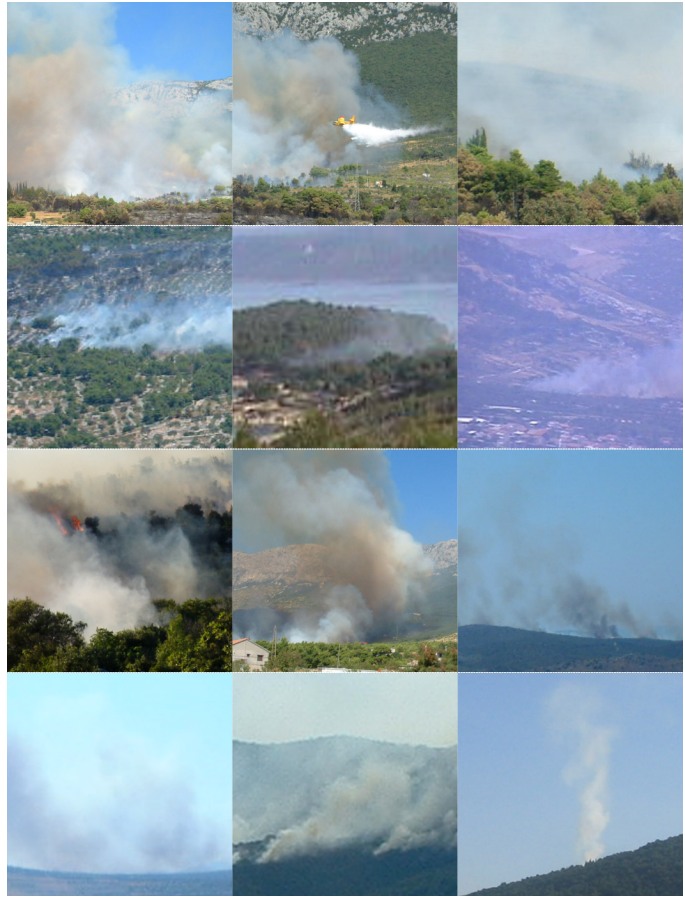

**Figure 4.17:** Samples from our dataset showing the class containing smoke.

Figure 4.19 shows the smoke detection result in a sample test photo. To detect smoke, which is a small fraction of the entire frame area, the photo is broken into small squares that partially overlap. Each square is analyzed separately by a trained model for smoke detection. If the probability of smoke classification is greater than the threshold value in any of the analyzed squares, the fire alarm system operator is alerted. The smoke presented in the photo is barely visible to the human eye, but our model has flawlessly marked the fragments where the smoke occurs. Very often, the smoke in the initial stage of a fire looks like this, so it is very important to be able to detect barely visible smoke, which gives the possibility of the fastest possible reaction, which is crucial in the success of a fire-fighting action. Figure 4.20 shows some sample photos containing smoke with overlaid Grad-CAM activation maps for the class that contains smoke. This visualization technique enables the analysis of which areas of the image influence the decisions generated by the convolutional neural network the most. We can observe that the most significant fragments are the areas on the border of smoke and background.

| Fraction | N teachers | Single acc. [%] | Ensemble acc. [%] | KD acc. [%] |
|----------|-----------|-----------------|-------------------|-------------|
| 1 | 1 | **93.31 (0.47)** | - | |
| | 2 | | 94.11 (0.42) | 93.64 (0.34) |
| | 3 | | 94.92 (0.37) | 94.42 (0.29) |
| | 4 | | 95.58 (0.28) | 95.12 (0.32) |
| | 5 | | 96.29 (0.28) | 96.03 (0.24) |
| 0.9 | 1 | 92.94 (0.39) | - | |
| | 2 | | 93.92 (0.53) | 93.51 (0.38) |
| | 3 | | 95.04 (0.48) | 94.71 (0.34) |
| | 4 | | 95.94 (0.35) | 95.52 (0.47) |
| | 5 | | **96.65 (0.47)** | **96.29 (0.43)** |
| 0.8 | 1 | 92.35 (0.41) | - | |
| | 2 | | 93.34 (0.42) | 92.95 (0.34) |
| | 3 | | 93.89 (0.57) | 93.57 (0.41) |
| | 4 | | 94.57 (0.39) | 94.21 (0.53) |
| | 5 | | 95.02 (0.47) | 94.69 (0.41) |
| 0.7 | 1 | 91.93 (0.45) | - | |
| | 2 | | 92.62 (0.47) | 92.39 (0.41) |
| | 3 | | 93.11 (0.38) | 92.64 (0.29) |
| | 4 | | 93.57 (0.43) | 93.03 (0.46) |
| | 5 | | 94.02 (0.35) | 93.71 (0.38) |
| 0.6 | 1 | 91.56 (0.51) | - | |
| | 2 | | 92.14 (0.44) | 91.91 (0.42) |
| | 3 | | 92.63 (0.33) | 92.27(0.34) |
| | 4 | | 93.12 (0.39) | 92.82 (0.40) |
| | 5 | | 93.68 (0.42) | 93.26 (0.35) |

**Table 4.2:** The accuracy of the smoke classification in our test set obtained by models trained with a different set of parameters.

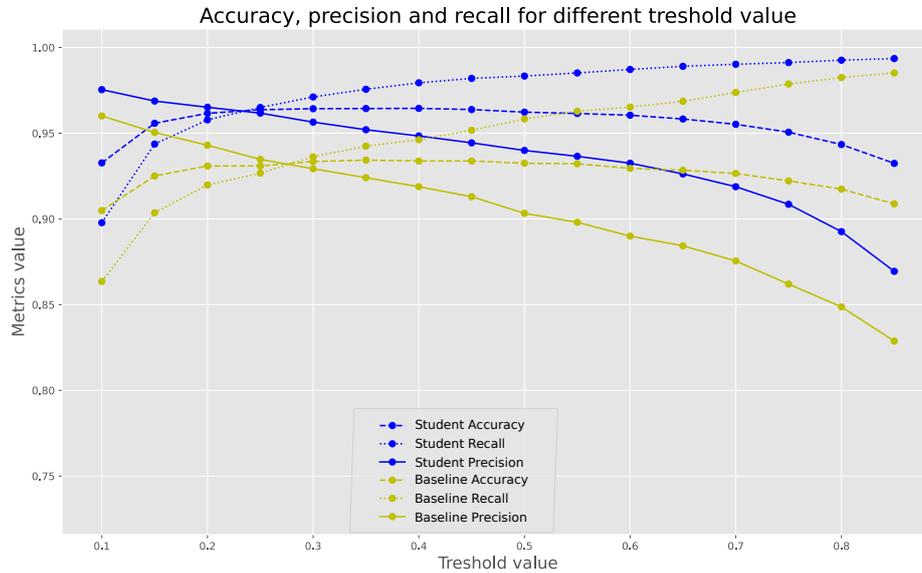

**Figure 4.18:** Precision-Recall characteristics for the best obtained baseline model (yellow lines) and the best obtained *student* model (blue lines). For the *student* model, the recall and precision curves are located noticeably higher than for the baseline model. The resulting *student* model is characterized by a smaller number of generated false positives and false negatives. This means that the use of such a model will contribute to increasing the reliability of the fire protection system while minimizing the number of generated false alarms.

## 4.4 Implementation summary

We have shown that $multi-teacher$ knowledge distillation framework has great potential for implementation applications. We have presented a description of our software package based on insights obtained during an in-depth $multi-teacher$ knowledge distillation analysis. Work on the library for a $multi-teacher$ knowledge distillation framework is completed. We showed a prototype implementation of the proposed methodology in the problem of non-invasive corrosion detection on aircraft fuselage, and we conducted the research in cooperation with Air Force Institute of Technology. We showed that DAIS imagining system can be successfully enhanced with a machine vision system based on our corrosion detection algorithm. Finally, we presented a description of the SmokeFinder project and outlined the problem and characteristics of forest fires in Poland. The model developed as a part of this project, which enables efficient classification of smoke in frames from observation cameras, is an example of the full implementation process of *multi−teacher* knowledge distillation framework. The smoke detection model was trained and then fully evaluated, and the best model to be launched in production was selected. Work on the SmokeFinder in terms of computer vision algorithms has been completed. Work is still underway to improve the hardware part. In summary, *multi−teacher* knowledge distillation solves many problems of generating machine learning models in a situation of constrained data resources and limited computing power because it gives the possibility to increase the generalization of the model without additional expenditure on computing power coupled with the inference phase.

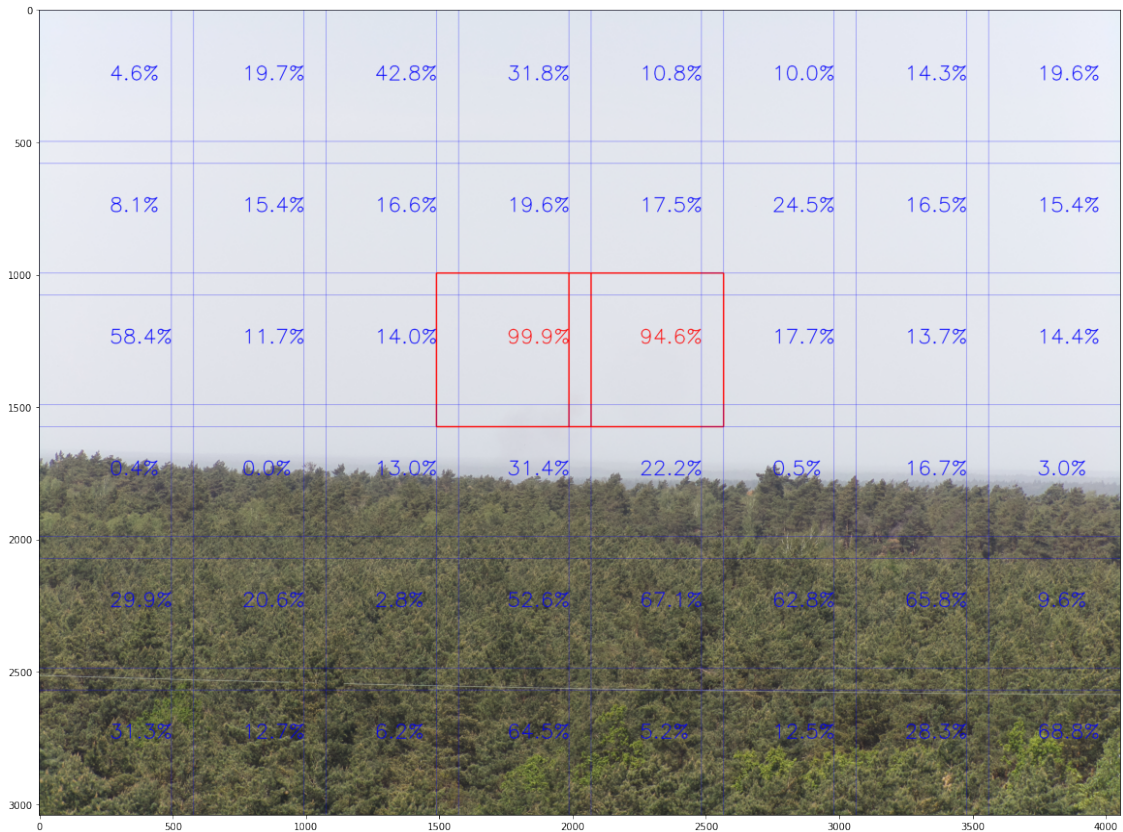

**Figure 4.19:** The process of analyzing a frame from surveillance camera. Individual squares represent the sections that were independently assessed for smoke presence and were the input for our detection model. The percentages in the middle of the squares represent the probability of smoke detection returned by the model. If the probability value is less than the threshold, the squares are blue; if the probability value exceeds the threshold, the squares are red. The smoke presented in the photo is barely visible to the human eye, but our model has flawlessly marked the fragments where the smoke occurs.

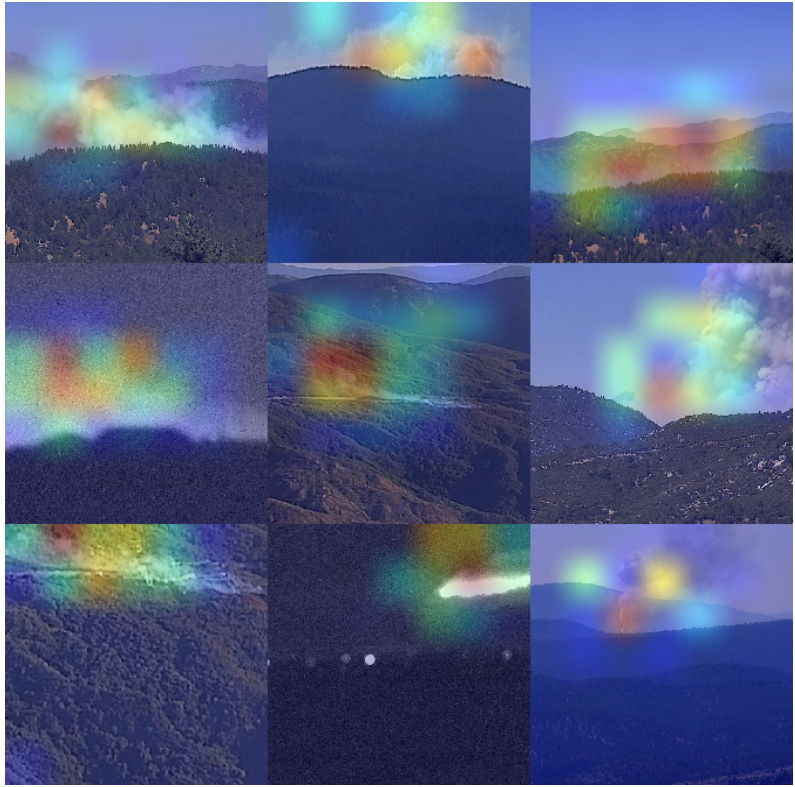

**Figure 4.20:** Grad-CAM activations on several sample photos containing smoke. The visualization shows which parts of the photo affect the classification of the photo as containing smoke by the model.

# Appendices

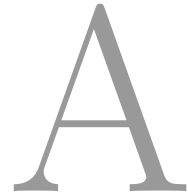

# Multi-teacher knowledge distillation results

Herein, as in the chapter 3 (Results), we present more experiment results for more datasets and different convolutional neural network architectures. Tables A.1, A.3, A.5, A.7 show the accuracy of the student classification according to the knowledge distillation method used for different $p-values$. The *mimicking all* method gives the best results. Knowledge distillation generates a model that is slightly weaker than full ensembling with $N$ times larger computational overhead. Compared to a single model, we observed increases in classification accuracy. Figures A.1, A.5, A.9, A.13 show the accuracy of the student classification, depending on the number of *teachers* used during the knowledge distillation process. Individual data series correspond to different methods of *teachers'* output mimicking by student model and different $p-value$. Figures A.2, A.6, A.10, A.14 show the accuracy of the student classification, depending on $p-value$. This time, individual data series correspond to the number of *teachers* used during different methods of *teachers* mimicking the knowledge distillation process. Tables A.1, A.3, A.5, A.7 presents the aggregated values for all *alpha* parameters we tested. On the other hand, Tables A.2, A.4, A.6, A.8 presents a similar comparison, but with separate cases for different values of the *alpha* parameter. Due to the significant increase in the various values to be shown, we printed information only for the *mimick all* method and the *number of teachers* equal to 3. Figures A.3, A.7, A.11, A.15 shows the dependence of the classification accuracy obtained on the *alpha* parameter used. Different data series mean that a different number of *teachers* is used in the training student model, the $p-value$ is set to 1. Figures A.4, A.8, A.12, A.16, show a twin relationship, but here the data series shows different $p-values$, the *number of teachers* used for training is set to 3.

| Teachers number | Fraction | Single accuracy [%] | Ensemble accuracy [%] | Multi-Teacher KD accuracy [%] | | |
|---|---|---|---|---|---|---|
| | | | | *Output avg.* | *Loss avg.* | Mimick all |
| 2 | 1 | 82.18 (0.55) | 83.35 (0.47) | 81.92 (0.39) | 82.08 (0.54) | **82.72 (0.25)** |
| | 0.9 | 81.29 (0.55) | 82.67 (0.49) | 81.46 (0.41) | 82.07 (0.34) | **82.68 (0.38)** |
| | 0.8 | 80.52 (0.65) | 81.42 (0.52) | 81.55 (0.52) | 81.03 (0.62) | **82.09 (0.24)** |
| | 0.7 | 79.52 (0.53) | 80.43 (0.43) | 81.47 (0.29) | 80.78 (0.34) | **81.28 (0.22)** |
| | 0.6 | 78.17 (0.91) | 79.21 (0.66) | 79.25 (0.66) | 79.94 (0.66) | **81.1 (0.5)** |
| 3 | 1 | 82.18 (0.55) | 84.04 (0.43) | 81.97 (0.61) | 81.18 (0.89) | **83.11 (0.22)** |
| | 0.9 | 81.29 (0.55) | 83.14 (0.51) | 78.76 (0.46) | 81.89 (0.25) | **82.76 (0.31)** |
| | 0.8 | 80.52 (0.65) | 81.89 (0.47) | 81.53 (0.5) | 81.48 (0.4) | **83.08 (0.31)** |
| | 0.7 | 79.52 (0.53) | 80.93 (0.39) | 81.82 (0.28) | 81.64 (0.4) | **82.14 (0.47)** |
| | 0.6 | 78.17 (0.91) | 80.14 (0.53) | 80.31 (0.46) | 80.61 (0.64) | **81.2 (0.51)** |
| 4 | 1 | 82.18 (0.55) | 84.21 (0.35) | 80.77 (0.94) | 82.84 (0.36) | **83.22 (0.21)** |
| | 0.9 | 81.29 (0.55) | 83.78 (0.42) | 80.42 (0.6) | 81.92 (0.43) | **82.73 (0.19)** |
| | 0.8 | 80.52 (0.65) | 82.17 (0.38) | 82.09 (0.48) | 81.83 (0.39) | **82.48 (0.23)** |
| | 0.7 | 79.52 (0.53) | 81.24 (0.45) | 81.02 (0.73) | 81.44 (0.36) | **82.33 (0.25)** |
| | 0.6 | 78.17 (0.91) | 80.61 (0.44) | 80.59 (0.45) | 79.94 (1.05) | **81.68 (0.67)** |
| 5 | 1 | 82.18 (0.55) | 84.35 (0.31) | 83.39 (0.3) | 82.82 (0.45) | **83.66 (0.31)** |
| | 0.9 | 81.29 (0.55) | 84.42 (0.39) | 82.71 (0.48) | 82.17 (0.49) | **83.0 (0.32)** |
| | 0.8 | 80.52 (0.65) | 82.39 (0.41) | 82.31 (0.43) | 81.26 (0.48) | **83.55 (0.22)** |
| | 0.7 | 79.52 (0.53) | 81.52 (0.42) | 81.41 (0.78) | 82.01 (0.5) | **82.19 (0.49)** |
| | 0.6 | 78.17 (0.91) | 80.83 (0.38) | 81.28 (0.56) | 80.91 (0.66) | **81.86 (0.41)** |

**Table A.1:** Classification accuracy comparison for a single model, ensemble model, and student model obtained using various multi-teacher knowledge distillation techniques of *teachers* mimicking. The following data are presented for the ResNet50 baseline model and the CIFAR10 dataset.

| Fraction | *Alpha* parameter | | | |
|---|---|---|---|---|
| | 0 | 0.25 | 0.5 | 0.75 |
| | Accuracy [%] | | | |
| 1 | *83.16 (0.31)* | **83.87 (0.33)** | 83 (0.49) | 82.4 (0.28) |
| 0.9 | 82.08 (0.52) | **83.95 (0.33)** | 83.33 (0.22) | 82.06 (0.53) |
| 0.8 | 82.93 (0.42) | **83.88 (0.07)** | *83.54 (0.64)* | 81.97 (0.19) |
| 0.7 | 80.66 (0.75) | 82.62 (0.84) | **83.09 (0.31)** | *82.92 (0.56)* |
| 0.6 | 80.26 (0.34) | 81.43 (0.27) | 82.14 (0.57) | **81.92 (0.31)** |

**Table A.2:** Classification accuracy for student models obtained using various *alpha* and $p - value$. The number of *teachers* was set to 3 and mimicking method was *mimick all*. The following data are presented for the ResNet50 baseline model and the CIFAR10 dataset.

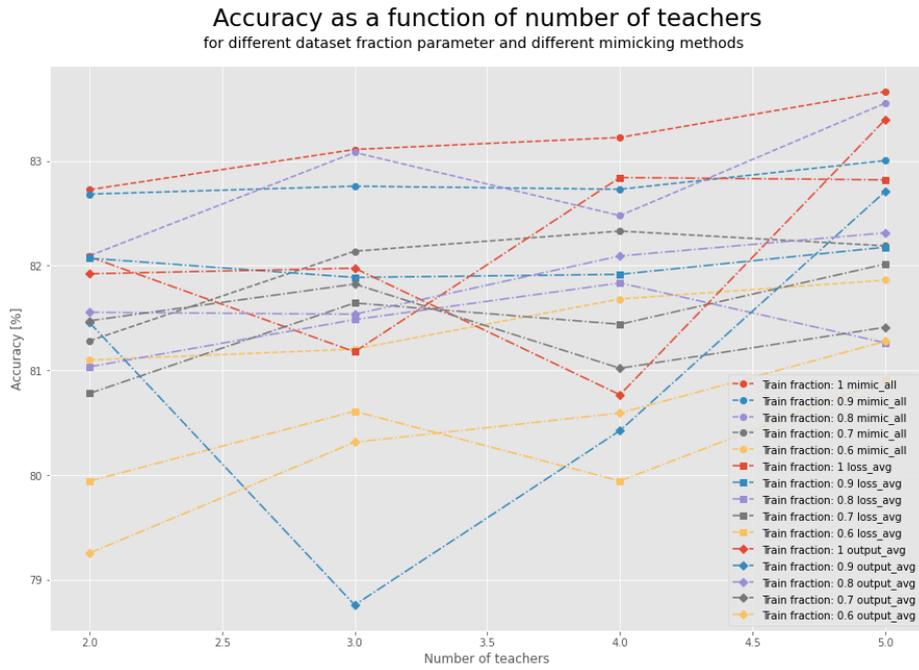

**Figure A.1:** Student classification accuracy depending on *teachers* number for different mimicking method and $p - value$. The following data are presented for the ResNet50 baseline model and the CIFAR10 dataset.

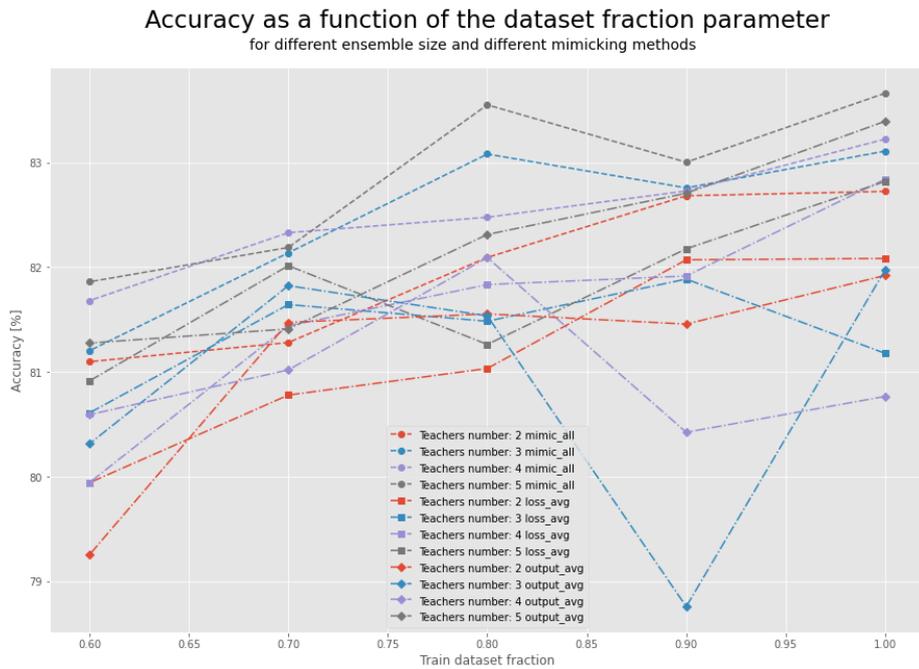

**Figure A.2:** Student classification accuracy depending on $p - value$ for different *teachers* number and used mimicking method. The following data are presented for the ResNet50 baseline model and the CIFAR10 dataset.

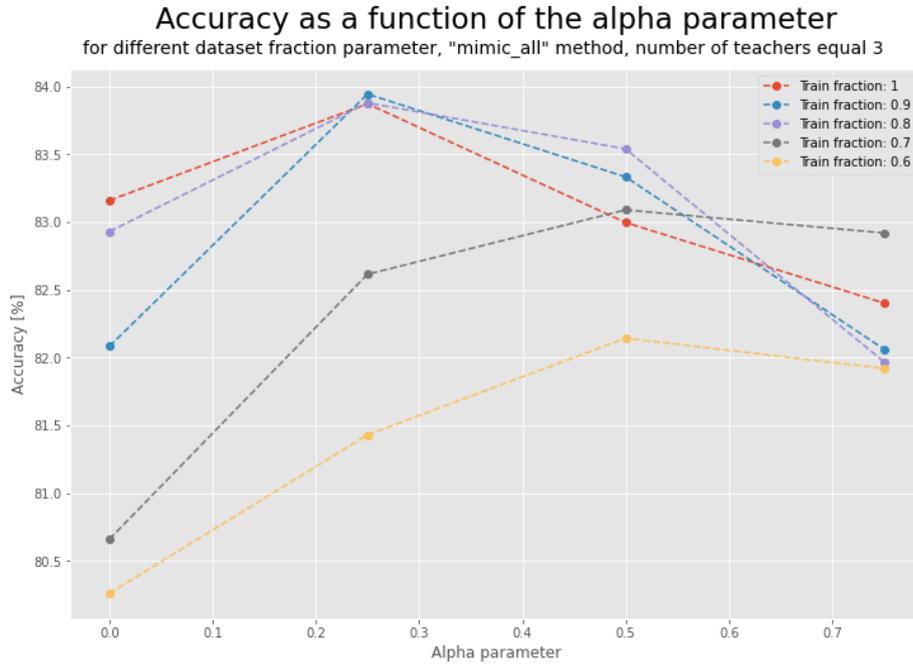

**Figure A.3:** Student classification accuracy depending on *alpha* parameter used, different data series for different *p−value*, the number of *teachers* used for training student are set to 3. The following data are presented for the ResNet50 baseline model and the CIFAR10 dataset.

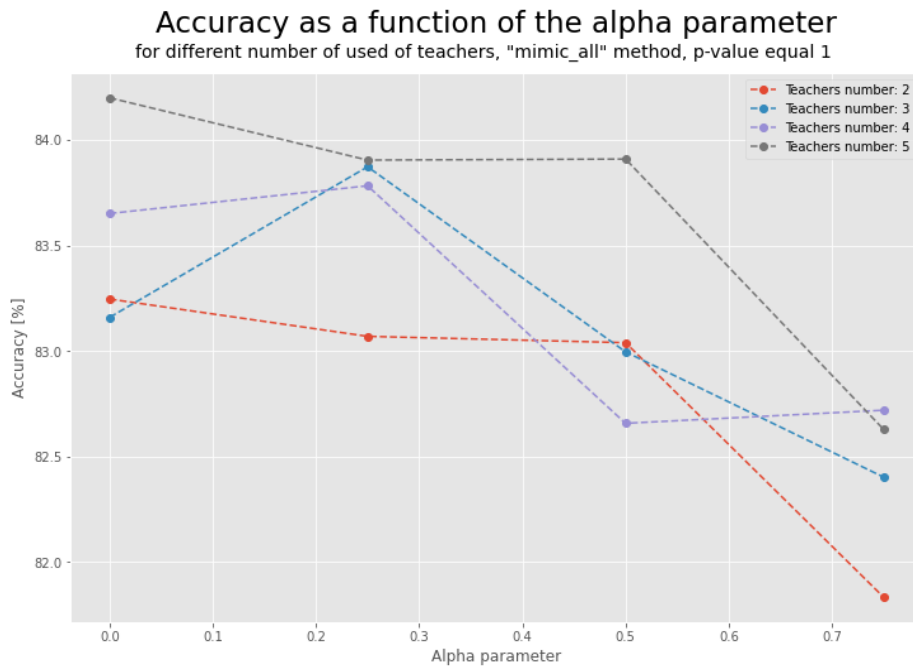

**Figure A.4:** Obtained classification accuracy depending on the *alpha* parameter used, different data series for different number of *teachers* used in training student model, the *p − value* is set to 1. The following data are presented for the ResNet50 baseline model and the CIFAR10 dataset.

| Teachers number | Fraction | Single accuracy [%] | Ensemble accuracy [%] | Multi-Teacher KD accuracy [%] | | |
|---|---|---|---|---|---|---|
| | | | | *Output avg.* | *Loss avg.* | *Mimick all* |
| 2 | 1 | 83.75 (0.22) | 84.76 (0.36) | 82.61 (0.4) | 83.06 (0.36) | **84.09 (0.57)** |
| | 0.9 | 82.63 (0.19) | 84.23 (0.27) | 82.23 (0.27) | 83.76 (0.28) | **83.98 (0.51)** |
| | 0.8 | 81.72 (0.77) | 82.54 (0.31) | 83.41 (0.55) | 82.43 (0.77) | **83.59 (0.31)** |
| | 0.7 | 80.98 (0.64) | 81.64 (0.54) | **83.17 (0.15)** | 82.48 (0.12) | 82.36 (0.68) |
| | 0.6 | 79.68 (0.97) | 80.54 (0.39) | 80.77 (0.83) | 80.73 (0.63) | **82.63 (0.95)** |
| 3 | 1 | 83.75 (0.22) | 85.45 (0.45) | 83.13 (0.74) | 81.79 (0.55) | **84.36 (0.72)** |
| | 0.9 | 82.63 (0.19) | 84.55 (0.27) | 82.76 (0.48) | 83.58 (0.34) | **83.79 (1.34)** |
| | 0.8 | 81.72 (0.77) | 83.18 (0.62) | 83.17 (0.35) | 83.35 (0.42) | **84.69 (0.55)** |
| | 0.7 | 80.98 (0.64) | 81.87 (0.41) | 83.41 (0.18) | **83.47 (0.37)** | 83.38 (1.34) |
| | 0.6 | 79.68 (0.97) | 81.41 (0.28) | 81.55 (0.25) | 82.38 (0.73) | **83.05 (0.68)** |
| 4 | 1 | 83.75 (0.22) | 85.6 (0.3) | 82.08 (0.4) | 83.99 (0.33) | **84.57 (0.74)** |
| | 0.9 | 82.63 (0.19) | 85.94 (0.69) | 81.97 (0.24) | 82.7 (0.21) | **84.1 (0.45)** |
| | 0.8 | 81.72 (0.77) | 83.83 (0.49) | 83.72 (0.58) | 82.97 (0.22) | **83.85 (0.47)** |
| | 0.7 | 80.98 (0.64) | 82.7 (0.24) | 82.02 (0.79) | 82.88 (0.14) | **83.58 (0.57)** |
| | 0.6 | 79.68 (0.97) | 81.42 (0.98) | 82.0 (0.55) | 81.62 (0.58) | **82.91 (1.33)** |
| 5 | 1 | 83.75 (0.22) | 85.88 (0.41) | 84.53 (0.37) | 84.32 (0.26) | **84.69 (1.04)** |
| | 0.9 | 82.63 (0.19) | 85.91 (0.26) | 83.82 (0.36) | 83.51 (0.22) | **84.22 (0.48)** |
| | 0.8 | 81.72 (0.77) | 84.17 (0.55) | 83.47 (0.52) | 82.38 (0.32) | **84.6 (0.27)** |
| | 0.7 | 80.98 (0.64) | 83.12 (0.51) | 83.46 (1.06) | **83.64 (0.19)** | 83.4 (0.92) |
| | 0.6 | 79.68 (0.97) | 82.13 (0.69) | 82.85 (0.42) | 81.93 (0.44) | **83.55 (0.84)** |

**Table A.3:** Classification accuracy comparison for a single model, ensemble model, and student model obtained using various multi-teacher knowledge distillation techniques of *teachers* mimicking. The following data are presented for the DenseNet121 baseline model and the CIFAR10 dataset.

| Fraction | Alpha parameter | | | |
|---|---|---|---|---|
| | 0 | 0.25 | 0.5 | 0.75 |
| | Accuracy [%] | | | |
| 1 | 84.62 (0.17) | **85.2 (0.1)** | 84.25 (0.51) | 83.37 (0.13) |
| 0.9 | 82.18 (0.35) | **85.43 (0.21)** | 84.99 (0.25) | 83.1 (0.28) |
| 0.8 | 84.91 (0.3) | 84.93 (0.01) | **85.12 (0.26)** | 83.81 (0.04) |
| 0.6 | 81.74 (0.69) | 83.88 (0.26) | **84.85 (0.22)** | 83.89 (0.72) |
| 0.7 | 81.98 (0.19) | 83.08 (0.13) | 83.55 (0.23) | **83.58 (0.24)** |

**Table A.4:** Classification accuracy for student models obtained using various *alpha* and *p − value*. The number of *teachers* was set to 3 and *teachers* mimicking method was *mimick all*. The following data are presented for the DenseNet121 baseline model and the CIFAR10 dataset.

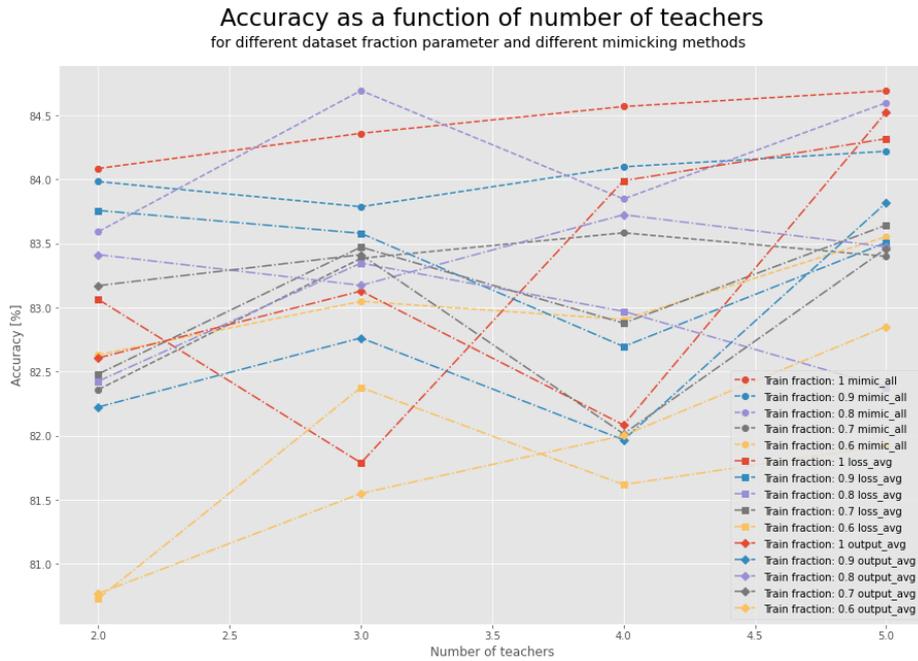

**Figure A.5:** Student classification accuracy depending on *teachers* number for different mimicking method and $p - value$. The following data are presented for the DenseNet121 baseline model and the CIFAR10 dataset.

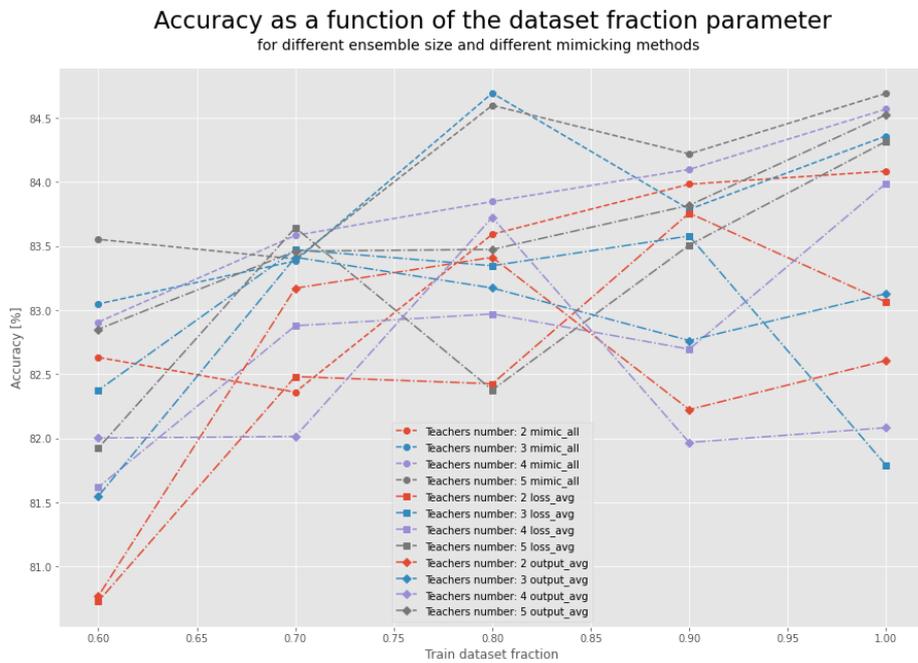

**Figure A.6:** Student classification accuracy depending on $p - value$ for different *teachers* number and used mimicking method. The following data are presented for the DenseNet121 baseline model and the CIFAR10 dataset.

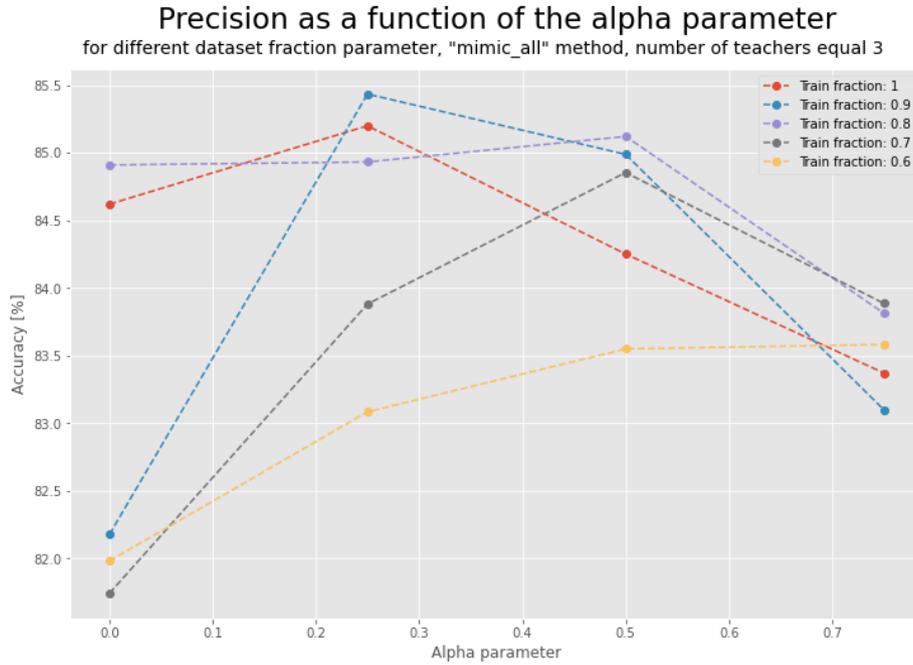

**Figure A.7:** Student classification accuracy depending on *alpha* parameter used, different data series for different $p-value$, the number of *teachers* used for training student are set to 3. The following data are presented for the DenseNet121 baseline model and the CIFAR10 dataset.

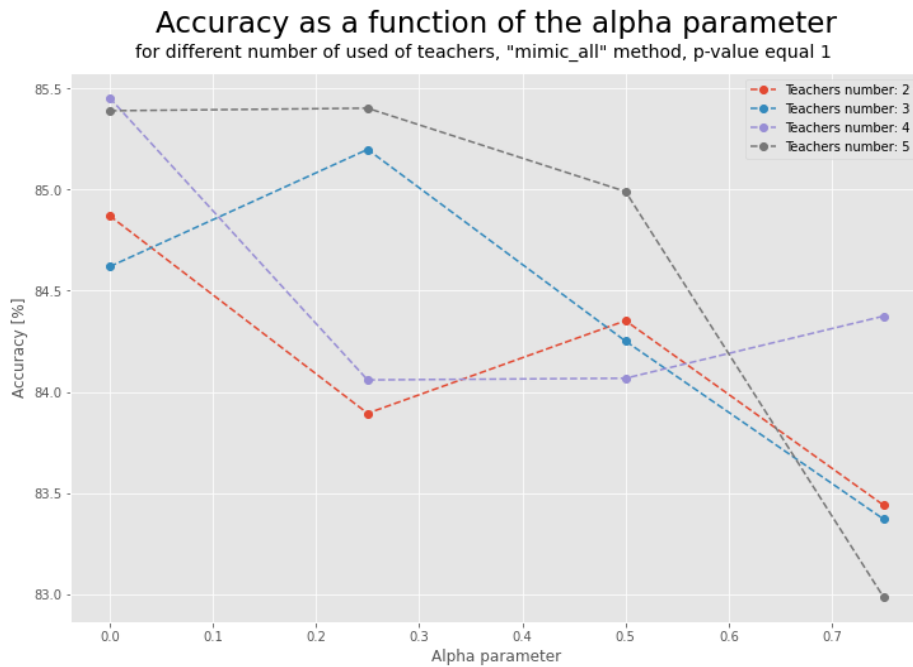

**Figure A.8:** Obtained classification accuracy depending on the *alpha* parameter used, different data series for different number of *teachers* used in training student model, the $p-value$ is set to 1. The following data are presented for the DenseNet121 baseline model and the CIFAR10 dataset.

| Teachers number | Fraction | Single accuracy [%] | Ensemble accuracy [%] | Multi-Teacher KD accuracy [%] | | |
|---|---|---|---|---|---|---|
| | | | | *Output avg.* | *Loss avg.* | *Mimick all* |
| 2 | 1 | 63.03 (0.77) | 64.0 (0.32) | 62.99 (0.27) | 62.86 (0.91) | **63.66 (0.9)** |
| | 0.9 | 65.89 (0.81) | 67.25 (0.85) | 62.95 (0.67) | 61.99 (0.44) | **63.28 (0.96)** |
| | 0.8 | 66.95 (0.55) | 67.53 (0.42) | 62.7 (0.32) | 61.85 (0.67) | **63.14 (0.82)** |
| | 0.7 | 66.52 (0.67) | 66.55 (0.49) | **63.09 (0.57)** | 61.71 (0.56) | 61.81 (0.52) |
| | 0.6 | 66.02 (0.74) | 66.53 (0.66) | 58.67 (1.27) | 60.36 (0.59) | **61.79 (1.42)** |
| 3 | 1 | 63.03 (0.77) | 64.89 (0.75) | 65.81 (1.01) | 65.78 (1.21) | **67.75 (1.09)** |
| | 0.9 | 65.89 (0.81) | 67.16 (0.5) | 65.44 (0.56) | 65.82 (0.21) | **67.12 (1.09)** |
| | 0.8 | 66.95 (0.55) | 67.86 (1.01) | 67.07 (0.7) | 65.63 (0.6) | **67.11 (0.63)** |
| | 0.7 | 66.52 (0.67) | 68.33 (0.88) | 65.58 (0.3) | **66.58 (0.92)** | 66.28 (1.54) |
| | 0.6 | 66.02 (0.74) | 67.64 (0.61) | 65.75 (0.89) | **66.26 (1.11)** | 65.88 (1.3) |
| 4 | 1 | 63.03 (0.77) | 65.64 (0.49) | 68.67 (1.05) | 68.87 (0.38) | **69.48 (0.91)** |
| | 0.9 | 65.89 (0.81) | 69.16 (0.73) | 66.4 (1.06) | 68.0 (0.43) | **69.21 (0.64)** |
| | 0.8 | 66.95 (0.55) | 67.6 (0.37) | 68.59 (0.7) | 67.22 (0.23) | **69.07 (0.95)** |
| | 0.7 | 66.52 (0.67) | 67.83 (0.55) | 66.53 (1.6) | 66.89 (0.42) | **68.79 (0.62)** |
| | 0.6 | 66.02 (0.74) | 68.7 (0.39) | 67.09 (0.34) | 66.18 (1.02) | **67.8 (0.94)** |
| 5 | 1 | 63.03 (0.77) | 65.32 (0.73) | **70.42 (0.22)** | 69.22 (0.72) | 70.26 (0.78) |
| | 0.9 | 65.89 (0.81) | 69.13 (0.67) | 68.87 (0.82) | 68.94 (0.73) | **69.62 (0.63)** |
| | 0.8 | 66.95 (0.55) | 68.7 (0.83) | 69.49 (0.45) | 68.42 (0.46) | **70.17 (0.69)** |
| | 0.7 | 66.52 (0.67) | 67.55 (0.78) | 68.68 (0.89) | **69.19 (0.74)** | 68.75 (1.0) |
| | 0.6 | 66.02 (0.74) | 69.3 (0.95) | 67.87 (0.65) | 68.43 (1.28) | **68.67 (1.24)** |

**Table A.5:** Classification accuracy comparison for a single model, ensemble model, and student model obtained using various multi-teacher knowledge distillation techniques of *teachers* mimicking. The following data are presented for the ResNet50 baseline model and the CIFAR100 dataset.

| Fraction | Alpha parameter | | | |
|---|---|---|---|---|
| | 0 | 0.25 | 0.5 | 0.75 |
| | Accuracy [%] | | | |
| 1 | 66.86 (0.31) | **69.2 (0.97)** | 67.48 (0.82) | 67.46 (0.05) |
| 0.9 | 66.28 (0.8) | 67.97 (0.53) | **68.21 (0.15)** | 66.29 (0.67) |
| 0.8 | 67.31 (0.05) | 67.44 (0.02) | **67.55 (0.51)** | 66.13 (0.06) |
| 0.6 | 64.42 (1.04) | **67.25 (1.14)** | 67.23 (0.1) | 67.15 (0.05) |
| 0.7 | 64.03 (0.41) | 65.75 (0.26) | **66.95 (0.44)** | 66.77 (0.96) |

**Table A.6:** Classification accuracy for student models obtained using various *alpha* and *p − value*. The number of *teachers* was set to 3 and mimicking method was *mimick all*. The following data are presented for the ResNet50 baseline model and the CIFAR100 dataset.

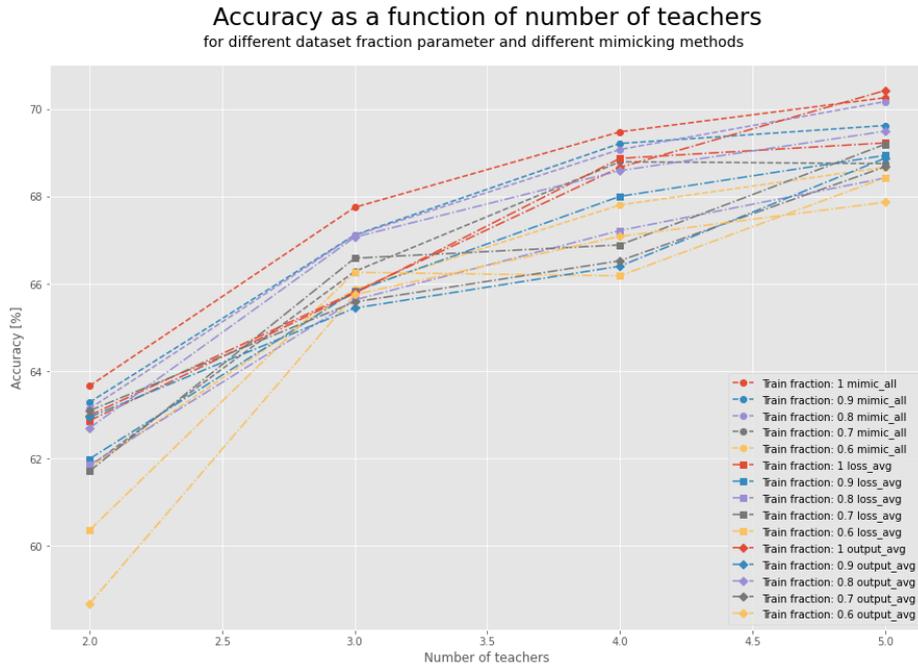

**Figure A.9:** Student classification accuracy depending on *teachers* number for different mimicking method and $p-value$. The following data are presented for the ResNet50 baseline model and the CIFAR100 dataset.

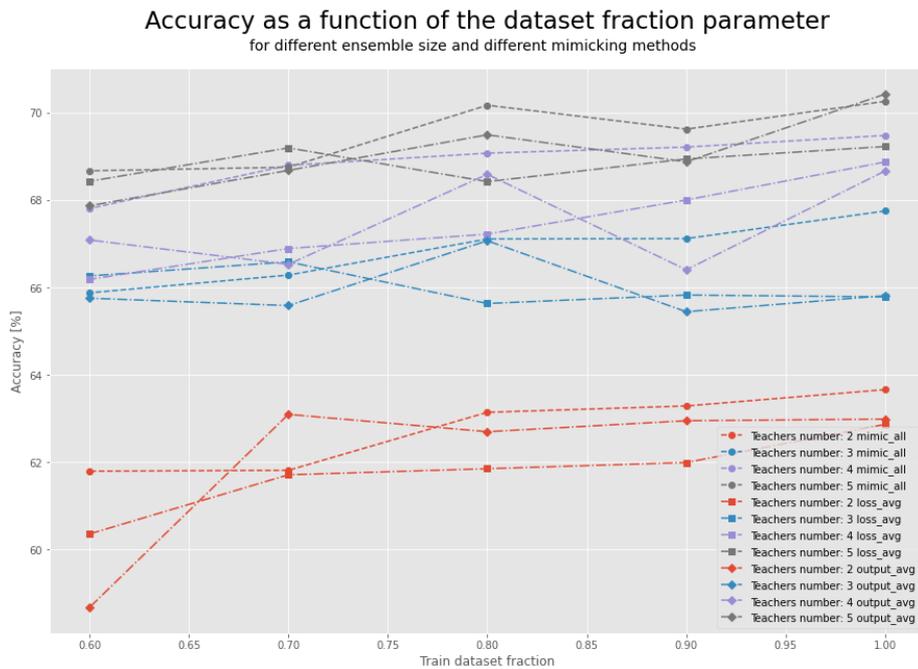

**Figure A.10:** Student classification accuracy depending on $p-value$ for different *teachers* number and used mimicking method. The following data are presented for the ResNet50 baseline model and the CIFAR100 dataset.

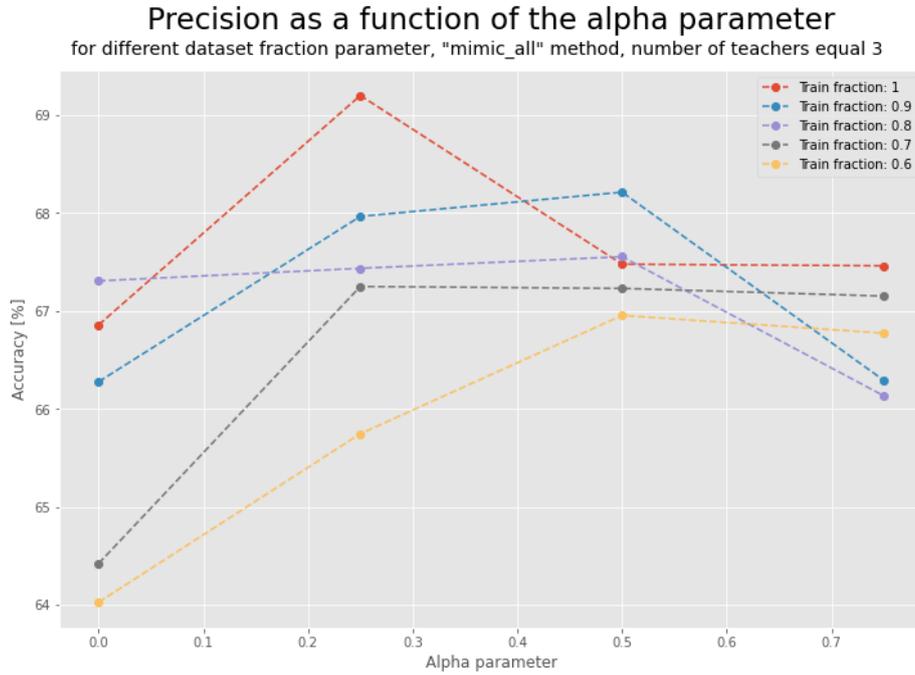

**Figure A.11:** Student classification accuracy depending on $alpha$ parameter used, different data series for different $p-value$, the number of $teachers$ used for training student are set to 3. The following data are presented for the ResNet50 baseline model and the CIFAR10 dataset.

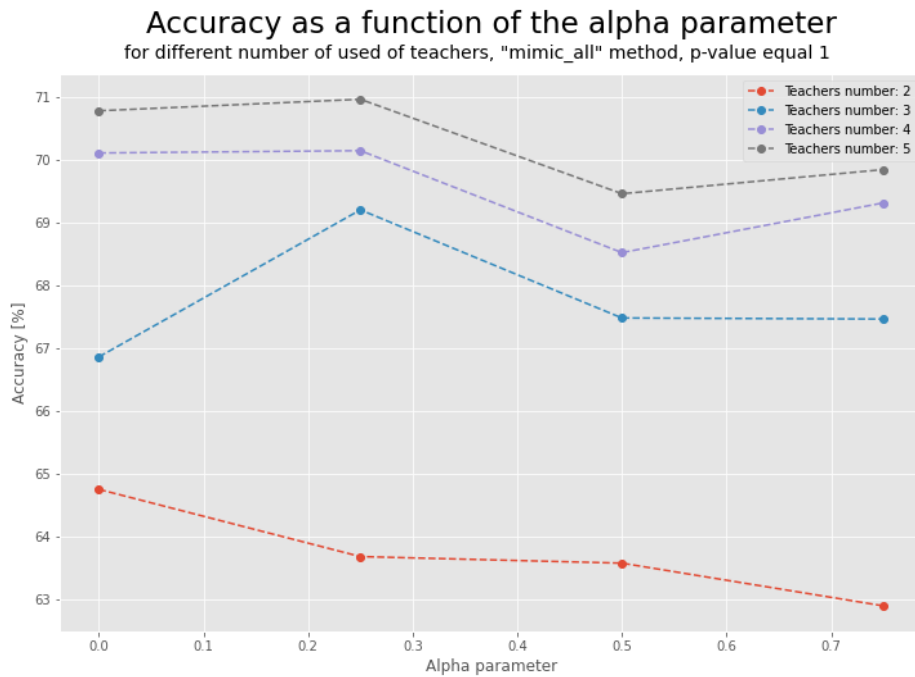

**Figure A.12:** Obtained classification accuracy depending on the $alpha$ parameter used, different data series for different number of $teachers$ used in training student model, the $p-value$ is set to 1. The following data are presented for the ResNet50 baseline model and the CIFAR100 dataset.

| Teachers number | Fraction | Single accuracy [%] | Ensemble accuracy [%] | Multi-Teacher KD accuracy [%] | | |
|---|---|---|---|---|---|---|
| | | | | *Output avg.* | *Loss avg.* | *Mimick all* |
| 2 | 1 | 64.61 (0.72) | 65.95 (0.54) | 64.3 (0.58) | 64.52 (1.67) | **65.22 (0.82)** |
| | 0.9 | 66.61 (0.72) | 68.96 (0.71) | 64.54 (0.41) | **64.8 (0.21)** | 64.28 (0.58) |
| | 0.8 | 67.64 (0.95) | 69.02 (0.72) | 63.25 (1.14) | **64.16 (1.61)** | 63.92 (0.46) |
| | 0.7 | 68.41 (0.91) | 68.38 (0.77) | **64.3 (0.56)** | 63.85 (0.47) | 64.19 (0.74) |
| | 0.6 | 67.52 (1.55) | 68.38 (0.6) | 62.5 (0.51) | 63.27 (1.18) | **63.94 (0.89)** |
| 3 | 1 | 64.61 (0.72) | 66.74 (0.81) | 67.71 (1.63) | 67.39 (0.42) | **69.07 (0.95)** |
| | 0.9 | 66.61 (0.72) | 68.74 (0.73) | 66.47 (0.25) | 67.41 (0.25) | **68.62 (1.01)** |
| | 0.8 | 67.64 (0.95) | 69.19 (0.39) | 66.81 (0.67) | 67.67 (0.4) | **69.02 (0.97)** |
| | 0.7 | 68.41 (0.91) | 68.42 (0.78) | **68.49 (0.35)** | 66.42 (0.43) | 68.3 (1.3) |
| | 0.6 | 67.52 (1.55) | 68.59 (0.87) | 64.65 (0.48) | 65.68 (0.3) | **67.2 (0.96)** |
| 4 | 1 | 64.61 (0.72) | 67.36 (0.81) | 67.65 (1.68) | 70.62 (0.56) | **70.89 (0.76)** |
| | 0.9 | 66.61 (0.72) | 70.58 (0.29) | 68.66 (0.71) | 68.63 (0.82) | **70.34 (0.7)** |
| | 0.8 | 67.64 (0.95) | 70.44 (0.1) | 69.83 (0.53) | 70.35 (0.67) | **70.47 (1.09)** |
| | 0.7 | 68.41 (0.91) | 69.8 (0.7) | 69.03 (1.49) | 69.35 (0.31) | **69.7 (0.46)** |
| | 0.6 | 67.52 (1.55) | 69.96 (0.76) | 68.48 (0.47) | 66.17 (1.08) | **69.16 (1.56)** |
| 5 | 1 | 64.61 (0.72) | 67.48 (0.59) | 70.92 (0.24) | 70.88 (0.57) | **71.35 (1.0)** |
| | 0.9 | 66.61 (0.72) | 70.7 (0.63) | 71.36 (0.38) | 70.07 (0.21) | **71.88 (0.81)** |
| | 0.8 | 67.64 (0.95) | 70.26 (0.85) | 71.15 (0.95) | 69.82 (0.42) | **71.32 (0.64)** |
| | 0.7 | 68.41 (0.91) | 69.27 (0.9) | 69.81 (0.57) | 70.44 (0.71) | **70.78 (0.55)** |
| | 0.6 | 67.52 (1.55) | 69.48 (0.33) | 68.66 (0.97) | 68.9 (1.07) | **70.4 (0.86)** |

**Table A.7:** Classification accuracy comparison for a single model, ensemble model, and student model obtained using various multi-teacher knowledge distillation techniques of *teachers* mimicking. The following data are presented for the DenseNet121 baseline model and the CIFAR100 dataset.

| Fraction | Alpha parameter | | | |
|---|---|---|---|---|
| | 0 | 0.25 | 0.5 | 0.75 |
| | Accuracy [%] | | | |
| 1 | 67.93 (0.14) | **70.24 (0.41)** | 69.63 (0.16) | 68.46 (0.13) |
| 0.9 | 67.96 (0.67) | 69.16 (0.01) | **69.56 (0.1)** | 67.99 (1.19) |
| 0.8 | 69.02 (0.26) | **70.24 (0.02)** | 69.13 (0.62) | 67.68 (0.2) |
| 0.6 | 67.3 (0.98) | **69.18 (1.66)** | 68.2 (0.06) | 69.04 (0.73) |
| 0.7 | 65.82 (0.79) | **67.69 (0.25)** | 67.66 (0.67) | 67.62 (0.2) |

**Table A.8:** Classification accuracy for student models obtained using various *alpha* and $p - value$. The number of *teachers* was set to 3 and mimicking method was *mimick all*. The following data are presented for the DenseNet121 baseline model and the CIFAR100 dataset.

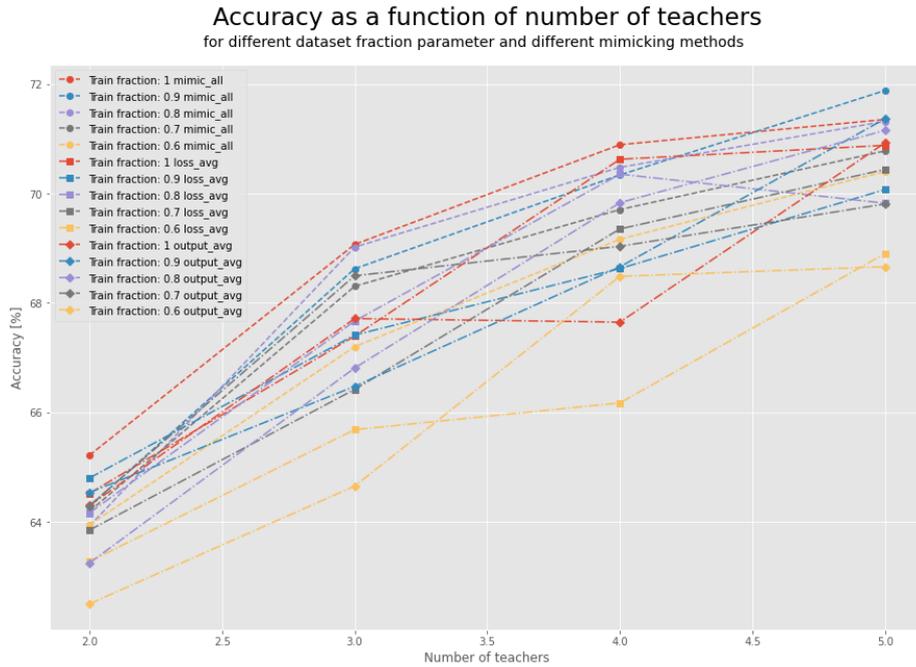

**Figure A.13:** Student classification accuracy depending on *teachers* number for different mimicking method and *p − value*. The following data are presented for the DenseNet121 baseline model and the CIFAR100 dataset.

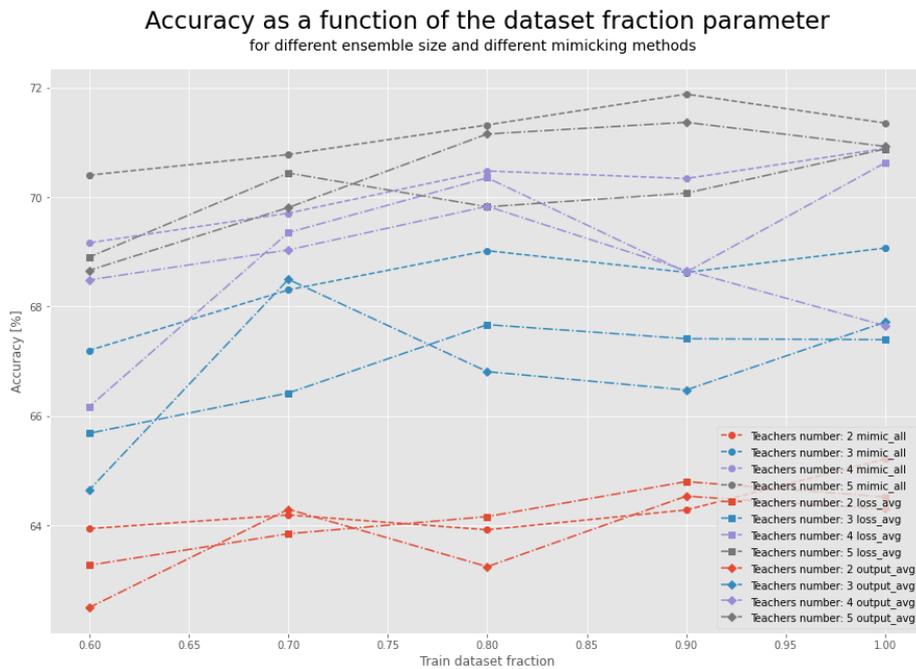

**Figure A.14:** Student classification accuracy depending on *p − value* for different *teachers* number and used mimicking method. The following data are presented for the DenseNet121 baseline model and the CIFAR100 dataset.

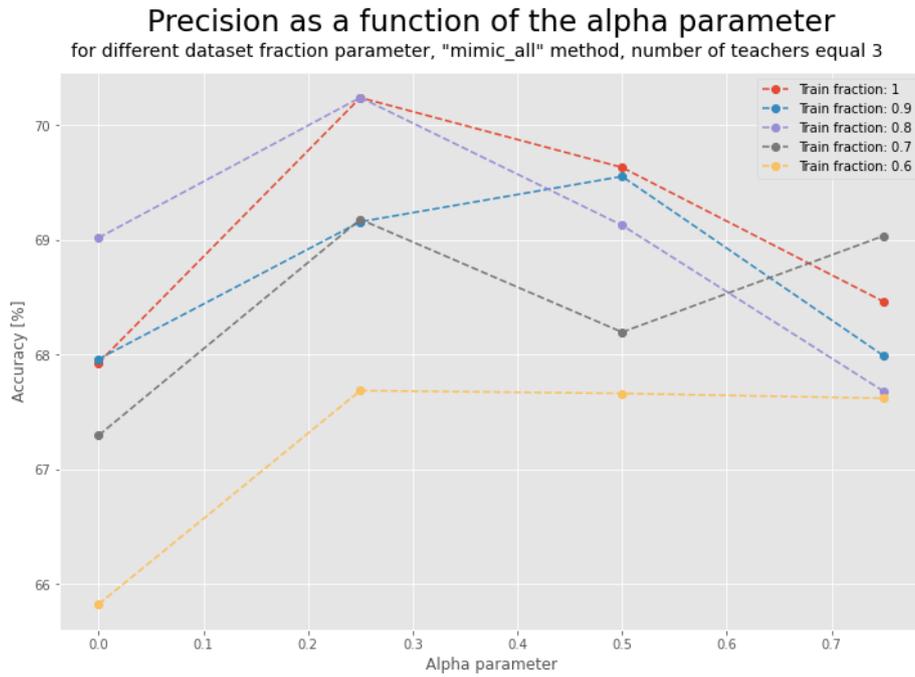

**Figure A.15:** Student classification accuracy depending on *alpha* parameter used, different data series for different $p - value$, the number of *teachers* used for training student are set to 3. The following data are presented for the DenseNet121 baseline model and the CIFAR100 dataset.

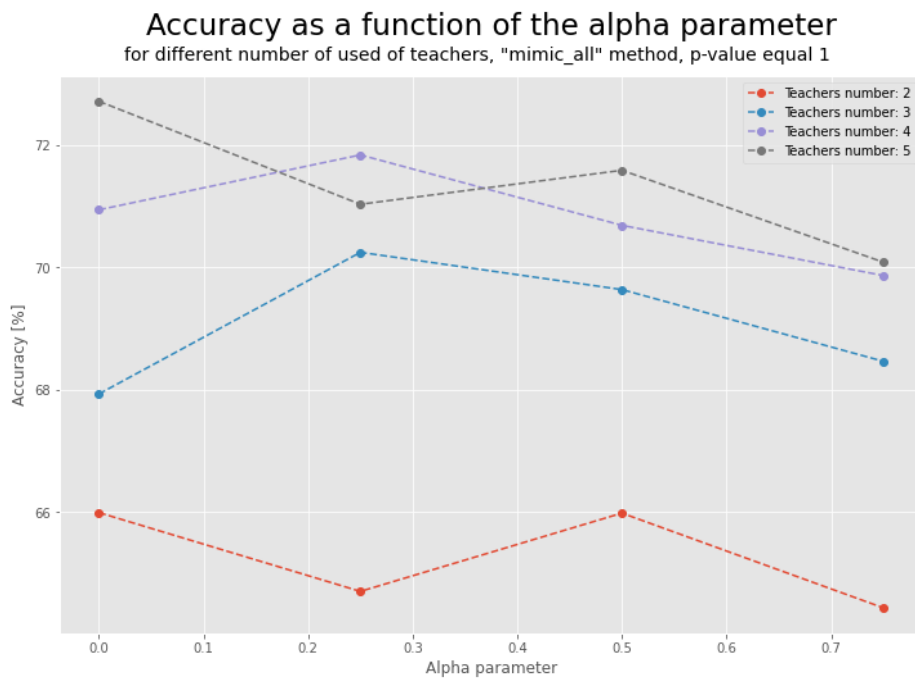

**Figure A.16:** Obtained classification accuracy depending on the *alpha* parameter used, different data series for different number of *teachers* used in training student model, the $p - value$ is set to 1. The following data are presented for the DenseNet121 baseline model and the CIFAR100 dataset.

# B
## Publications list

**Author's publications**

- Konrad Zuchniak et al. "Corrosion detection on aircraft fuselage with multi-teacher knowledge distillation". In: *International Conference on Computational Science*. Springer. 2021, pp. 318–332

- Anna Kolonko et al. "SuperNet: using supermodeling in developing more efficient data models". In: *AMMCS interdisciplinary conference*. 2021, p. 118

- Zdzislaw Burda, Pawel Wojcieszak, and Konrad Zuchniak. "Dynamics of wealth inequality". In: *Comptes Rendus Physique* 20.4 (2019), pp. 349–363

- Konrad Zuchniak. "Application of Committee Scoring Rules in decision-making process in ensemble classifiers". In: *CGW 18 International Workshop*. Academic Computer Centre CYFRONET AGH. 2018, pp. 43–44

- Konrad Zuchniak. "Cropping input image can lead to a better training of convolutional neural networks". In: *CGW 17 International Workshop*. Academic Computer Centre CYFRONET AGH. 2017, pp. 37–38